\documentclass{article}

\usepackage{arxiv}

\usepackage[utf8]{inputenc} % allow utf-8 input
\usepackage[T1]{fontenc}    % use 8-bit T1 fonts
\usepackage{hyperref}       % hyperlinks
\usepackage{url}            % simple URL typesetting
\usepackage{booktabs}       % professional-quality tables
\usepackage{amsfonts}       % blackboard math symbols
\usepackage{nicefrac}       % compact symbols for 1/2, etc.
\usepackage{microtype}      % microtypography
\usepackage{lipsum}
\usepackage{amssymb,amsmath,mathrsfs,amstext,amsthm,cite}
\usepackage{bm}
\usepackage{graphicx,url,epstopdf,multirow,color,rotating,pifont,url,stmaryrd}
\usepackage[ruled, vlined, lined,boxed,linesnumbered]{algorithm2e}
\usepackage{algorithmic}
\usepackage{tabularx,subcaption}
\usepackage{titling}

\theoremstyle{definition}
\newtheorem{exmp}{Example}[section]

\newcommand{\vect}{\mathrm{vec}}

\newcommand{\real}[1]{\mathrm{I \! R} \mathit{^{#1}}}
\newcommand{\trans}{^{\mbox{\tiny {\sffamily T}}}}

\newcommand{\diag}{\mathrm{diag}}

\newcommand{\Abf}{{\bm A}}
\newcommand{\Bbf}{{\bm B}}

\newcommand{\Gbf}{{\bm G}}

\newcommand{\Ibf}{{\bm I}}

\newcommand{\Sbf}{{\bm S}}

\newcommand{\Xbf}{{\bm X}}

\newcommand{\Zbf}{{\bm Z}}

\newcommand{\xbf}{{\bm x}}

\newcommand{\zbf}{{\bm z}}

\newcommand{\greekbold}[1]{\mbox{\boldmath $#1$}}

\newcommand{\betabf}{\greekbold{\beta}}

\newcommand{\gammabf}{\greekbold{\gamma}}

\newcommand{\lambdabf}{\greekbold{\lambda}}

\newcommand{\Lambdabf}{\greekbold{\Lambda}}

\newcommand{\Bcal}{{\mathcal B}}

\title{Sparse Symmetric Tensor Regression for Functional Connectivity Analysis}

\author{
  Da Xu\thanks{Email: xudastar@berkeley.edu. The work was done in partial satisfaction of the requirements for the graduate degree of Division of Biostatistics, University of California, Berkeley. Program committee chair: Lexin Li, email: lexinli@berkeley.edu.} \\
}
\date{\vspace{-5ex}}

\begin{document}
\maketitle

\begin{abstract}
Tensor regression models, such as CP regression \cite{zhou2013tensor} and Tucker regression \cite{li2018tucker}, have many successful applications in neuroimaging analysis where the covariates are of ultrahigh dimensionality and possess complex spatial structures. The high-dimensional covariate arrays, also known as tensors, can be approximated by low-rank structures and fit into the generalized linear models. The resulting tensor regression achieves a significant reduction in dimensionality while remaining efficient in estimation and prediction.
Brain functional connectivity is an essential measure of brain activity and has shown significant association with neurological disorders such as Alzheimer's disease. The symmetry nature of functional connectivity is a property that has not been explored in previous tensor regression models. In this work, we propose a sparse symmetric tensor regression that further reduces the number of free parameters and achieves superior performance over symmetrized and ordinary CP regression, under a variety of simulation settings. We apply the proposed method to a study of Alzheimer's disease (AD) and normal ageing from the Berkeley Aging Cohort Study (BACS) and detect two regions of interest that have been identified important to AD.
\end{abstract}

% keywords can be removed
% \keywords{Tensor regression, Higher-order matrix, Functional connectivity, Neuroimage}

\section{Introduction}
\label{sec:introduction}

Brain functional connectivity reveals the synchronization of brain systems through correlations in neurophysiological measures of brain activity. When measured during resting-state, it maps the intrinsic functional architecture of the brain \cite{Varoquaux2013}. Brain connectivity analysis is at the foreground of neuroscience research for studying cognitive ability and neurological disease pathologies \cite{Castellanos2013}. Accumulated evidence has suggested that alternations in brain connectivity network are predictive of cognitive function and decline, and hold crucial insights of pathologies of neurological disorders \cite{Fox2010}. In this article, we tackle the problem of association modelling between functional connectivity and phenotypic outcome. 
Our motivating example is a study of Alzheimer's disease (AD) and normal aging from the Berkeley Aging Cohort Study (BACS). This example also embodies a more general class of functional connectivity analysis and association modelling problems. AD, characterized by progressive impairment of cognitive and memory functions, is an irreversible neurodegenerative disorder and the leading form of dementia in elderly subjects. With the aging of the worldwide population, the number of affected people is drastically increasing, and thus it is an international imperative to understand, diagnose, and treat this disorder. Numerous studies have demonstrated that brain networks degrade during both symptomatic AD and the preclinical phase, in which amyloid-beta (A$\beta$) plaques accumulate \cite{Hedden2009,Sheline2010,Mormino2011,Brier2014}. A$\beta$ is a form of protein that is toxic to neurons in the brain, and it accumulates outside neurons and forms sticky buildup called A$\beta$ plaques. A$\beta$ plaques destroy synapses, i.e., contact points via which nerve cells relay signals to one another, and eventually lead to nerve cell death. A$\beta$ plaques are the hallmark neuropathology markers of Alzheimer's disease (AD), and are also commonly found in elderly normal controls. Our study aims to establish an interpretable association model between brain connectivity network and A$\beta$ deposition, and to identify links among individual brain subregions that mostly affect A$\beta$. This, in turn, will deepen our understanding of both AD pathologies as well as normal aging. The A$\beta$ deposition was measured by Pittsburgh compound-B positron emission tomography (PIB-PET) imaging, and the brain connectivity was measured by resting-state functional magnetic resonance imaging (rs-fMRI).  

As subject lies in the scanner at rest, rs-fMRI data is acquired as a four-dimensional array, with around $10^4$ to $10^5$ spatial voxels over $100$ to $200$ time points captured about every two seconds. Directly studying brain connectivity network at the voxel level is not desirable, since most connections would be due to close spatial proximity \cite{Craddock2013}. As such, most studies parcellate the brain scan by mapping voxels to prespecified brain regions, and average time series of the voxels within the same region, resulting in a region by time form of matrix data. Then, for each subject, an undirected brain network is constructed, taking the form of a connectivity matrix, where nodes represent brain regions, and links measure interaction and dependence between nodes through some correlation measure. A good number of correlation measures have been proposed to depict the connectivity matrix. Common choices include Pearson correlation and partial correlation in the time domain \cite{Bullmore2009, Zalesky2012, Ryali2012, Chen2013}, coherence, partial coherence, mutual information, and partial mutual information in the frequency domain \cite{Fiecas2011,Cassidy2015}. Two previous papers have detailed discussions for such methods \cite{Craddock2013,Cassidy2015}. In our analysis, it is \emph{not} our intension to argue which correlation measure is the best connectivity measure. Instead, our method can take \emph{any} connectivity matrix as input for our association modelling. It is also useful to note that, in spite of the choice of correlation measure, the resulting connectivity is always a \emph{symmetric} matrix. 

After the construction of a brain connectivity network, most clinical applications of functional connectivity analysis have been focusing on the comparison of networks across different populations of subjects, for instance, between the group with a neurological disorder and the normal control \cite{Meskaldji2013,Varoquaux2013,Castellanos2013,ChenKang2015}. A number of methods directly modelled biological and clinical phenotypes given the connectivity network using machine learning techniques \cite{wee2012resting,wee2014group}. Most of those solutions, however, construct the connectivity network as a few global or node (region) based \emph{network metrics}. Some commonly used network metrics include global efficiency, characteristic path length, local efficiency and clustering coefficient \cite{Rubinov2010,Kim2014}. Summarizing a network in the form of a set of network metrics transforms the problem into a classical statistical learning framework, and has proven to shed useful insight on disease pathologies. However, we must carefully consider the extent to which each network metric provides a meaningful representation of brain function \cite{Fornito2013}. In addition, most network metrics are defined on a binary network, i.e., the link between brain regions is either zero or one, which in turn require appropriate thresholding of the correlation matrix. On the other hand, there is no unanimous agreement on what is the best thresholding rule, nor what network metrics best characterize brain functions. More recently, there have emerged a few solutions that directly utilized connectivity matrix as features in predictive modelling, again using machine learning techniques \cite{chen2011classification,zhu2013connectome,jie2014topological}. Nevertheless, interpretation and parameter tuning of such models are often challenging. 

\begin{figure}[t]
\centering
\includegraphics[scale=0.3]{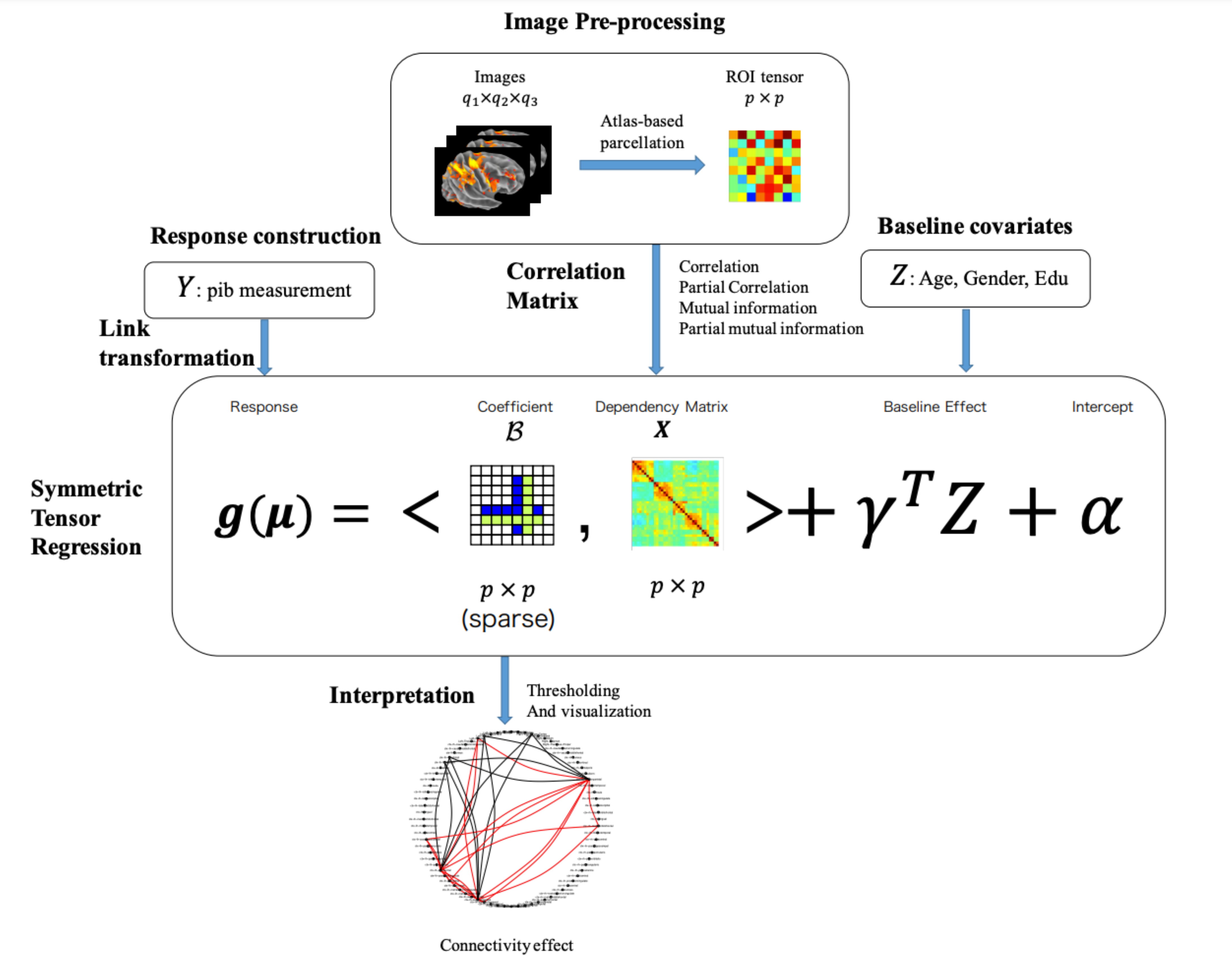} 
\caption{A schematic overview of the proposed sparse symmetric tensor regression for association modeling of brain functional connectivity.}
\label{fig:scheme}
\end{figure}

In this work, we describe a class of regression models that directly associate brain connectivity to the biological and clinical phenotypes. A schematic overview of our proposed method is given in Figure~\ref{fig:scheme}. The new model takes the phenotype, which can be either categorical or continuous, as a scalar response, and the symmetric connectivity matrix as a predictor. The regression coefficient is given by a symmetric matrix, which fully characterizes the effect of individual links of brain regions on the response. The model is capable of incorporating additional covariates such as age and gender. Classical regression models treat predictors as a vector and estimate a corresponding vector of regression coefficients. For our problem, directly vectorizing the symmetric matrix predictor is not desirable, as it would both results in a very high dimensionality, and destroys all the inherent structure embedded in the matrix predictor and coefficient. To circumvent the curse of dimensionality, as well as to preserve the spatial structures, we introduce a symmetric low rank decomposition on the associated matrix coefficient. It effectively reduces the dimension and leads to efficient estimation and prediction. Our proposal extends a recent regression model for a predictor of multidimensional array, a.k.a.\ tensor \cite{ZhouLiZhu2013}. Since the matrix is two-dimensional tensor, our method can be viewed, in a sense, as a modification of the CP tensor regression \cite{ZhouLiZhu2013}. On the other hand, the nature of the correlation matrix suggests the symmetry of the coefficient matrix and thus imposes additional constraint. This constraint, in turn, leads to a completely new and more challenging optimization problem than that of \cite{ZhouLiZhu2013}. Accordingly, we develop a highly scalable optimization algorithm based on the proximal gradient method.

Our proposal is novel and useful in several ways. First, it offers a parsimonious model that is both flexible and interpretable. It adopts the widely used generalized linear regression framework and can handle both continuous and discrete phenotypes. It models the connectivity matrix as input directly, leading to a straightforward interpretation of the effect of individual links between brain regions on the phenotype, while avoiding the issue of selecting thresholding rule or network metrics. Second, it permits both individual variations of functional connectivity and inference at the individual level. It is known that functional connectivity may exhibit significant variations across individuals \cite{Sporns2013}. In our method, a connectivity network is constructed for each individual and fed into the regression as input. It does not require different individuals to have the same connectivity pattern. Moreover, the regression framework allows inference and prediction to be carried out for individual subject, which has useful clinical implications. Third, our method can take any connectivity matrix as input, and thus works for correlation measures in both time and frequency domains. This alleviates the potential issue of temporal variability in response latency in functional imaging data \cite{MihyeShenZhu2015}. Finally, we comment that our method is primarily motivated by rs-fMRI-based functional connectivity analysis; however, the method is equally applicable to other imaging modalities with symmetric structure and structural connectivity analysis.

\section{Preliminary}
\label{sec:preliminary}

We provide a brief review of the key notations and operations we use in this work. An extensive reference can be found in the survey paper \cite{KoldaBader09Tensor}.

\subsection{Notation}
\label{sec:notation}

A \emph{tensor} is a multidimensional array. \emph{Fiber} of a tensor is defined by fixing every index but one, and is the higher order analogue of matrix row and column. The \emph{vec operator} stacks the entries of a $D$-dimensional tensor $\Bbf \in \real{p_1 \times \cdots \times p_D}$ into a column vector, such that an entry $b_{i_1\ldots i_D}$ maps to the $j$-th entry of $\vect \, \Bbf$ where $j = 1 + \sum_{d=1}^D (i_d-1) \prod_{d'=1}^{d-1} p_{d'}$. The \emph{mode-$d$ matricization}, $\Bbf_{(d)}$, maps a tensor $\Bbf$ into a $p_d \times \prod_{d' \ne d} p_{d'}$ matrix, such that the $(i_1,\ldots,i_D)$ element of the array $\Bbf$ maps to the $(i_d,j)$ element of the matrix $\Bbf_{(d)}$, where $j = 1 + \sum_{d'\ne d} (i_{d'}-1) \prod_{d''<d',d'' \ne d} p_{d''}$. When $D=1$, we observe that $\vect \, \Bbf$ is the  same as vectorizing the mode-1 matricization $\Bbf_{(1)}$. 

The inner product $\langle \cdot, \cdot \rangle$ between two arrays is defined as $\langle \Bcal,\Xbf \rangle = \langle \vect \Bcal, \vect \Xbf \rangle = \sum_{i_1,\ldots,i_D} \beta_{i_1\ldots i_D} x_{i_1 \ldots i_D}$.

\subsection{CP and Tucker Tensor regression}
\label{sec:cp-tucker}
Let $Y$ denote the univariate response, $\Xbf$ denote the $D$-dimensional tensor predictor and let $\Zbf \in \real{p_0}$ denote the vector of additional covariates. In \cite{ZhouLiZhu2013} the authors proposed a tensor predictor regression to model the association between $Y$ and a general tensor $\Xbf \in \real{p_1 \times \ldots \times p_D}$ after adjusting for $\Zbf$. They adopted the classical generalized linear model (GLM) setting \cite{McCullaghNelder83GLMBook}, where $Y$ belongs to an exponential family with probability mass function or density
\begin{eqnarray*}
p(y|\theta,\phi) = \exp\left\{ \frac{y \theta - b(\theta)}{a(\phi)} + c(y,\phi) \right\},
\end{eqnarray*}
with $\theta$ and $\phi>0$ denoting the natural and dispersion parameters, and the corresponding link function 
\begin{eqnarray} \label{eqn:glm} 
g(\mu) =  \gammabf \trans \Zbf + \langle \Bcal,\Xbf \rangle.
\end{eqnarray}
The coefficient tensor $\Bcal$ involves a number of unknown parameters at the exponential order $\prod_{j=1}^{D} p_j$, and CP tensor regression imposes a low rank structure on $\Bcal$ \cite{ZhouLiZhu2013}. In particular, $\Bcal$ is assumed to admit a structure of rank-$R$ canonical decomposition or parallel factors:
\begin{eqnarray} \label{eqn:R-CP-decomp}
\Bcal_{\text{CP}} = \sum_{r=1}^R \betabf_1^{(r)} \circ \cdots \circ \betabf_D^{(r)},   
\end{eqnarray}
where $\betabf_d^{(r)} \in \real{p_d}, d=1,\ldots,D,r=1,\ldots,R$ are all column vectors, $\circ$ denotes the outer product, and $\Bbf$ cannot be written as a sum of less than $R$ outer products. For convenience, we use the shorthand $\Bcal_{\text{CP}} = \llbracket \Bbf_1,\ldots,\Bbf_D \rrbracket$ such that $\Bbf_d = [\betabf_d^{(1)}, \ldots, \betabf_d^{(R)}] \in \real{p_d \times R}$. Notice that the number of free parameters in $\Bbf$ is substantially reduced from the exponential order to the linear order $\sum_{j=1}^{D} p_j$. With the low-rank structure embedded in GLM, the optimization (for maximum likehood estimation) is made feasible by the key observation that, although $g(\mu)$ in is not joinly linear in $(\Bbf_1, \ldots,\Bbf_D)$, it is linear in each $\Bbf_d$. It suggests an algorithm that alternately updates $\Bbf_d$ while keeping other components fixed, so each update step reduces to a standard GLM fit. Furthermore, the authors introduced sparsity constraint to penalize the columns of $\betabf_{d}^{(r)}$, which in turn leads to selection of tensor regions that are most relevant to the response. Similarly, the regularized optimization can be decomposed into individual penalized GLM steps. 

In a follow-up work \cite{li2018tucker}, the authors propose the more flexible tensor regression model using the Tucker decomposition \cite{KoldaBader09Tensor}. It further reduces the number of free parameters and is able to better accommodate images with skewed dimensions. In the Tucker regression, the coefficient array $\Bbf$ admits a higher-order singular value decomposition:
\begin{eqnarray} \label{eqn:tucker}
\Bcal_{\text{Tucker}} = \sum_{r_1=1}^{R_1} \cdots \sum_{r_D=1}^{R_D} g_{r_1,\ldots,r_D} \betabf_{1}^{(r_1)} \circ \cdots \circ \betabf_{D}^{(r_D)}, 
\end{eqnarray}
where $\betabf_{d}^{(r_d)} \in \real{p_d}$ are all column vectors for $d=1,\ldots,D,r_d=1,\ldots,R_d$, and $g_{r_1,\ldots,r_D}$ are constants. We employ a similar shorthand: $\Bcal_{\text{Tucker}} = \llbracket \Gbf; \Bbf_1,\ldots,\Bbf_D \rrbracket$, where $\Gbf \in \real{R_1 \times \cdots \times R_D}$ is a $D$-dimensional \emph{core tensor} with entries $(\Gbf)_{r_1\ldots r_D} = g_{r_1,\ldots,r_D}$. 
Compared with the CP decomposition, Tucker decomposition allows $R_d$ to vary across dimensions so $\Bbf_d$ can have \emph{different} ranks. Therefore, the CP decomposition is a special case where the core tensor $\Gbf$ is super-diagonal.
In practice, the neuroimaging analysis can benefit from the flexibility of Tucker regression.
Similar to the CP decomposition, when combined with the GML, the systematic part is also jointly linear in $\Gbf$ and $\Bbf_d$, which suggests update $\Gbf$ and $\Bbf_d$ alternatively in a block ascent fashion. The sparsity constraints on $\betabf_{d}^{(r)}$ have been found helpful for avoiding overfitting.

Both tensor regression models exploit the low-rank structure of the tensor signal and reduce the dimensionality. The empirical evidence shows that they provide sound low rank approximations to the potentially high-rank signals. 

\section{Methods}
\label{sec:methods}

\subsection{Symmetric Tensor Regression}
\label{sec:sym-tensor}

Similar to the setup in CP and Tucker regression, we let $Y$ denote the univariate response, whereas the proposed method is readily extendable to the multivariate response setting as well. Let $\Xbf \in \real{p \times \ldots \times p}$ denote a $D$-dimensional symmetric tensor predictor. In this article, our application focuses on the two-dimensional symmetric matrices. Nevertheless, we develop the methodology for a general $D$-dimensional symmetric tensor. We use $\Zbf \in \real{p_0}$ to represent the vector of the additional covariates, such as age, gender and education. The lower-case triplets $(y_i, \xbf_i, \zbf_i)$, $i=1,\ldots,n$ denote the observed independent sample instances of $(Y, \Xbf, \Zbf)$. 

Without loss of generality, we omit the intercept term in the GLM model (given by \eqref{eqn:glm}), as we can always center the data in advance. For our purpose, we use $\Bcal \in \real{p_1 \times \ldots \times p_D}$ to capture the individual effect of the elements of $\Xbf$ on $Y$, after adjusting for the intercept term $\alpha \in \real{}$ and the covariate effect $\gammabf \in \real{p_0}$ of $\Zbf$. 

When $\Xbf \in \real{p \times \ldots \times p}$ is a symmetric tensor given by such as a correlation matrix, the corresponding coefficient tensor $\Bcal$ should also be symmetric. It introduces an additional constraint to the parameterization of $\Bcal$. We continue to adopt the GLM formulation but require $\Bcal$ to be symmetric in that 
\begin{eqnarray} \label{eqn:sy-CP-decomp}
\Bcal = \llbracket \lambdabf; \Bbf,\ldots,\Bbf \rrbracket, 
\end{eqnarray}
where $\lambdabf = (\lambda_1,\ldots,\lambda_R)$, and $\Bbf \in \real{p \times R}$. In other words, the symmetry of $\Bcal$ requires all the individual matrices $\Bbf_1, \ldots, \Bbf_D$ in \eqref{eqn:R-CP-decomp} to be the same up to a constant. The parameterization in \eqref{eqn:sy-CP-decomp} follows the symmetric nature of the coefficient tensor, and leads to  straightforward interpretation. In addition, we further reduce the complexity of tensor model from the order of $pRD$ to $pR$. However, the additional symmetric constraint brings considerable complications to the optimization as we shall no longer update $\Bbf$ iteratively. We leave the optimization details to the next section. 

In addition to the imposed low-rank structure, we also add the sparsity requirement on $\Bcal$, so that only a subset of elements of $\Bcal$ are nonzero. The sparsity constraint, which is imposed via the usual $\ell_1$ type penalty such as Lasso \cite{Tibshirani1996}, allows one to identify individual links among a set of brain regions whose interactions directly affect the phenotypic outcome. It is known that the $\ell_1$ regularization improves the generalization performance and also aids the interpretation of the fitted model. Putting the pieces together, we reach the following sparse low-rank regularization problem:
\begin{eqnarray} \label{eqn:obj-L1}
\underset{\Bbf}{\text{minimize}} \, \ell(\gammabf, \lambdabf, \Bbf) + \rho \|\vect \Bbf\|_1,
\end{eqnarray}
where $\ell(\gammabf, \lambdabf, \Bbf)$ is the negative log-likelihood function of the GLM with the parameters $\Bcal$ following a symmetric CP structure as in \eqref{eqn:sy-CP-decomp}, and $\| \cdot \|_1$ is the $\ell_1$ norm. The extra symmetry constraint causes the optimization problem in \eqref{eqn:obj-L1} to become much more challenging than that of the original tensor regression in \cite{ZhouLiZhu2013}. In the next section, we propose a highly efficient and scalable optimization algorithm to estimate the regression coefficient tensor $\Bcal$. 

\subsection{Optimization Algorithm}
\label{sec:algorithm}
We divide and conquer the optimization of \eqref{eqn:obj-L1} by alternatively updating $\gammabf$, $\lambdabf$ and $\Bbf$. The key solution is to employ a proximal gradient method to estimate $\Bbf$. 
In the first step, we update $\gammabf$ given $\lambda$ and $\Bbf$. This is simply the classical GLM with offset $\langle {\mathcal B}, \xbf_i \rangle$. After that, we update $\lambdabf$ given $\gammabf$ and $\Bbf$. We note that
\begin{eqnarray*}
\langle {\mathcal B},\xbf_i \rangle 
= \langle \vect \, {\mathcal B}, \vect \, \xbf_i \rangle 
= (\vect \, \xbf_i)\trans (\Bbf \odot \cdots \odot \Bbf) \lambdabf,
\end{eqnarray*}
where $\odot$ denotes the Khatri-Rao product \cite{RaoMitra71GenInv}. Then the problem becomes a GLM with $R$-dimensional covariates $(\vect \, \xbf_i)\trans (\Bbf \odot \cdots \odot \Bbf)$ and offset $\gammabf \trans \zbf_i$. 

Thirdly, we update $\Bbf$ given $\gammabf$ and $\lambdabf$, and toward that end, we employ the proximal gradient method. The gradient descent is based on the first-order approximation to the loss function at the current point $\Bbf^{(t)}$,
\begin{eqnarray*} 
s(\Bbf \mid \Bbf^{(t)}, \delta) 
& = & \ell(\Bbf^{(t)}) + \langle \nabla \ell(\Bbf^{(t)}), \Bbf - \Bbf^{(t)} \rangle + \frac{1}{2 \delta} \|\Bbf - \Bbf^{(t)}\|_{\text{F}}^2 + \rho \|\vect \Bbf\|_1 \\
& = & \frac{1}{2 \delta} \|\Bbf - \{\Bbf^{(t)} - \delta \nabla \ell(\Bbf^{(t)})\}\|_{\text{F}}^2 + \rho \|\vect \Bbf\|_1, 
\end{eqnarray*}
where $\| \cdot \|_{\text F}$ denotes the Frobenius norm, and $\delta$ is a constant determined during the line search. The term $(2 \delta)^{-1} \|\Bbf - \Bbf^{(t)}\|_{\text{F}}^2$ acts as a trust region and shrinks the next iterate towards $\Bbf^{(t)}$. The above surrogate function $s$ is minimized by soft-thresholding $\Bbf^{(t)} - \delta \nabla \ell(\Bbf^{(t)})$ at the threshold value $\rho \delta$, usually with a small number of proximal gradient steps such as five steps. To calculate the gradient $\nabla \ell(\Bbf^{(t)})$, we have 
\begin{eqnarray*}
\nabla \eta (\Bbf) & = & \left[ (\Bbf^{\odot (D-1)} \Lambdabf) \trans \otimes \Ibf_p \right] \left( \sum_{d=1}^D \vect \, \Xbf_{(d)} \right), \\
\nabla \ell (\Bbf) & = & \frac{(y-\mu)\mu'(\eta)}{\sigma^2} \left[ (\Bbf^{\odot (D-1)} \Lambdabf) \trans \otimes \Ibf_p \right] \left( \sum_{d=1}^D \vect \, \Xbf_{(d)} \right),
\end{eqnarray*}
where $\eta = \alpha + \gammabf \trans \Zbf + \langle {\mathcal B}, \Xbf \rangle$, and $\Bbf^{\odot (D-1)}$ denotes Khatri-Rao product of $D-1$ copies of $\Bbf$. To see this result, we first assume a general CP decomposition ${\mathcal B} = \llbracket \lambdabf; \Bbf_1, \ldots, \Bbf_D \rrbracket$. Then we have 
\begin{eqnarray*}
	{\mathcal B}_{(d)} &=& \Bbf_d \Lambdabf [\Bbf_D \odot \cdots \Bbf_{d+1} \odot \Bbf_{d-1} \cdots \odot \Bbf_1]^T,
\end{eqnarray*}
where $\Lambdabf = \diag(\lambdabf)$. Given that $\vect (\Abf_1 \Abf_2 \Abf_3) = (\Abf_3\trans \otimes \Abf_1) \vect \Abf_2$, we have 
\begin{eqnarray*}
	\vect \, {\mathcal B}_{(d)} = [(\Bbf_D \odot \cdots \Bbf_{d+1} \odot \Bbf_{d-1} \cdots \odot \Bbf_1) \Lambdabf \otimes \Ibf_{p_d}] \vect (\Bbf_d).
\end{eqnarray*}
Then $\langle {\mathcal B}, \Xbf \rangle = (\vect \, \Xbf_{(d)}) \trans [(\Bbf_D \odot \cdots \Bbf_{d+1} \odot \Bbf_{d-1} \cdots \odot \Bbf_1) \Lambdabf \otimes \Ibf_{p_d}] \vect (\Bbf_d)$, and $\nabla_{\vect \, \Bbf_d} \eta({\mathcal B}) = [\Lambdabf(\Bbf_D \odot \cdots \Bbf_{d+1} \odot \Bbf_{d-1} \cdots \odot \Bbf_1) \trans \otimes \Ibf_{p_d}] (\vect \, \Xbf_{(d)})$. Now, under the symmetric CP decomposition \eqref{eqn:sy-CP-decomp} with $\Bbf_1 = \cdots = \Bbf_D = \Bbf$, we thus have
\begin{eqnarray*}
	\nabla \eta(\Bbf) = \sum_{d=1}^D \nabla_{\vect \, \Bbf_d} \eta({\mathcal B})
	= \left[ (\Bbf^{\odot (D-1)} \Lambdabf) \trans \otimes \Ibf_p \right] \left( \sum_{d=1}^D \vect \, \Xbf_{(d)} \right).
\end{eqnarray*}
Finally, by using the chain rule,
\begin{eqnarray*}
	\nabla \ell(\Bbf) &=& \frac{(y-\mu) \mu'(\eta)}{\sigma^2} \nabla \eta(\Bbf) \\
	&=& \frac{(y-\mu)\mu'(\eta)}{\sigma^2} \left[ (\Bbf^{\odot (D-1)} \Lambdabf) \trans \otimes \Ibf_p \right] \left( \sum_{d=1}^D \vect \, \Xbf_{(d)} \right).
\end{eqnarray*}

We summarize the above optimization procedure of solving \eqref{eqn:obj-L1} in Algorithm~\ref{algo:sym-tensor-reg}. Lines \ref{algo:update-gamma}-\ref{algo:update-lambda} are for updating regular coefficients $\gammabf$ and tensor scalars $\lambdabf$,  respectively. Lines \ref{algo:update-B-start}-\ref{algo:update-B-end} are the proximal gradient loops for updating $\Bbf$. Line search (lines 9-12) is necessary for each proximal gradient step to guarantee a monotone algorithm. In practice, a small number of proximal gradient steps (such as five steps) suffice for sufficient decrease in objective value. The Nesterov acceleration \cite{BeckTebloulle09FISTA} can be further employed to speed up the proximal gradient steps. The initial step length $\delta_0$ can be determined by a crude estimate of the gradient Lipschitz constant of the GLM model \cite{ZhouLi12MatrixLasso}. Because each block update decreases the objective value, the sequence of objective values converges as long as the objective function is bounded below. We terminate the algorithm when the relative change in objective value is less than $10^{-4}$. Global convergence of the iterates to a stationary point can be established under mild regularity conditions \cite{Tseng01BlockCoordinateDescent}. Due to the lack of convexity, a stationarity point can be a local minimum, a saddle point, or a global minimum. In practice, we use the CP regression outcome to construct an initial point that is already close to the global optimum. This empirical approach is observed to achieve better performance. We introduce the construction method in Section~\ref{sec:initial}.

\begin{algorithm}[t!]
\SetKwInOut{Input}{Input}\SetKwInOut{Output}{Output}
\Input{response $y_i$, regular covariate $\zbf_i$, tensor covariate $\xbf_i$}
\Output{MLE $\widehat \gammabf$, $\widehat \lambdabf$, $\widehat \Bbf$}
Initialize $\gammabf^{(0)}, \lambdabf^{(0)}, \Bbf^{(t)}$ \;
\Repeat{objective value converges}{
${\mathcal B}^{(t)} \gets \llbracket \lambdabf^{(t)}; \Bbf^{(t)}, \ldots, \Bbf^{(t)} \rrbracket$ \;
Update $\gammabf$ by solving a GLM with covariate $\zbf_i$ and offset $\langle {\mathcal B}^{(t)}, \xbf_i \rangle$ \label{algo:update-gamma} \;
Update $\lambdabf$ by solving a GLM with covariate $(\vect \, \xbf_i)\trans \Bbf^{(t) \odot D}$ and offset $\langle \gammabf^{(t+1)}, \zbf_i \rangle$ \label{algo:update-lambda} \;
$\Sbf \gets \Bbf^{(t)}$ \label{algo:update-B-start} \;
\For{$s=1, \ldots,5$}{
$\delta = \delta_0$ \;
\Repeat{$\ell(\Sbf) \le s(\Bbf \mid \Bbf^{(t)}, \delta)$}{
$\Sbf \gets \text{soft-thresholding}(\Sbf - \delta \nabla \ell(\Sbf), \rho \delta)$ \;
$\delta \gets \delta / 2$ \;
}
}
$\Bbf^{(t+1)} \gets \Sbf$ \label{algo:update-B-end} \;
}
\caption{Estimation algorithm of the symmetric tensor regression.}
\label{algo:sym-tensor-reg}
\end{algorithm}

\subsection{Symmetric CP Regression and Estimation}
\label{sec:symcp}
Other than imposing the symmetric constraint on $\mathcal B$ at the modelling stage, we may also consider symmetrize $\Bcal$ in an ad-hoc fashion after we obtain the standard CP regression outcome:
\begin{eqnarray} \label{eqn:symcp}
\Bcal_{\text{CPsym}} = \sum_{r=1}^R \sum_{\pi \in \Pi_D} \betabf_{\pi_1}^{(r)} \circ \cdots \circ \betabf_{\pi_D}^{(r)}/D! 
= \sum_{r=1}^R \sum_{\pi \in \Pi_D} \llbracket \Bbf_{\pi_1},\ldots,\Bbf_{\pi_D} \rrbracket /D!,
\end{eqnarray}
where $\Pi_D$ is the set of all the permutations of $(1,\cdots,D)$. For intuition, notice that when $D=2$, the above formulation (\ref{eqn:symcp}) simply becomes $\Bcal_{\text{CP}} \trans + \Bcal_{\text{CP}} /2$. Here, the symmetry requirement is achieved by the ad-hoc symmetrization step, and the solution also follows a low-rank tensor decomposition. 

We may take a step further and adopt the decomposition in (\ref{eqn:symcp}) to the GLM framework. It leads to another sparse low rank regression problem that more resembles the standard CP regression, which has the same number of parameters as CP regression (in order of $pRD$), and thus can be viewed as a relaxation of the symmetric tensor relaxation.
% \begin{eqnarray} \label{eqn:,cp-obj-L1}
% \min \ell(\gammabf, \lambdabf, \Bbf_{1}, \cdots, \Bbf_{D}) + \rho \sum_{d=1}^D\|\vect \Bbf_{d}\|_1.
% \end{eqnarray}

For the parameter estimation, note that when $\Xbf$ is symmetric, we have $\Xbf_{i_1,i_2,\cdots,i_D} = \Xbf_{\pi_1,\cdots,\pi_D}$ for any permutation $\pi$ of $(i_1,\cdots, i_D)$, as well as $\Xbf_{(1)} = \cdots = \Xbf_{(D)}$. As a consequence, we can expect:
$$
\langle \Xbf, \betabf_1 \circ \cdots \circ \betabf_D \rangle = \langle \Xbf, \betabf_{\pi_1} \circ \cdots \circ \betabf_{\pi_D} \rangle,
$$
since the subscript index of $\Xbf$ can be randomly permuted without changing the corresponding element.
As a consequence, for the systematic part $\langle \Xbf,\Bcal \rangle$ in GLM framework, we have:
\begin{eqnarray*}
\langle \Xbf, \Bcal_{\text{CPsym}} \rangle &=& \big \langle \Xbf,\sum_{r=1}^R \sum_{\pi \in \Pi_D} \betabf_{\pi_1}^{(r)} \circ \cdots \circ \betabf_{\pi_D}^{(r)}/D! \big\rangle \\
&=&  \big \langle \Xbf, \sum_{r=1}^R \betabf^{(r)}_1 \circ \cdots \circ \betabf^{(r)}_D \big \rangle \\
 &=& \big\langle \Bbf_{d}, \Xbf_{(d)}(\Bbf_D \odot \cdots \odot \Bbf_{d+1}, \odot \Bbf_{d-1}, \odot \Bbf_{1}) \big \rangle,\text{ for } d=1,\cdots, D,
\end{eqnarray*}
which is exactly the systematic part of the standard CP regression. We thus establish the equivalence between the symmetrized and standard CP regression in the systematic part of GLM. On the other hand, it is obvious that the $\ell_1$ regularization under the new parameterization of $\Bcal_{\text{CPsym}}$ is also equivalent to that of the standard CP regression: we first add the $D!$ permutation terms (which all consists of the same $\Bbf_1, \ldots, \Bbf_D$) and then divide the total sum by the same factor of $D!$. 
Therefore, we can simply adopt the same update rules from the standard CP regression, by employing the block update algorithm proposed in \cite{ZhouLiZhu2013}, to first compute the non-symmetry solution ${\Bcal}_{\text{CP}}$ in each step, and then symmetrize it as in (\ref{eqn:symcp}). We do not explore the symmetrization for the Tucker tensor regression, because the core tensor causes extra complications for the modelling and interpretation.

The key purpose of introducing the symmetric CP regression is to construct an initial point for Algorithm \ref{algo:sym-tensor-reg}. The objective function for the symmetric tensor regression is non-convex in general, while the CP regression enjoys a coordinate-wise convexity. Therefore, we effectively leverage the relaxation of our problem (which is the symmetric CP regression in this case) to construct a reasonable initial value rather than using the random or all-zero initializations.

\subsection{Constructing Initial Value for Symmetric Tensor Regression}
\label{sec:initial}

The non-convexity of the objective function often brings uncertainty and instability to our optimization. To recover the true signal tensor, we need to accurately estimate both $\lambdabf$ and $\Bbf$ (suggested by \eqref{eqn:sy-CP-decomp}).
We use the following toy example to illustrate on several potential issues caused by the non-convexity of our problem, when the starting point is not well-initialized.

\begin{exmp}
We first generate the true tensor signal as follow:
\[
\Bcal_0 =
  \begin{bmatrix}
    0 & 1 \\
    1 & 0
  \end{bmatrix}
  =
  1 \times \begin{bmatrix}
    0.707 \\
    0.707
  \end{bmatrix}
  \times
  \begin{bmatrix}
    0.707 & 0.707 
  \end{bmatrix}
  + (-1) \times
  \begin{bmatrix}
    -0.707 \\
    0.707
  \end{bmatrix}
  \times
  \begin{bmatrix}
    -0.707 & 0.707 
  \end{bmatrix}
\] 
and generate the response according to $Y = \langle \Xbf, \Bcal_0 \rangle + \epsilon$, where $\Xbf$ is a randomly generated correlation matrix and $\epsilon \sim N(0,\sigma^2)$, and the sample size is $n=1000$. We do not consider the scalar covariates in this case. We choose $\sigma$ such that the signal-to-noise ration is 10:1. We generate a dataset with $n=1000$ samples. Note that $\Bcal_0$ has a simple rank-2 decomposition, with $\betabf^{(1)} = (0.707, 0.707)$, $\betabf^{(2)} = (-0.707, 0.707)$ and $\lambdabf=(1,-1)$. 

However, if we set the initial value for $\betabf^{(1)}$ and $\betabf^{(2)}$ as $(1,0)$ and $(0,0)$, the symmetric tensor regression under Algorithm \ref{algo:sym-tensor-reg} always converges to the following local stationary point
% \[
% \hat{\Bcal} =
%   \begin{bmatrix}
%     -0.454 & 0.463 \\
%     0.463 & -0.471
%   \end{bmatrix}
%   =
%   1 \times \begin{bmatrix}
%     -0.674 \\
%     0.686
%   \end{bmatrix}
%   \times
%   \begin{bmatrix}
%     -0.674 & 0.686 
%   \end{bmatrix}
%   + 0
% \] 
where $\hat{\betabf}^{(1)} = (-0.674, 0.686)$, $\hat{\betabf}^{(2)} = (0,0)$ and $\hat{\lambdabf} = (1,0)$. By checking the optimization dynamics, we find that with this initialization, $\lambda_2$ never escapes zero during the updates and the second rank is always voided. By conducting extensive simulations, we recognize that when $\hat{\lambdabf}$ is identified with the wrong signs (even partially) at initialization, the algorithm will be stuck in the local optimum where certain ranks are voided.
\end{exmp}

\vspace{0.1cm}

\begin{exmp}
We now consider a slightly different $\Bcal_0$:
\[
\Bcal_0 =
  \begin{bmatrix}
    1 & 1 \\
    1 & 0
  \end{bmatrix}
  =
  1 \times \begin{bmatrix}
    1.376 \\
    0.851
  \end{bmatrix}
  \times
  \begin{bmatrix}
    1.376 & 0.851 
  \end{bmatrix}
  + (-1) \times
  \begin{bmatrix}
    0.851 \\
    -1.376
  \end{bmatrix}
  \times
  \begin{bmatrix}
    0.851 & -1.376 
  \end{bmatrix}
\] 
and we generate the dataset described in the previous example. If we choose the initial value for $\betabf^{(1)}$ and $\betabf^{(2)}$ as $(0,50)$ and $(1,1)$, the symmetric tensor regression converges to the local stationary point
% \[
% \hat{\Bcal} =
%   \begin{bmatrix}
%     0.662 & 1.124 \\
%     1.124 & -0.042
%   \end{bmatrix}
%   =
%   1 \times \begin{bmatrix}
%     1.767 \\
%     5.696
%   \end{bmatrix}
%   \times
%   \begin{bmatrix}
%     1.767 & 5.696 
%   \end{bmatrix}
%   + (-1) \times 
%   \begin{bmatrix}
%     5.699 \\
%     1.568
%   \end{bmatrix}
%   \times
%   \begin{bmatrix}
%     5.699 & 1.568 
%   \end{bmatrix},
% \] 
where $\hat{\betabf}^{(1)} = (1.767, 5.696)$, $\hat{\betabf}^{(2)} = (5.699, 1.568 )$ and $\hat{\lambdabf} = (1,-1)$. In this example, even though $\lambdabf$ is correctly identified, the initial values are so distant from the global optimum that $\hat{\betabf}^{(1)}$ and $\hat{\betabf}^{(2)}$ converges to the local stationary points as well.
\end{exmp}

The above two examples emphasize the importance of initial values for our problem. It is not our focus here to discuss non-convex optimization techniques for tensor regression, however, we observe empirically that the symmetric CP regression introduced in Section~\ref{sec:symcp} is more stable with respect to initialization and often recovers the true signal better. Nevertheless, it has $pR(D-1)$ extra parameters than the proposed symmetric tensor regression. Therefore, we first consider the low-rank \emph{symmetric} decomposition to approximate the estimated $\hat{\Bcal}_{\text{CPsym}}$: $\hat{\Bcal}_{\text{CPsym}} \approx \llbracket  \lambdabf_{\text{init}},\underbrace{ \Bbf_{\text{init}},\ldots,\Bbf_{\text{init}} \rrbracket}_{R}$, and use it as the initial values. Toward this end, 
we need to solve the following optimization problem:

\begin{equation}\label{eqn:sym-initial}
\Bbf_{\text{init}} = \arg\min_{\beta_1,\ldots, \beta_R}\big\|\hat{\Bcal}_{\text{CPsym}} - \sum_{r=1}^R \lambda_r \cdot \underbrace{\beta_r \circ \ldots \circ \beta_r}_{D} \big\|_F^2,
\end{equation}

where we use $\|\cdot\|_F$ to denote the Frobenius norm. When $D=2$, the problem is equivalent to finding the eigen-decomposition of $\hat{\Bcal}_{\text{CPsym}}$. 
When $D>2$, it is also referred to as the higher-order eigen-decomposition where several methods have been proposed to effectively find approximations to the solution \cite{nie2017low,batselier2016symmetric}. We do not provide in-depth discussions on those work and we focus on the $D=2$ case.

\begin{exmp}
Using the same simulation setup from the previous examples, we now let $\Bcal_0$ be given by \texttt{two box}, \texttt{three box} and \texttt{cross}, which are all symmetric signals (first row of Figure~\ref{fig:random-vs-const}). We start by directly applying Algorithm \ref{algo:sym-tensor-reg} with $R=3$ under random initialization, and provided the estimated $\hat{\Bcal}$ in the second row of Figure~\ref{fig:random-vs-const}. We then use constructed initial values for Algorithm \ref{algo:sym-tensor-reg}, and provide the results in the third row of Figure~\ref{fig:random-vs-const}. 

\begin{figure}[htb]
\caption{Using the random initial values and constructed initial values described in Example 3.3 for Algorithm \ref{algo:sym-tensor-reg}, with $n=1000$. The results are averaged over ten random initialization.}
\label{fig:random-vs-const}
\begin{center}
\begin{tabular}{m{.15\textwidth} m{.17\textwidth} m{.17\textwidth} m{.17\textwidth}}
$\Bcal_0$& \includegraphics[scale = 0.13]{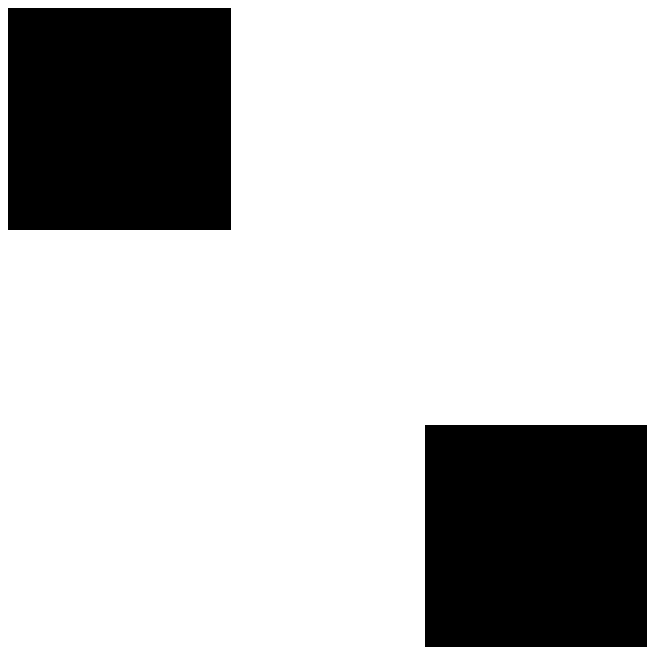} & \includegraphics[scale = 0.13]{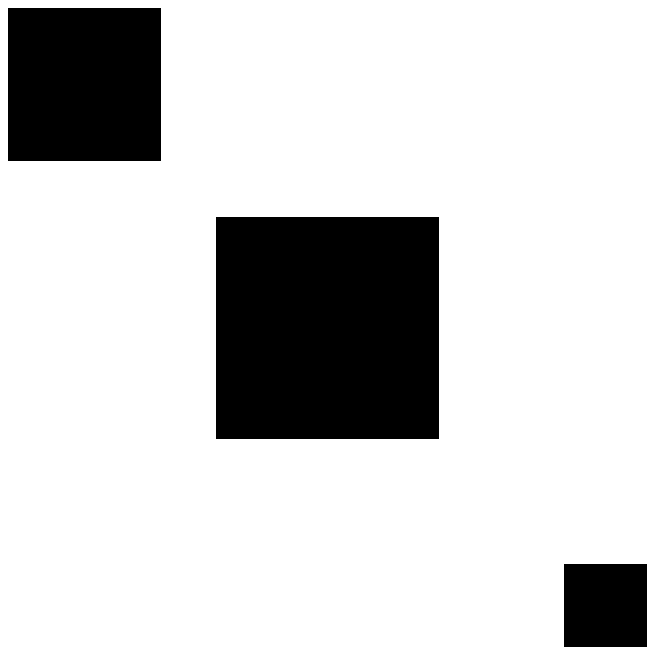} & \includegraphics[scale = 0.13]{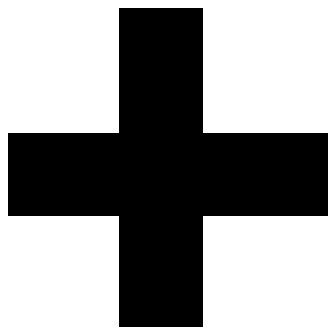} \\
$\hat{\Bcal}_{\text{random initial value}}$ & \includegraphics[scale = 0.13]{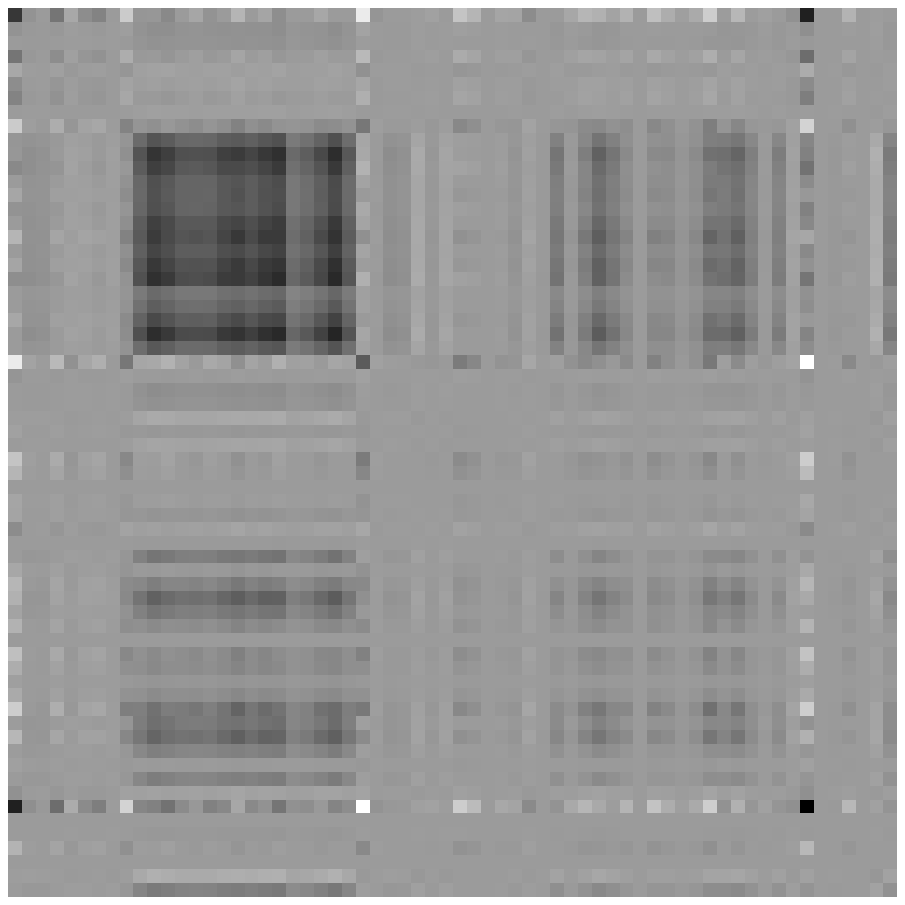} & \includegraphics[scale = 0.13]{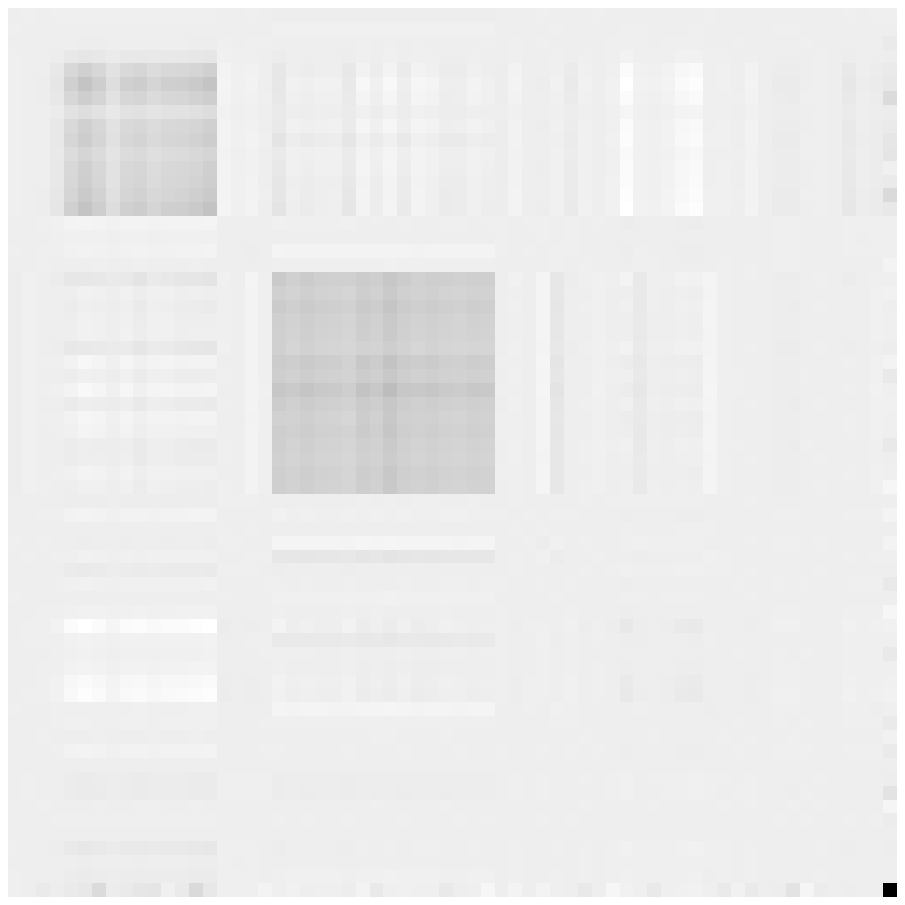} & \includegraphics[scale = 0.13]{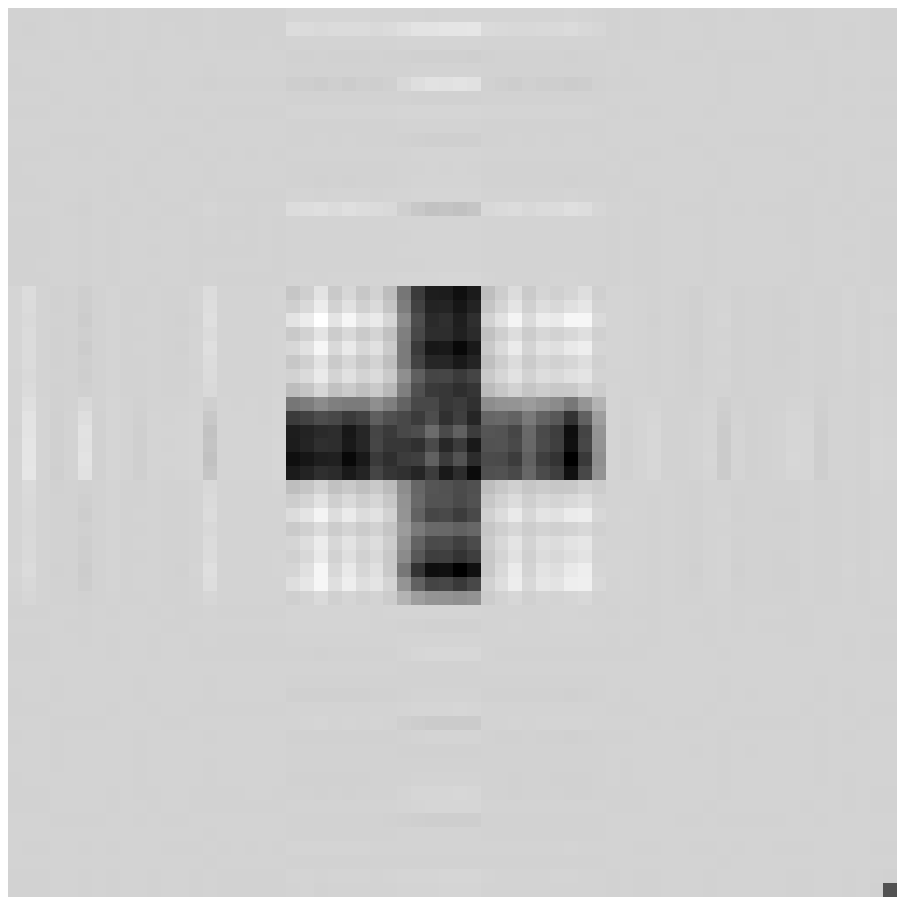} \\
$\hat{\Bcal}_{\text{constructed initial value}}$ & \includegraphics[scale = 0.13]{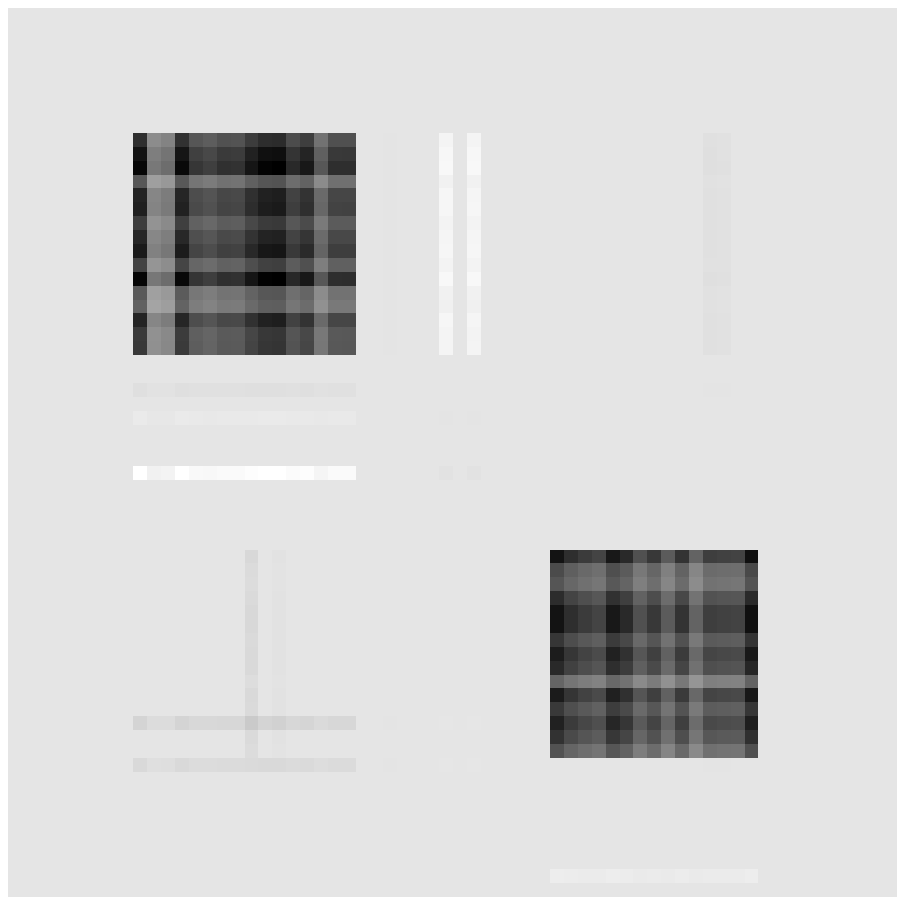} & \includegraphics[scale = 0.13]{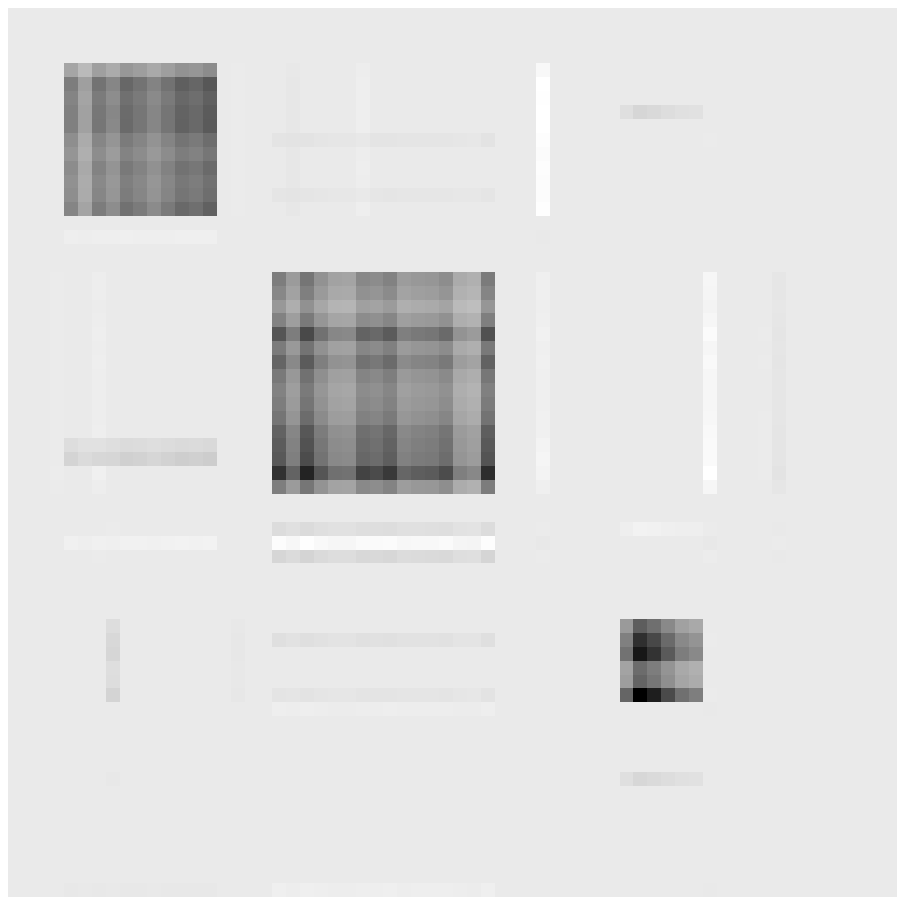} & \includegraphics[scale = 0.13]{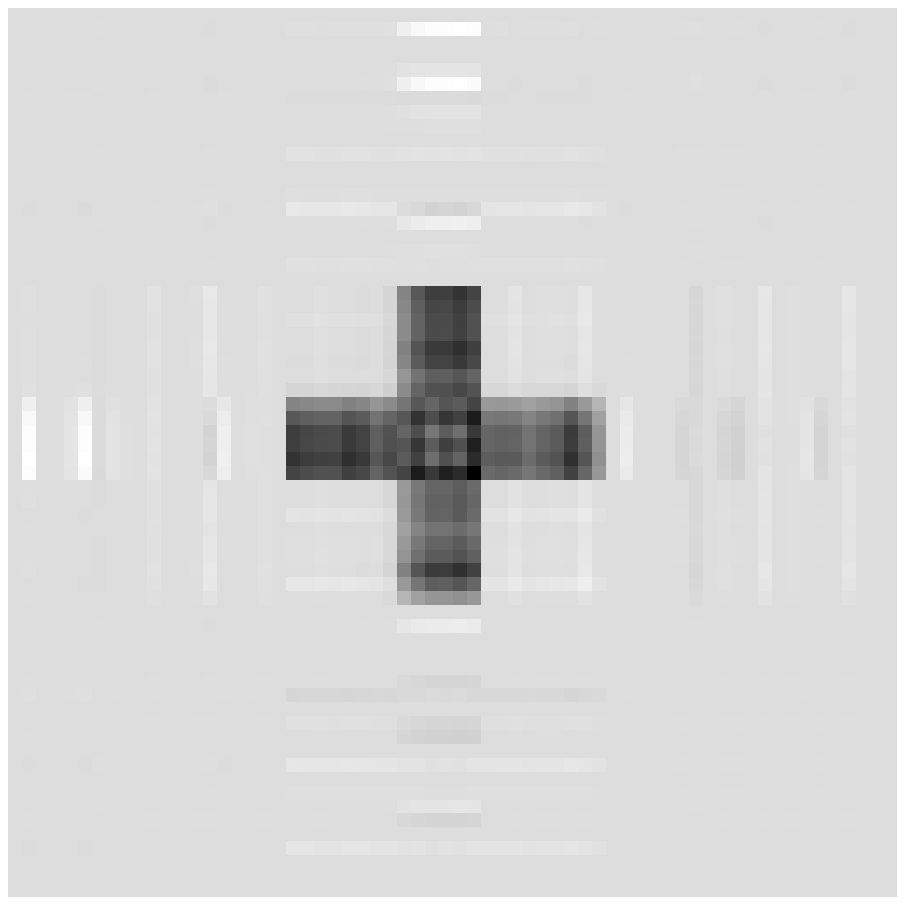} \\
\end{tabular}
\end{center}
\end{figure}
\end{exmp}
From this example, we see that the constructed initial values improves the performance of Algorithm \ref{algo:sym-tensor-reg} significantly. In the following simulation studies and real data analysis, we use the constructed initial value for the proposed symmetric tensor regression.

\section{Numerical Results}
\label{sec:results}

\subsection{Simulations}
\label{sec:simulations}

\begin{figure}
\begin{center}
\label{fig:sim-shape0}
\begin{tabular}{ccccc}
$\Bcal$ & n=500 & n=700 & n=900 & n=1100 \\ \hline
\includegraphics[scale=0.11]{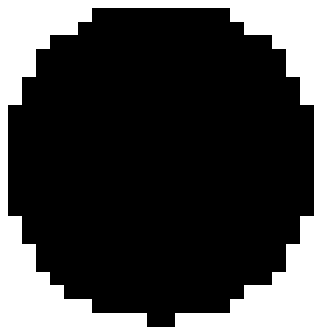} & \includegraphics[scale=0.11]{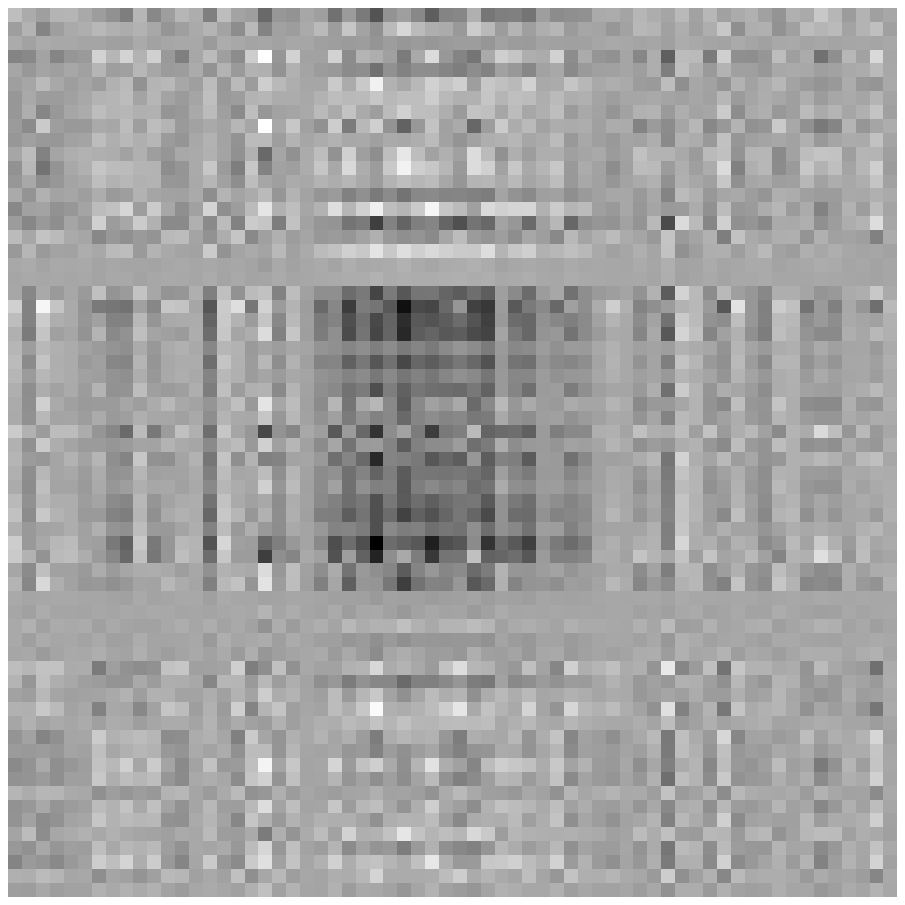} & \includegraphics[scale=0.11]{./fig-ball_500_TR1.eps}  & \includegraphics[scale=0.11]{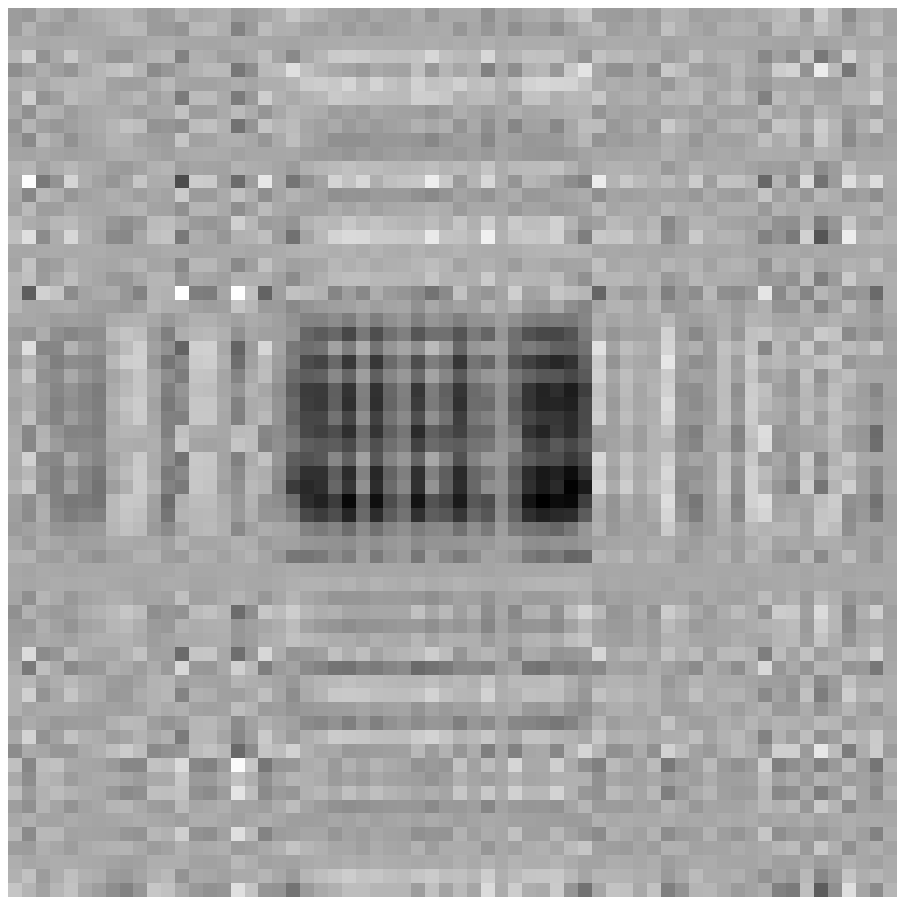} & \includegraphics[scale=0.11]{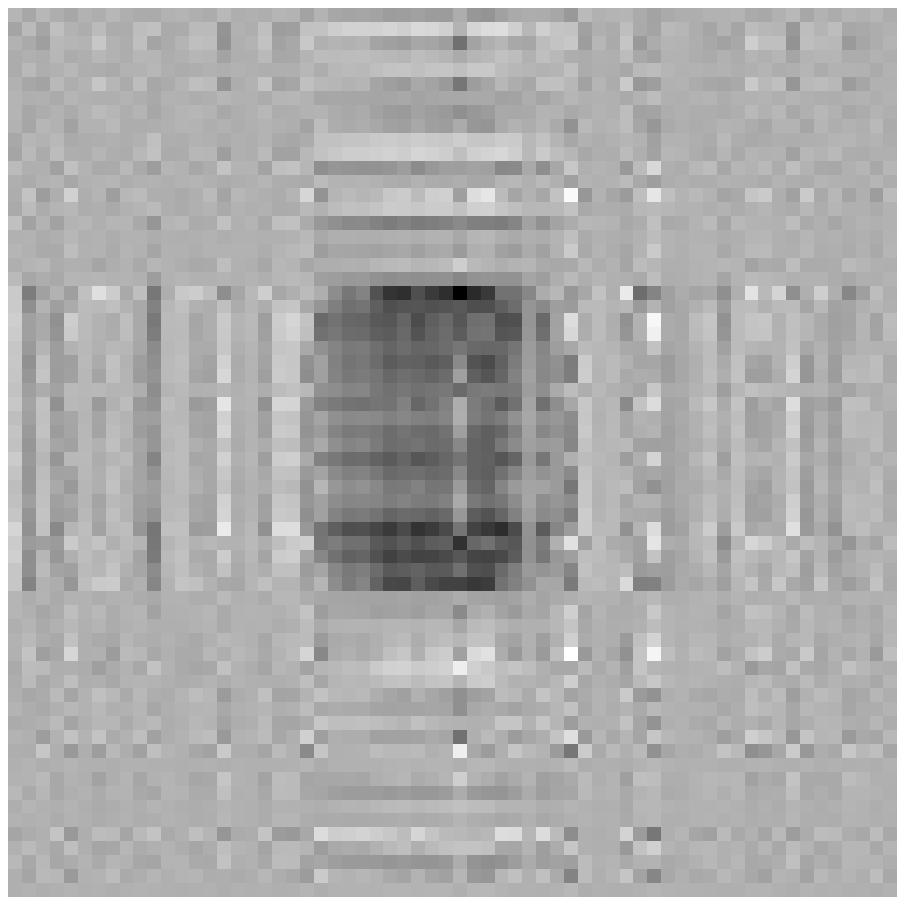} \\

&  \includegraphics[scale=0.11]{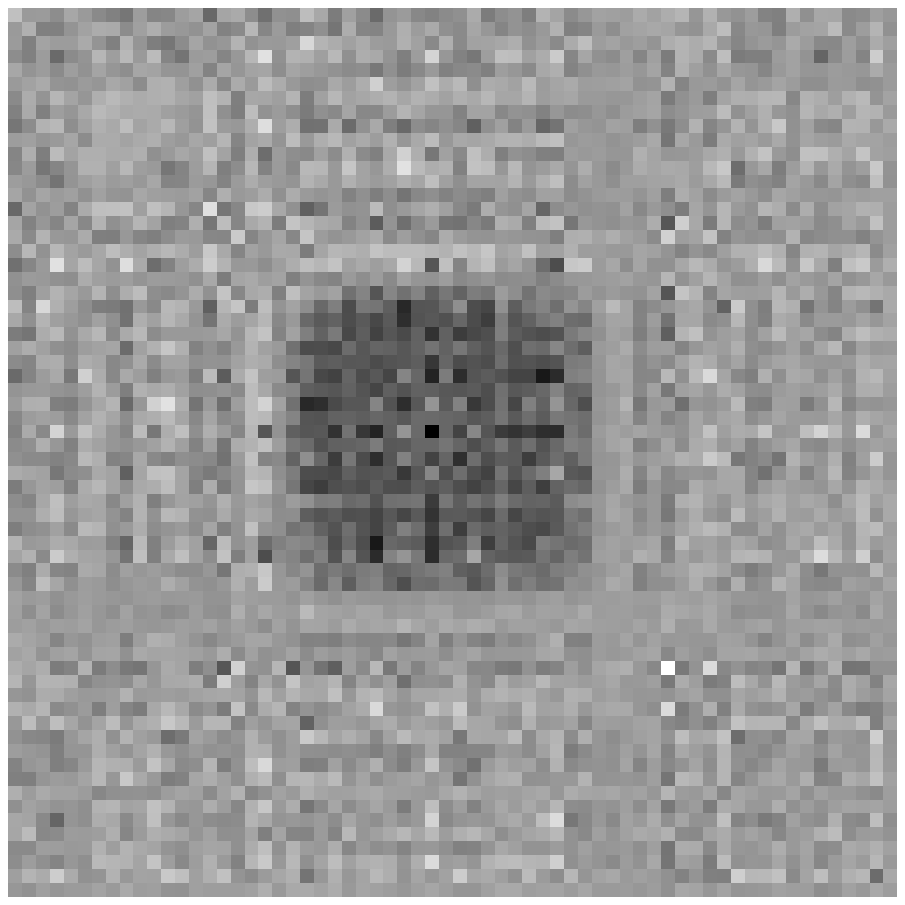} & \includegraphics[scale=0.11]{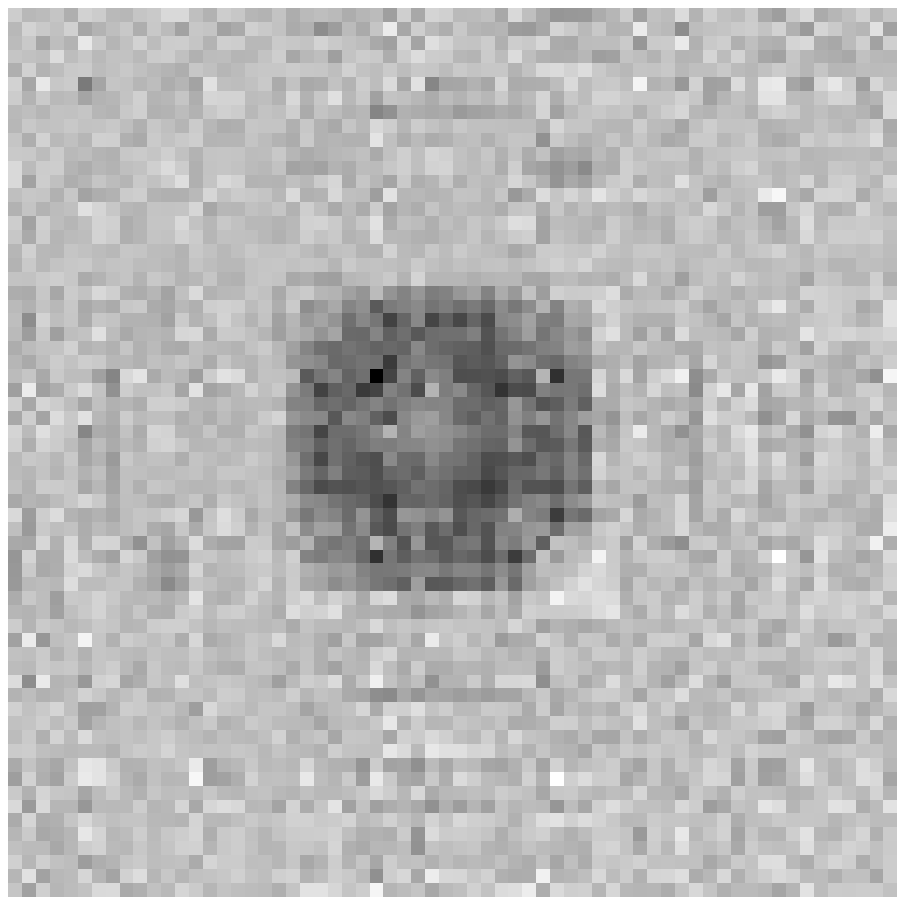}  & \includegraphics[scale=0.11]{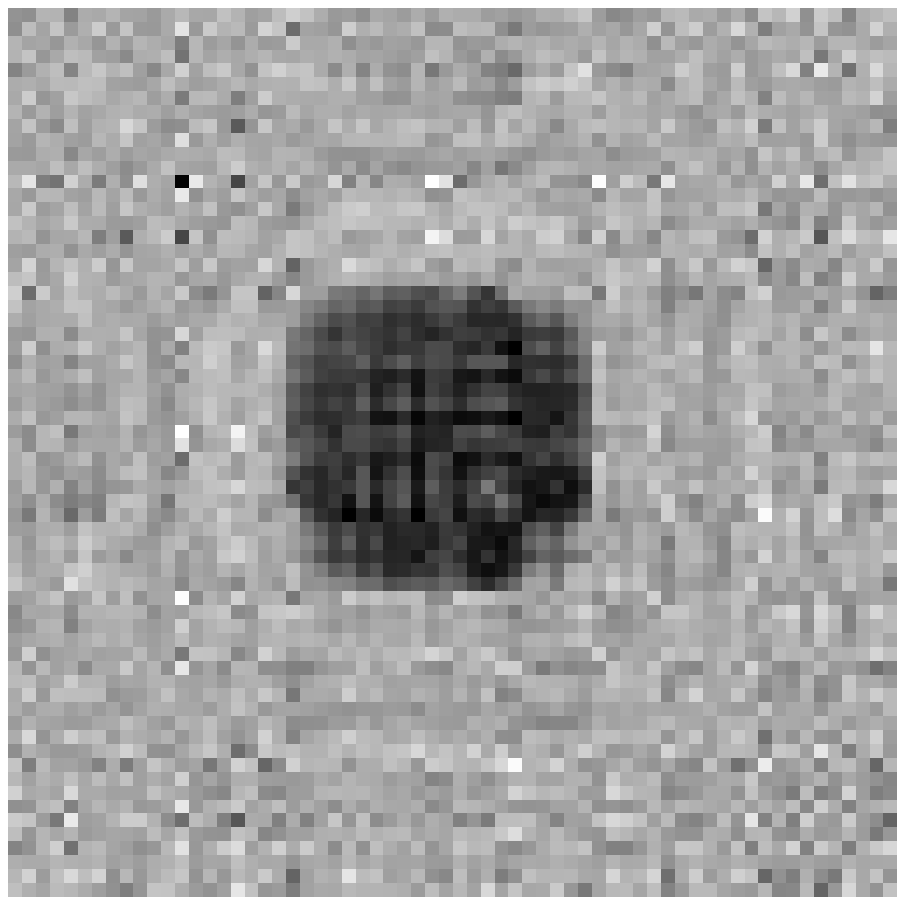} & \includegraphics[scale=0.11]{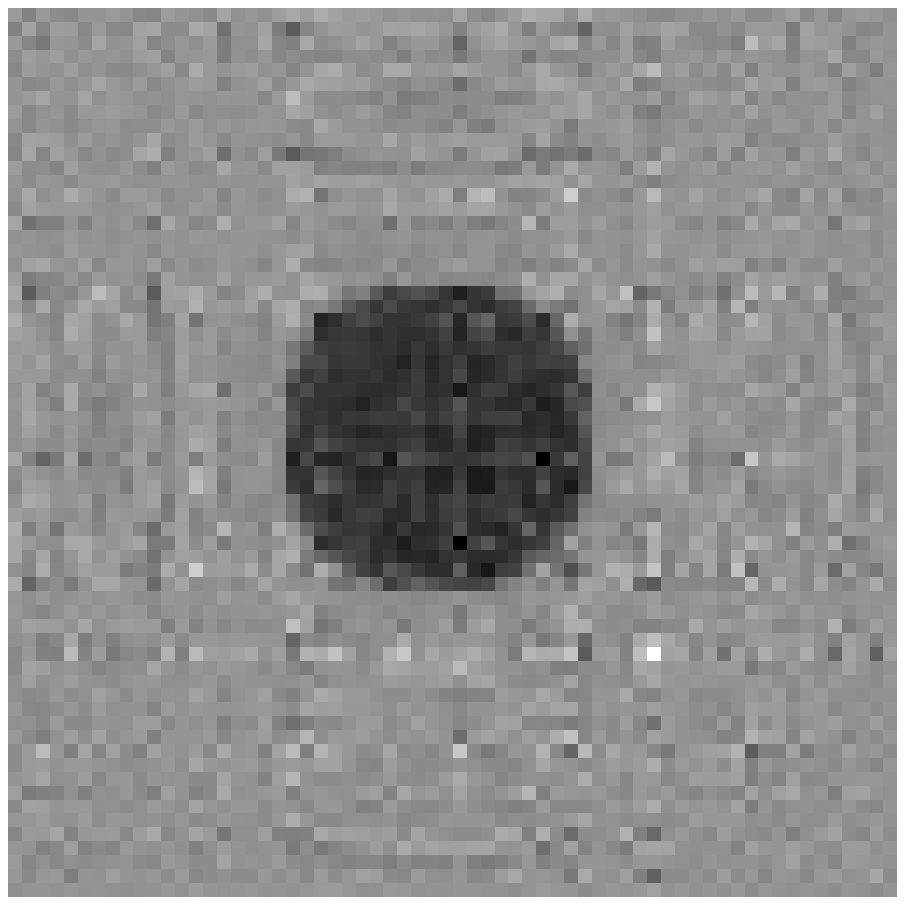} \\

&  \includegraphics[scale=0.11]{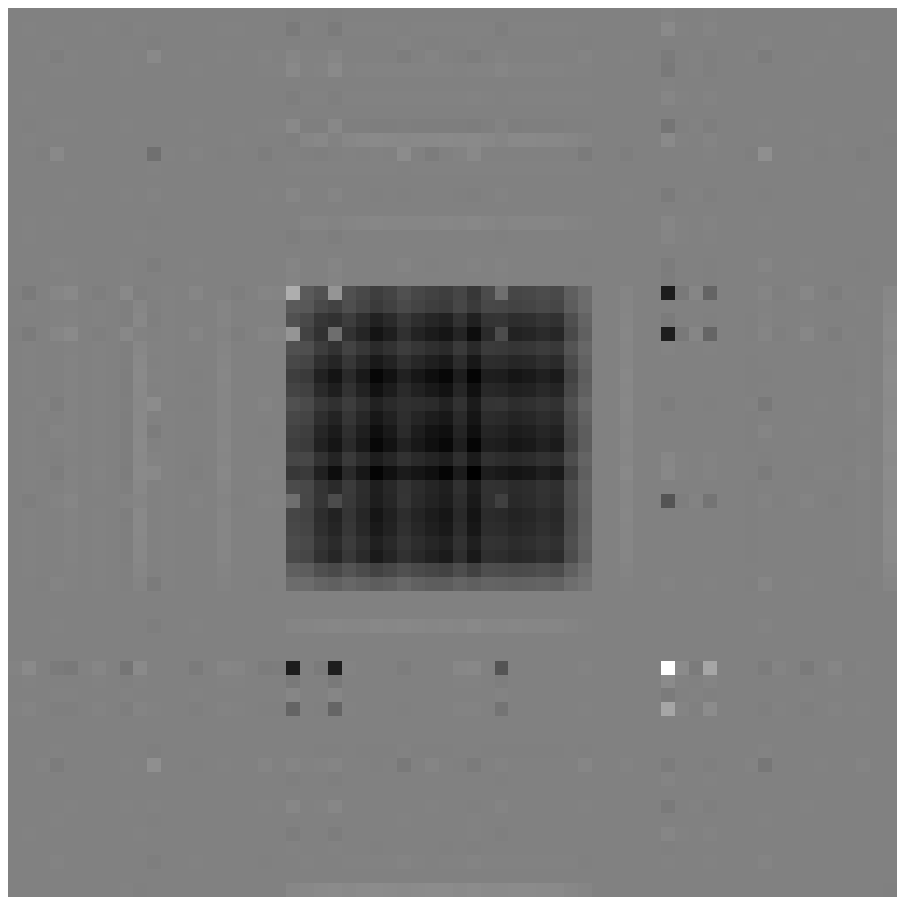} & \includegraphics[scale=0.11]{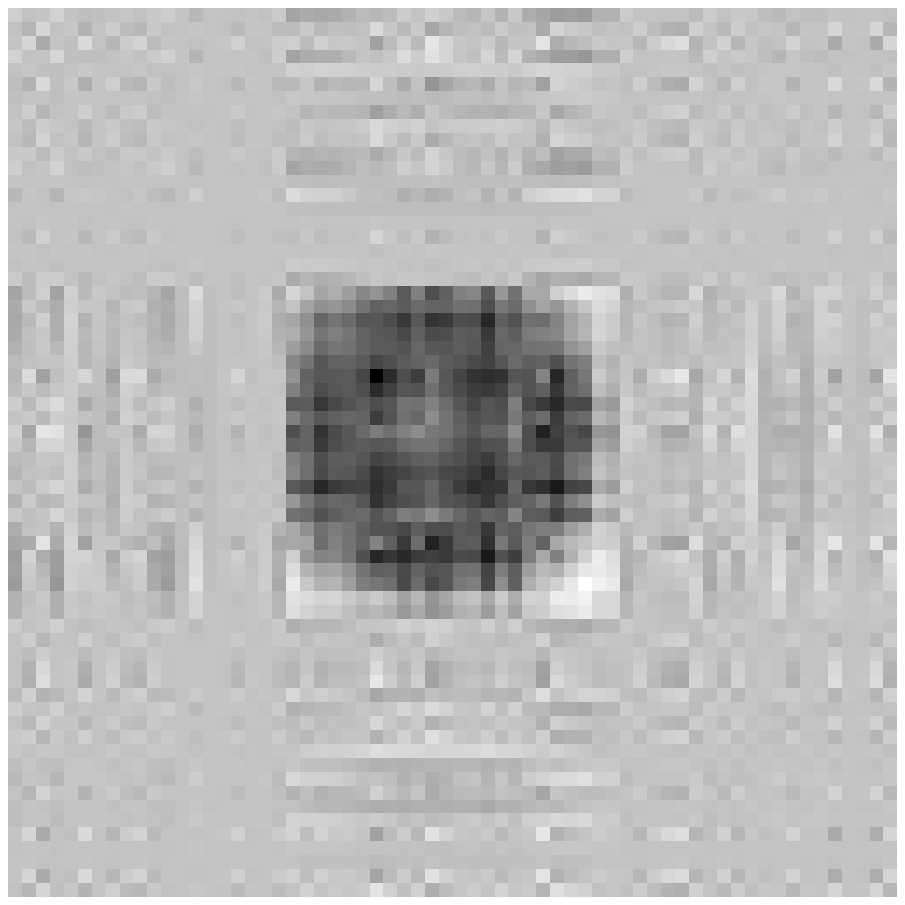}  & \includegraphics[scale=0.11]{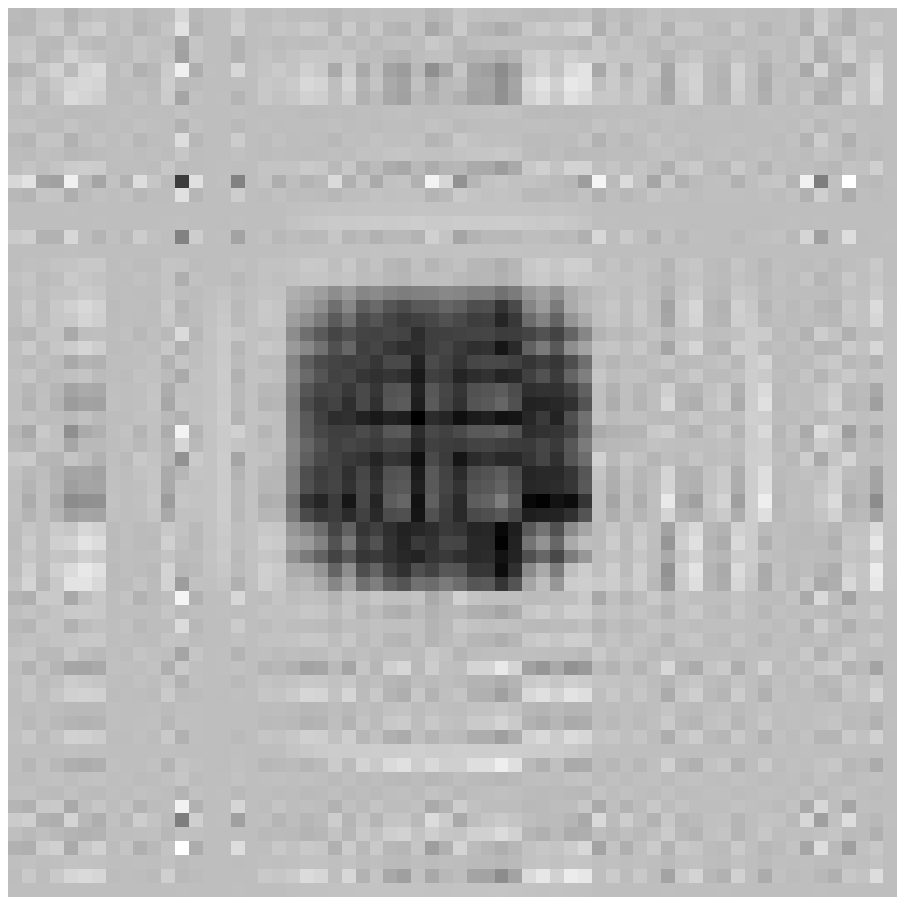} & \includegraphics[scale=0.11]{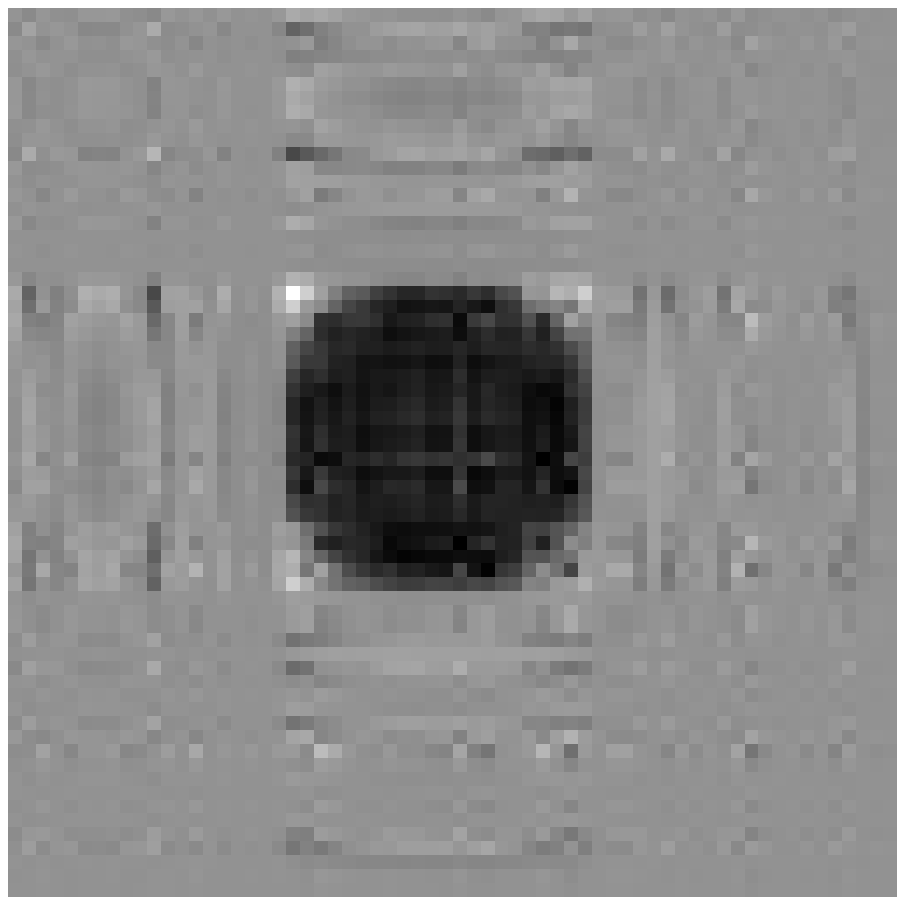} \\ \hline

\includegraphics[scale=0.11]{./fig-cross.eps} & \includegraphics[scale=0.11]{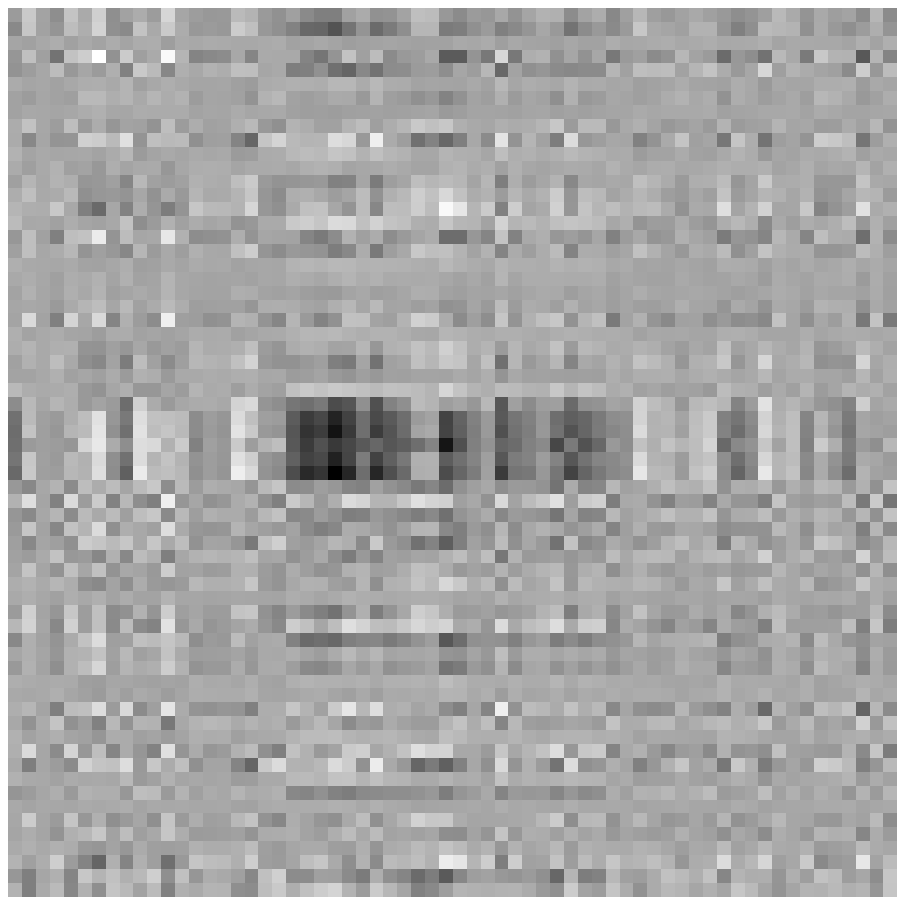} & \includegraphics[scale=0.11]{./fig-cross_500_TR1.eps}  & \includegraphics[scale=0.11]{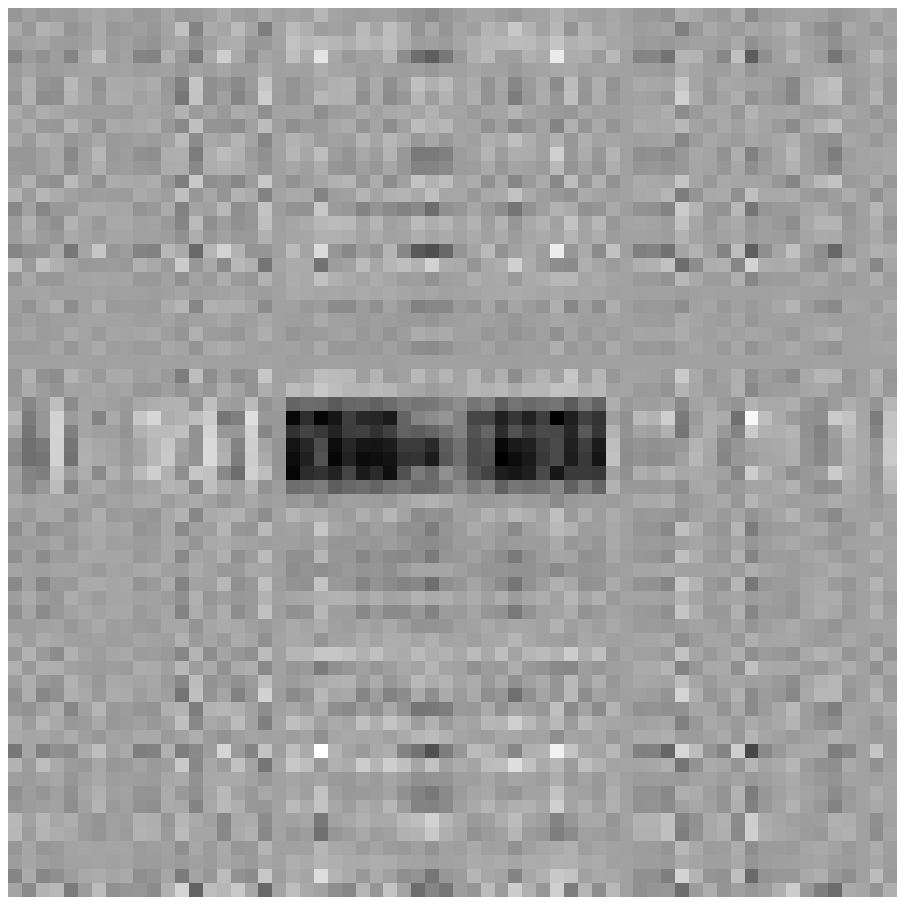} & \includegraphics[scale=0.11]{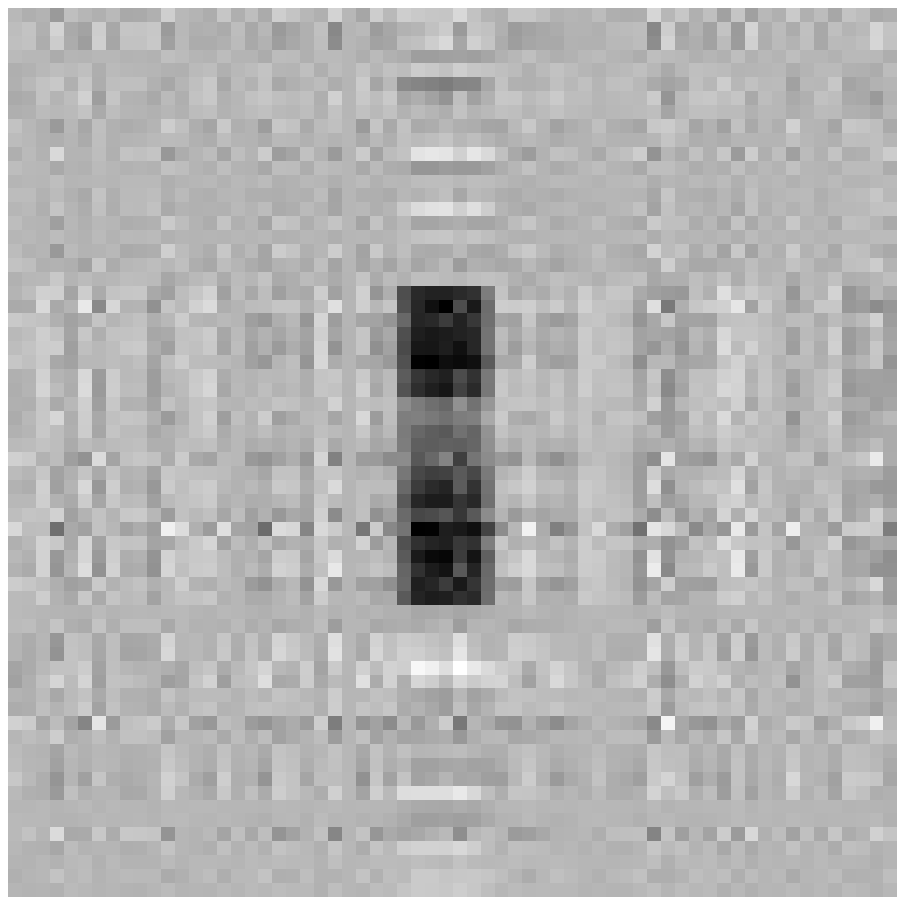} \\

&  \includegraphics[scale=0.11]{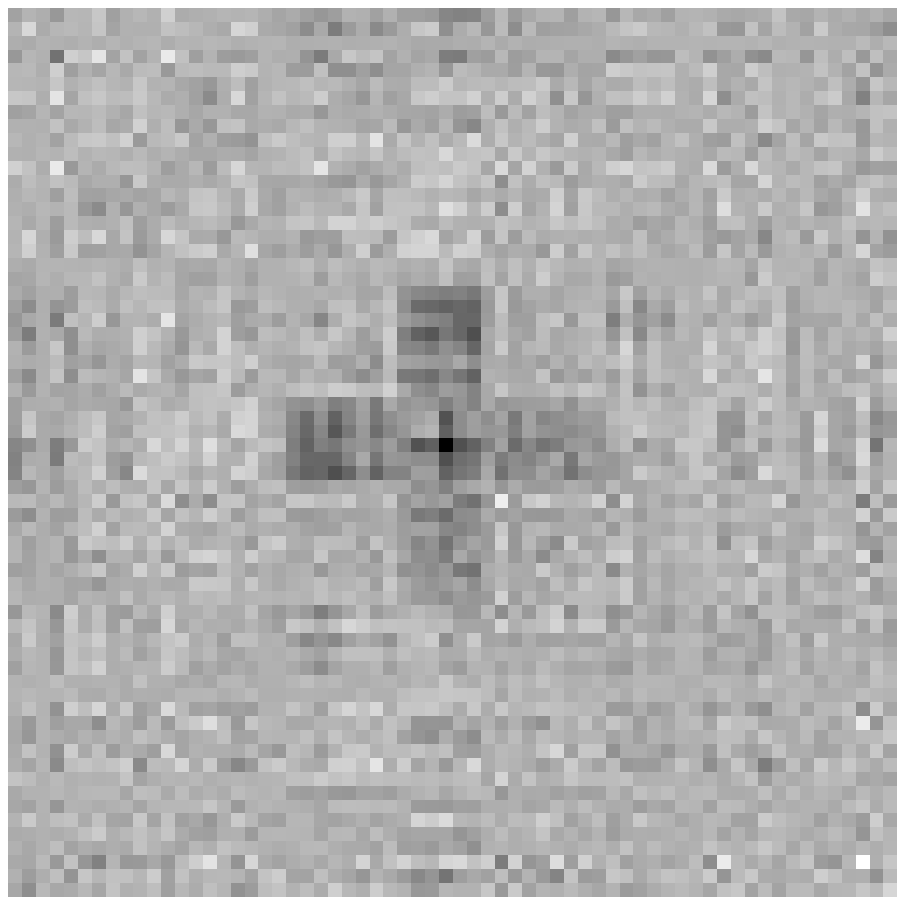} & \includegraphics[scale=0.11]{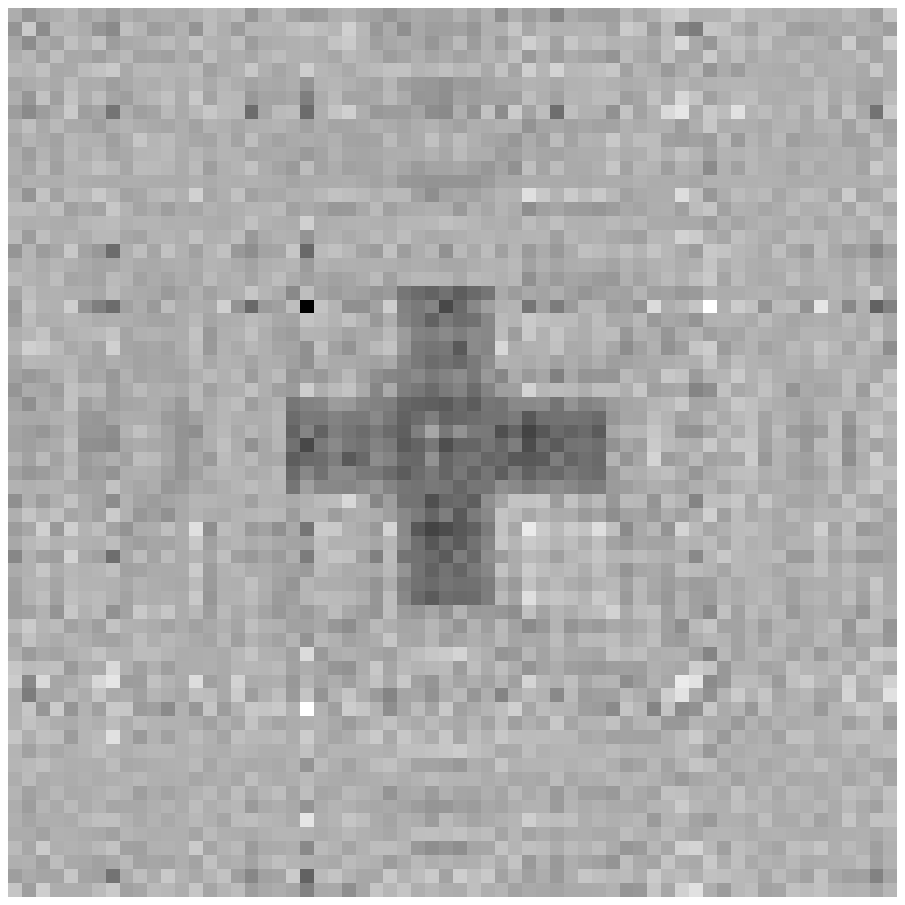}  & \includegraphics[scale=0.11]{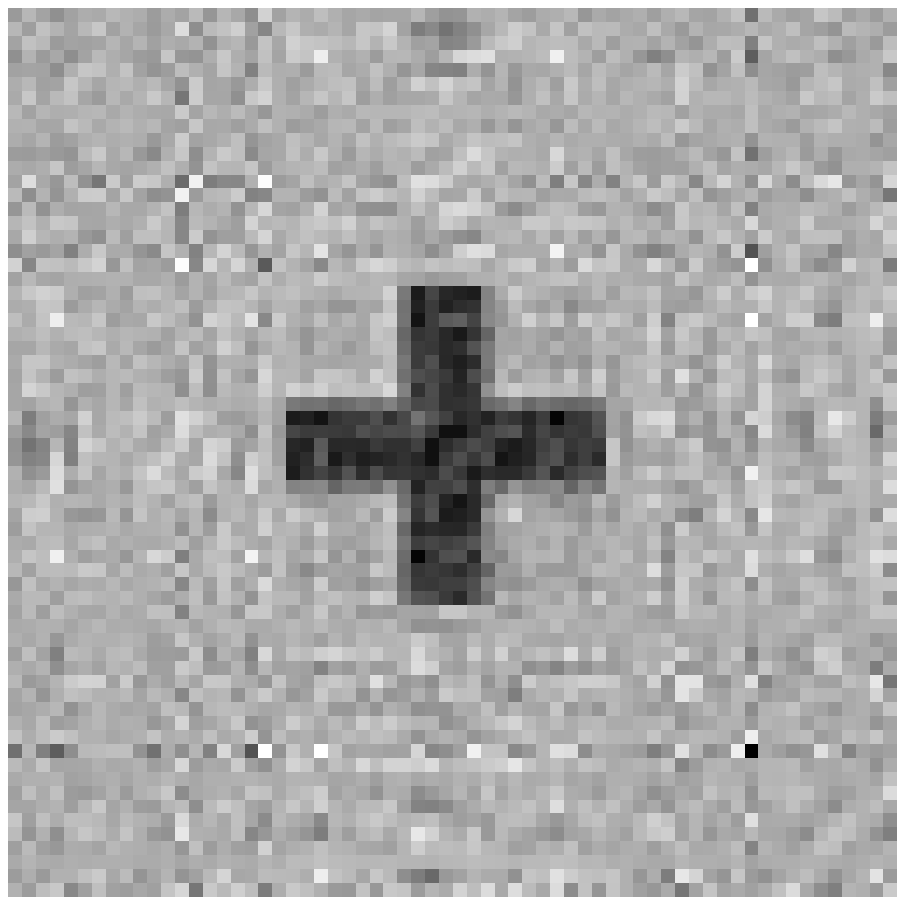} & \includegraphics[scale=0.11]{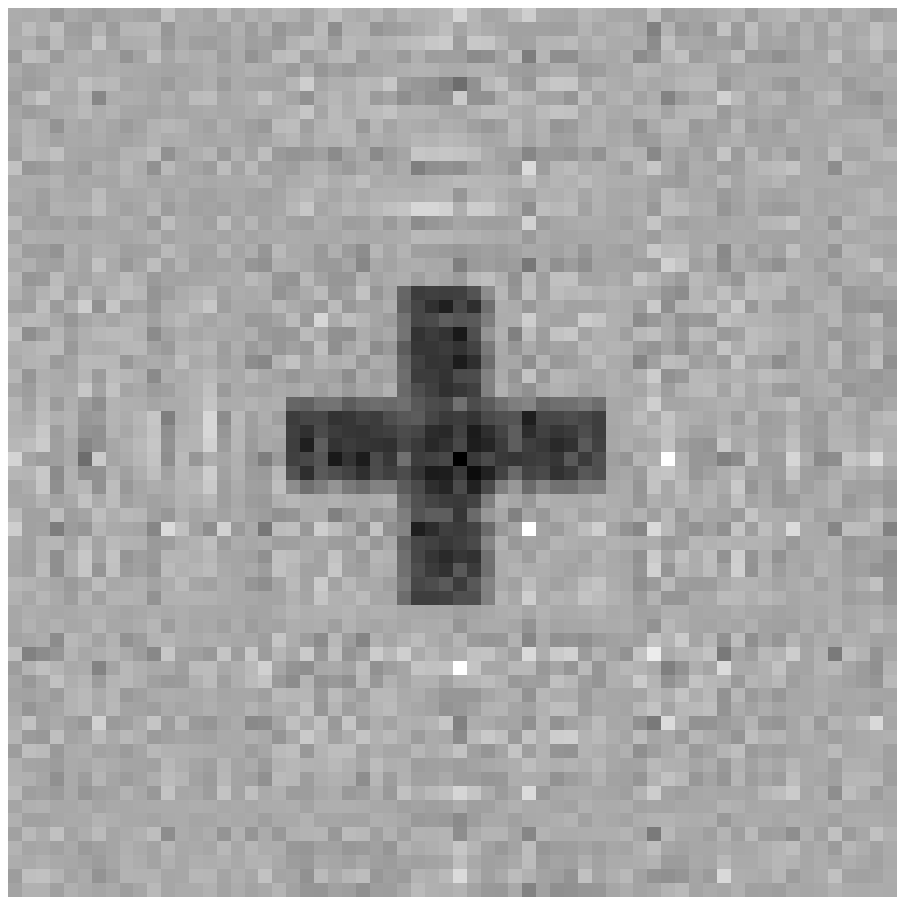} \\

&  \includegraphics[scale=0.11]{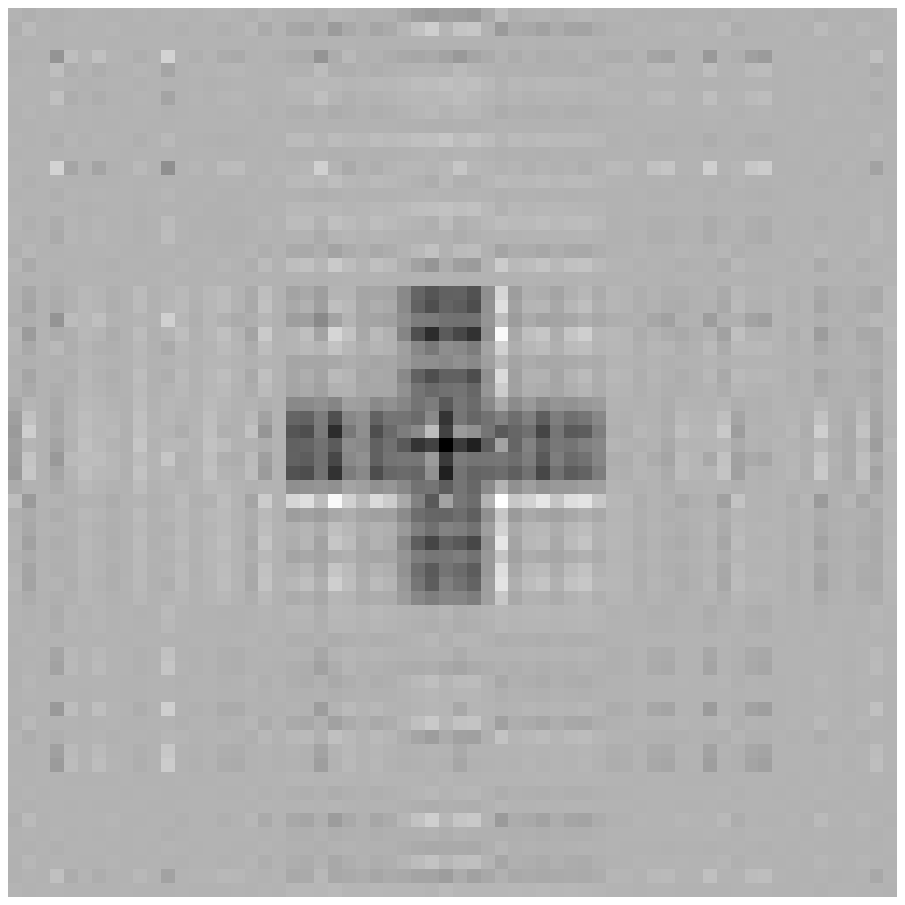} & \includegraphics[scale=0.11]{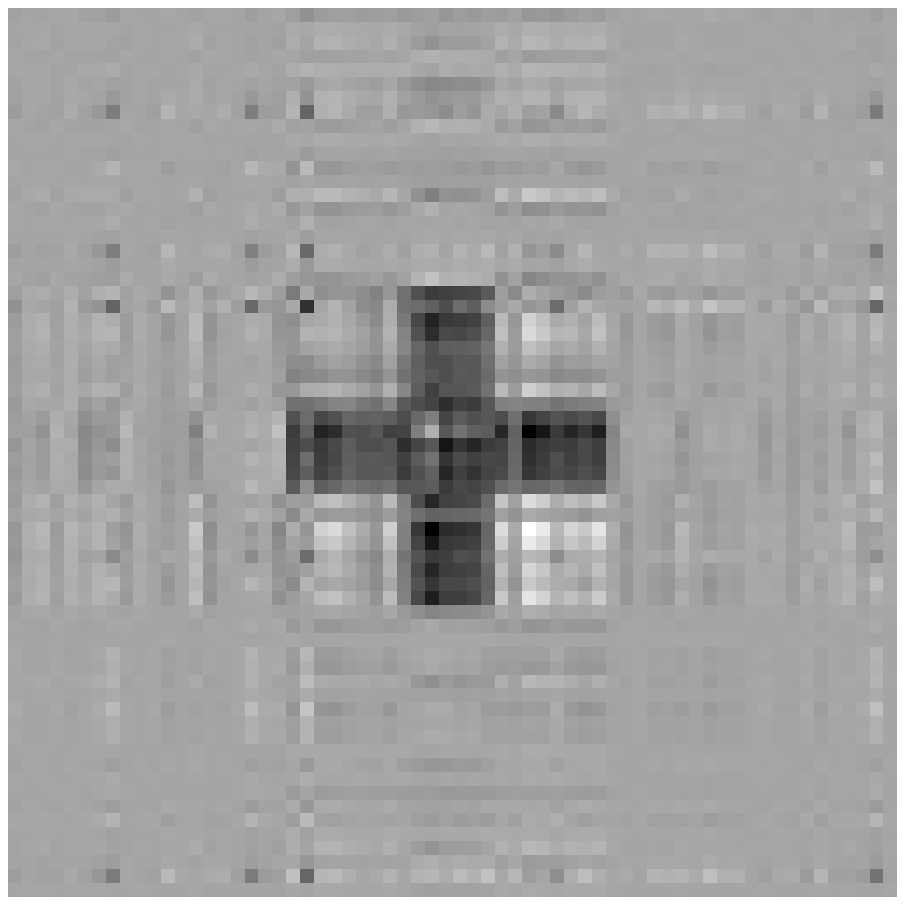}  & \includegraphics[scale=0.11]{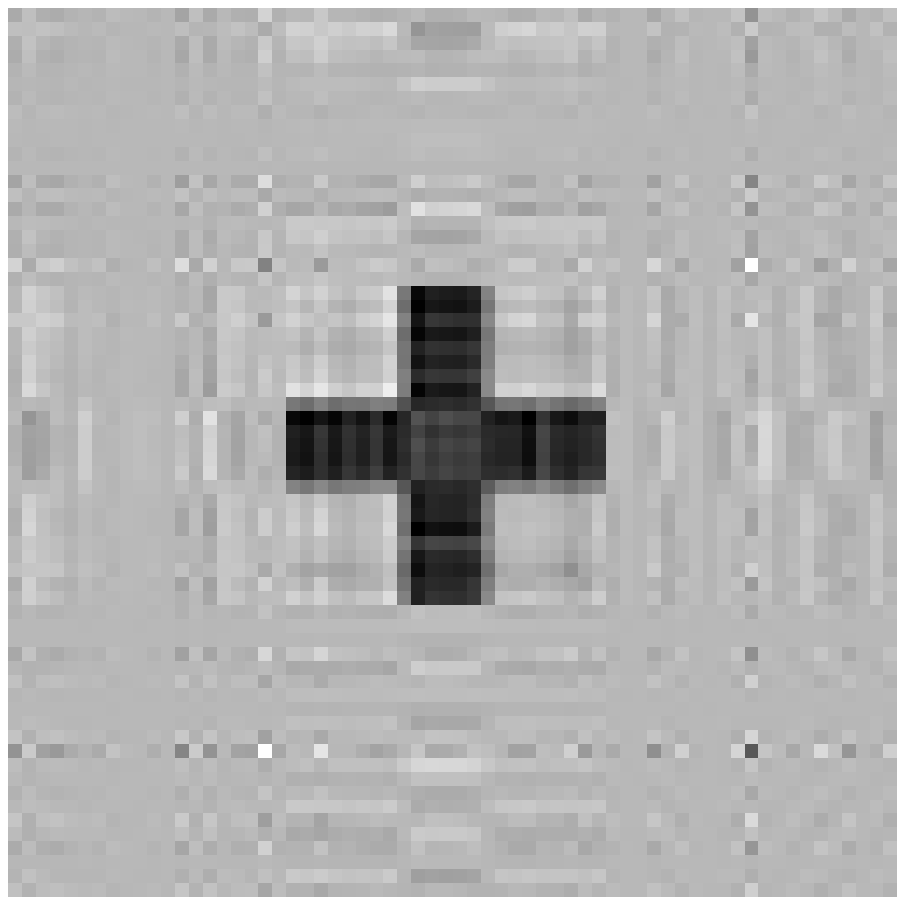} & \includegraphics[scale=0.11]{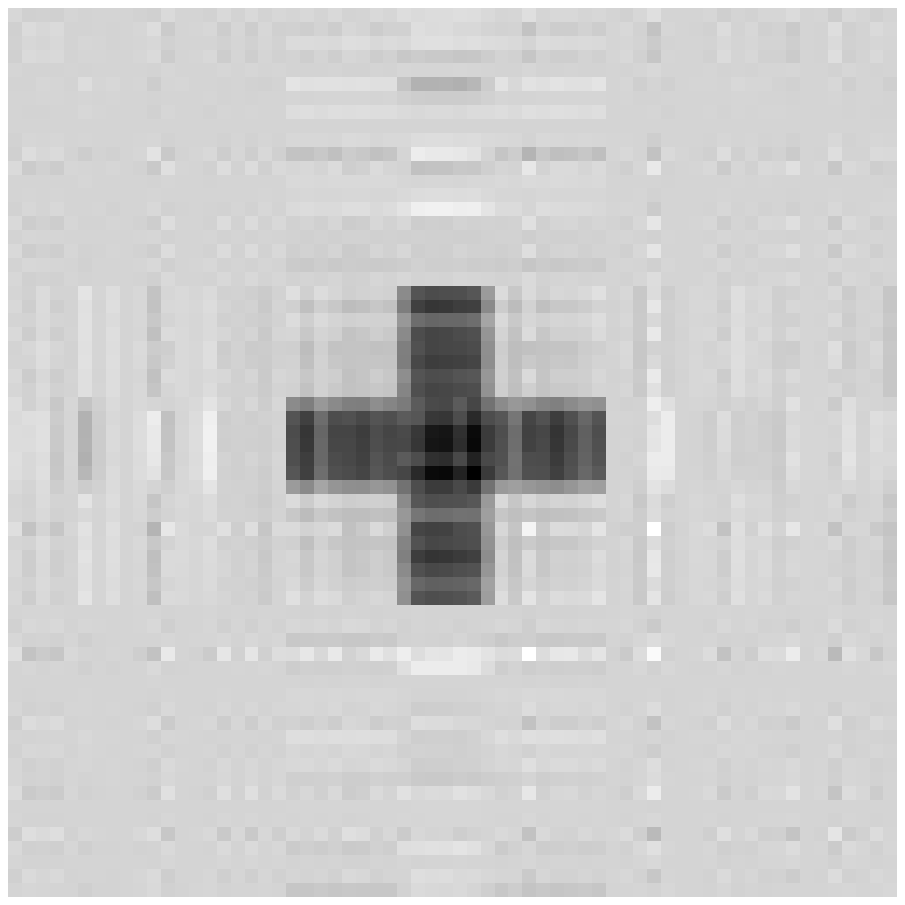} \\ \hline

\includegraphics[scale=0.11]{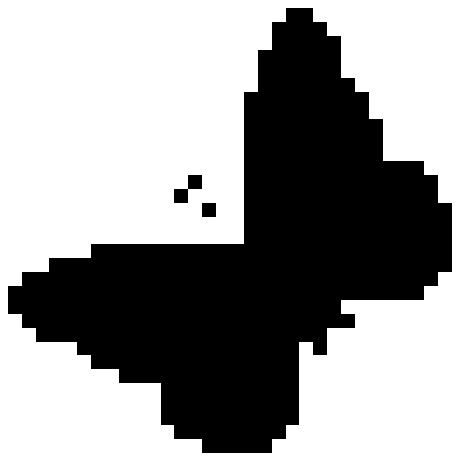} & \includegraphics[scale=0.11]{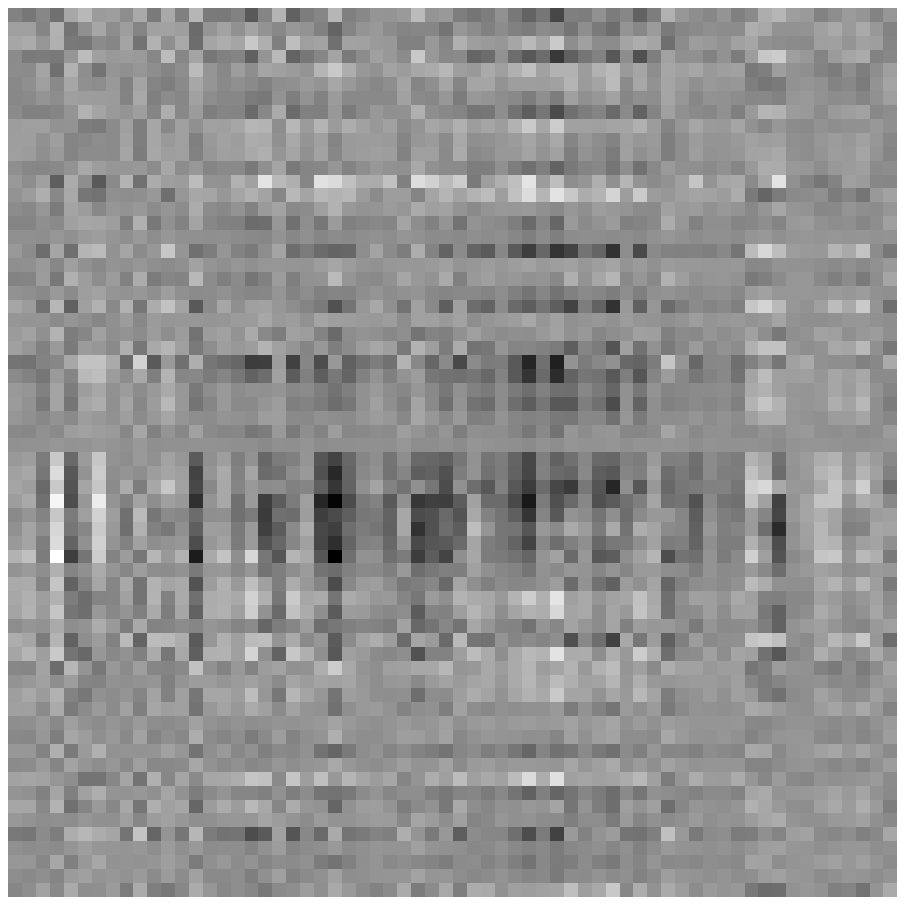} & \includegraphics[scale=0.11]{./fig-butterfly_500_TR1.eps}  & \includegraphics[scale=0.11]{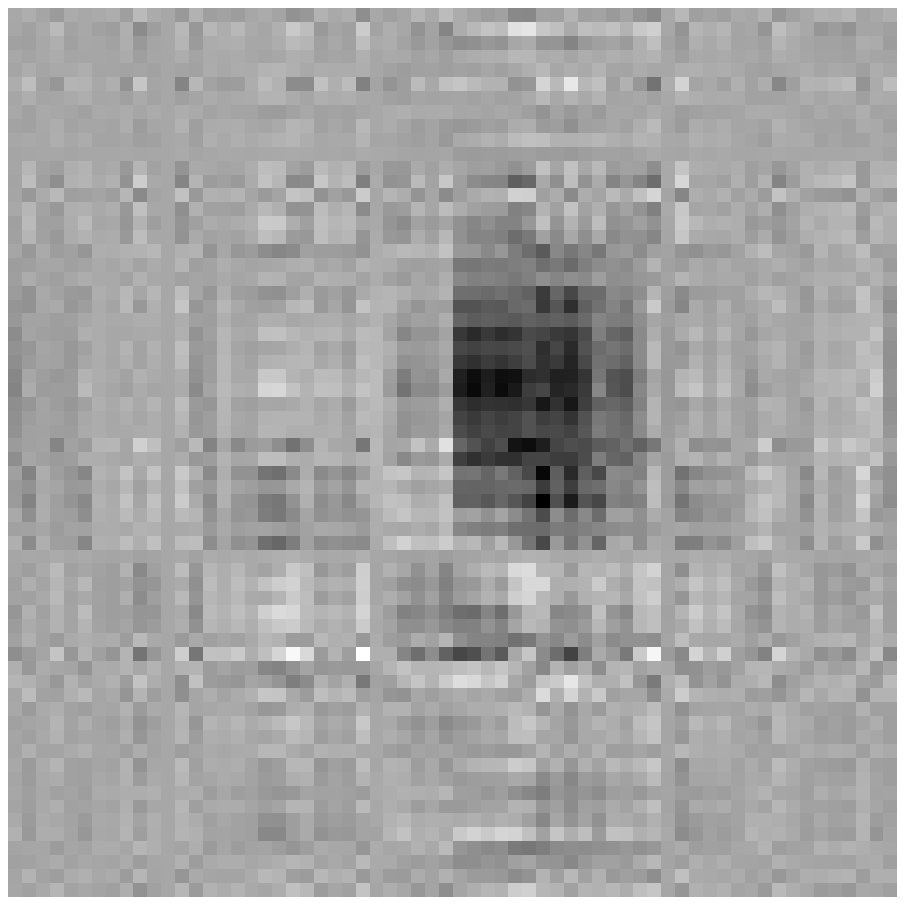} & \includegraphics[scale=0.11]{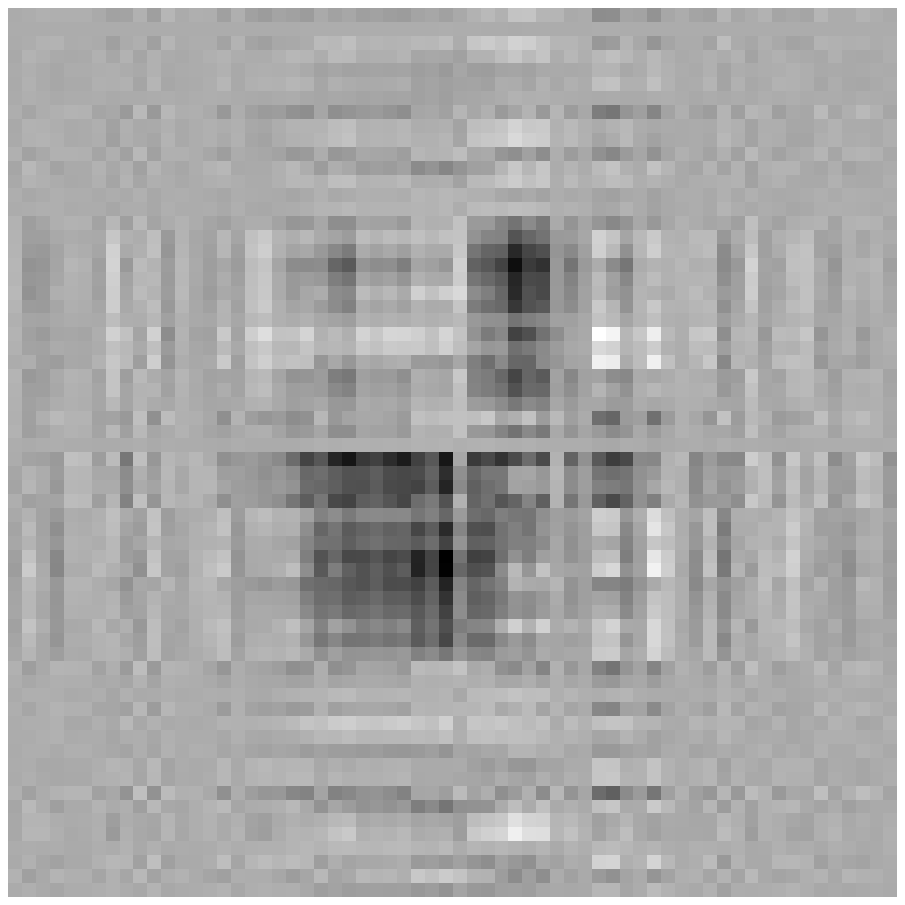} \\

&  \includegraphics[scale=0.11]{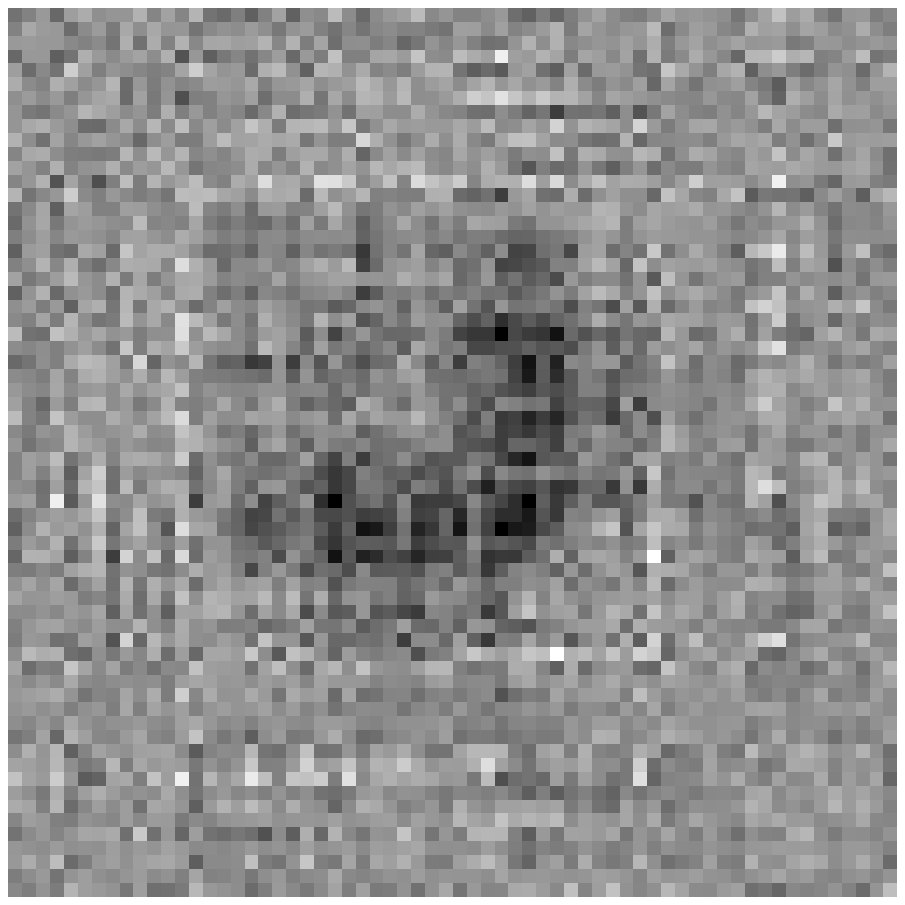} & \includegraphics[scale=0.11]{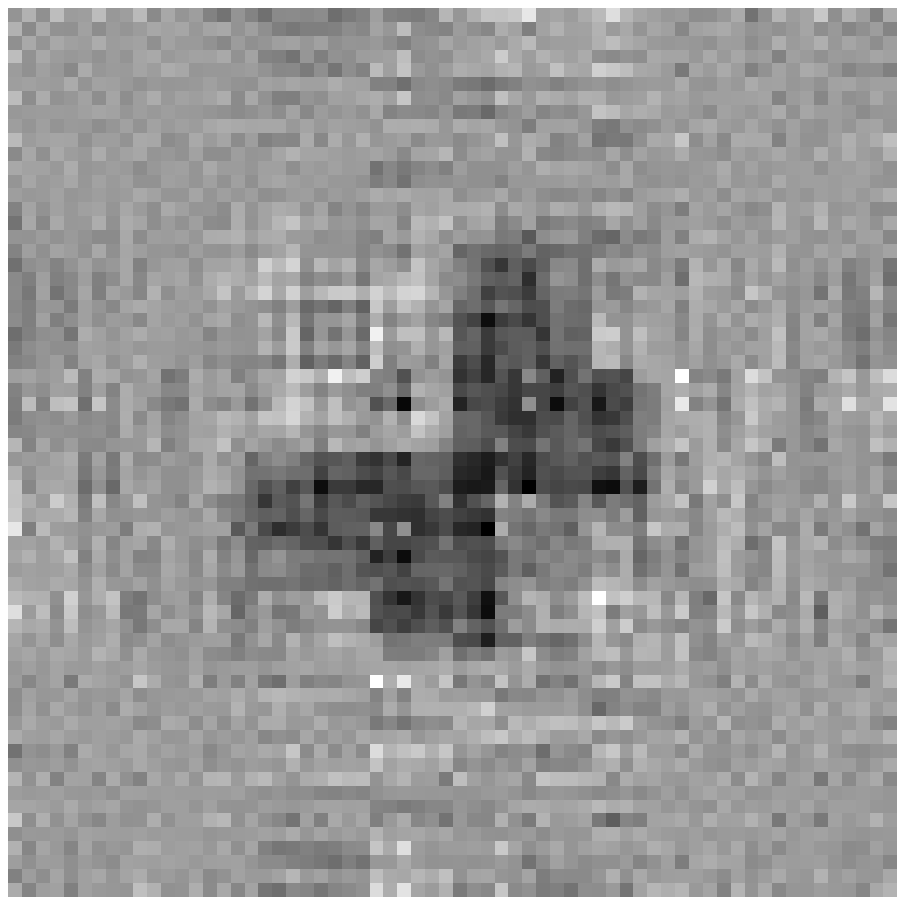}  & \includegraphics[scale=0.11]{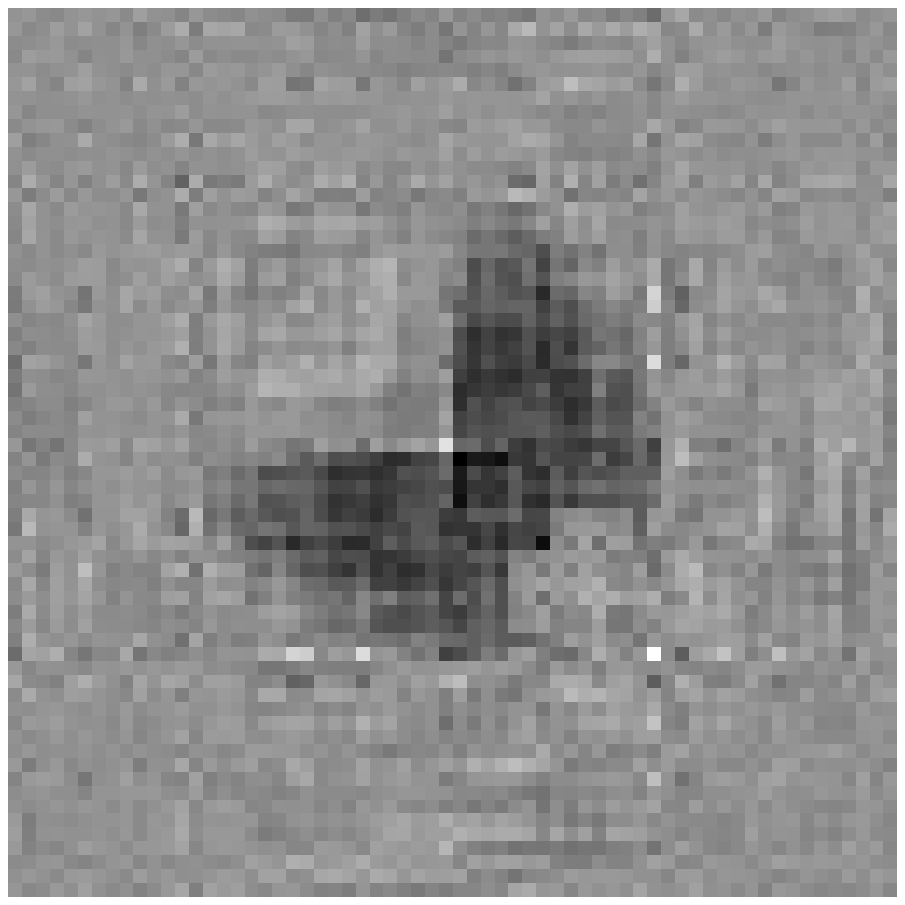} & \includegraphics[scale=0.11]{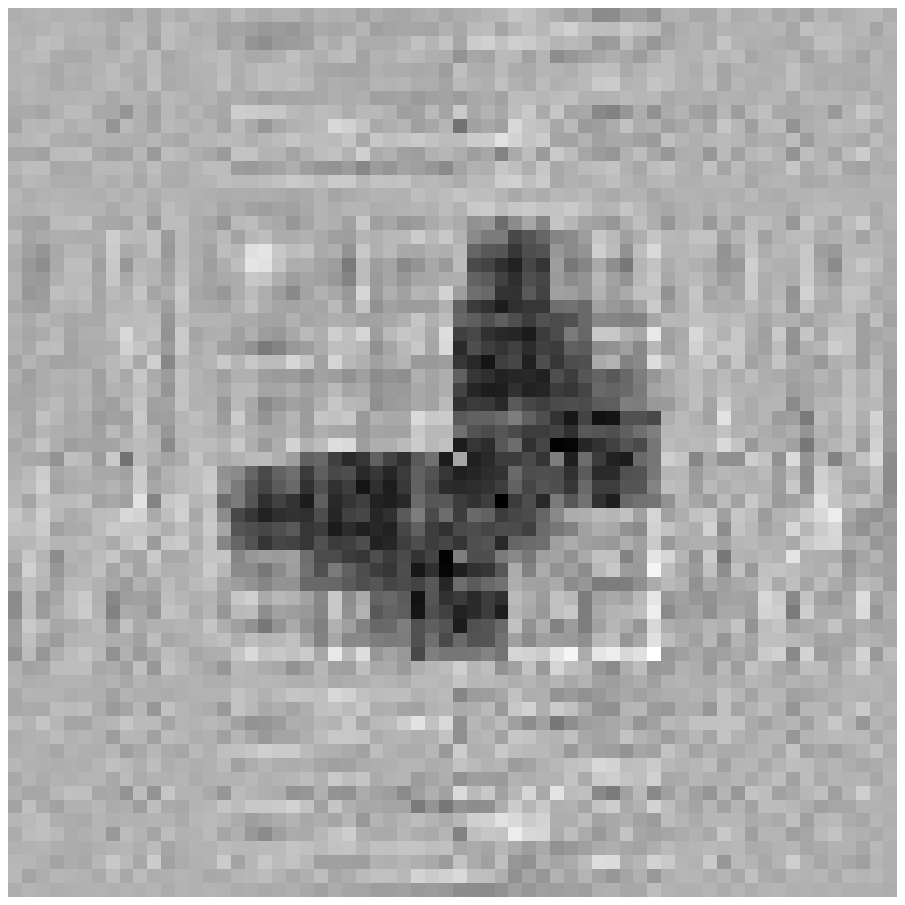} \\

&  \includegraphics[scale=0.11]{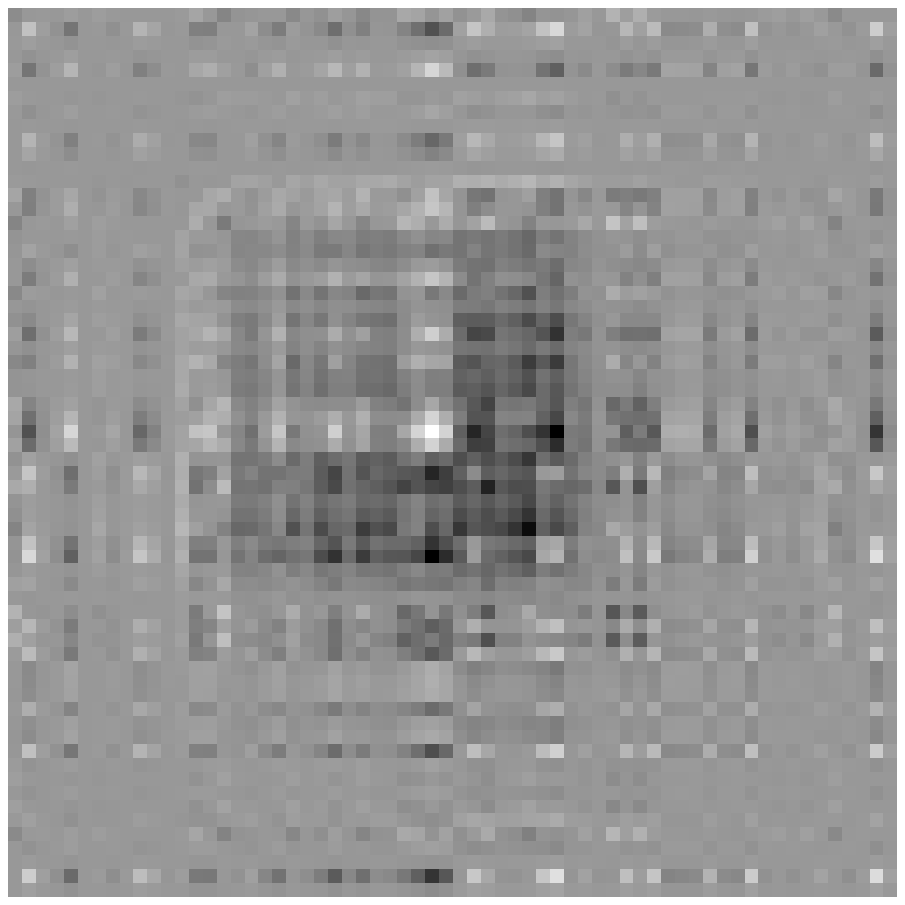} & \includegraphics[scale=0.11]{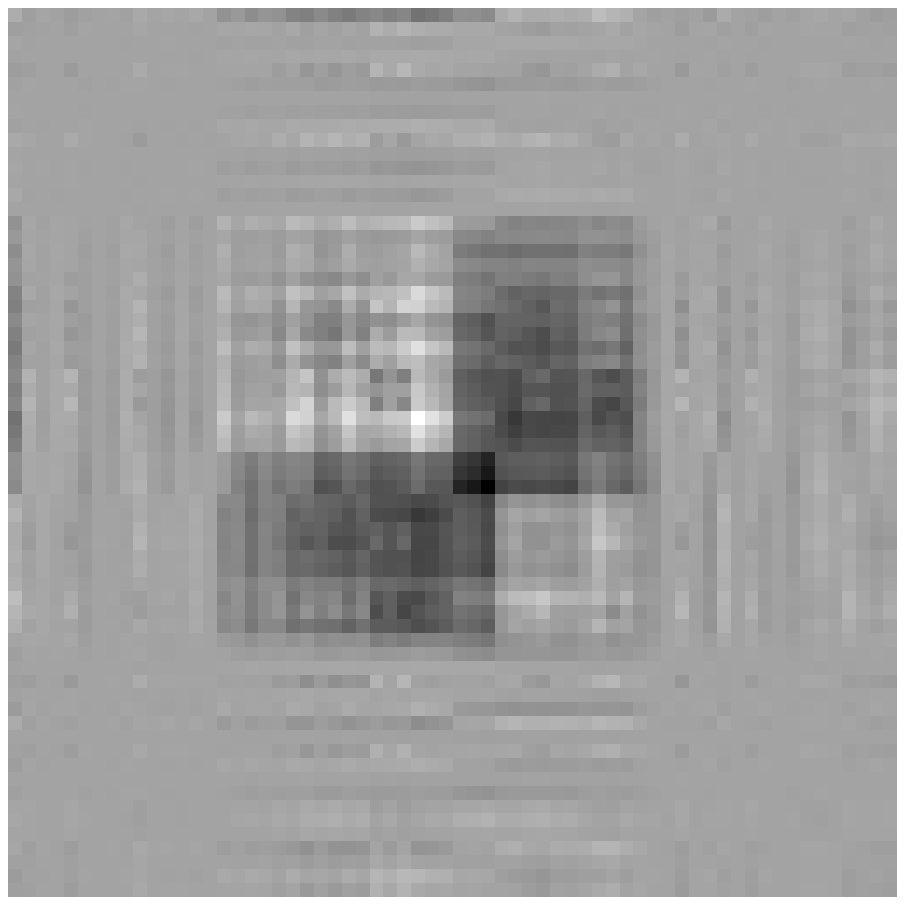}  & \includegraphics[scale=0.11]{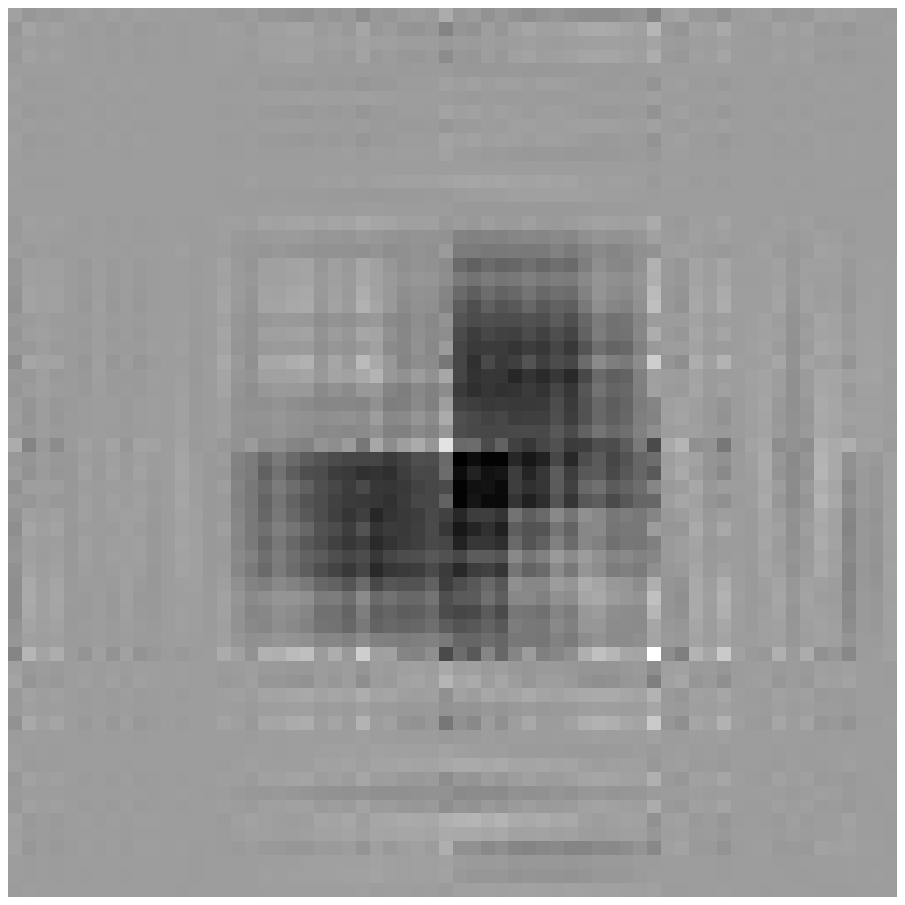} & \includegraphics[scale=0.11]{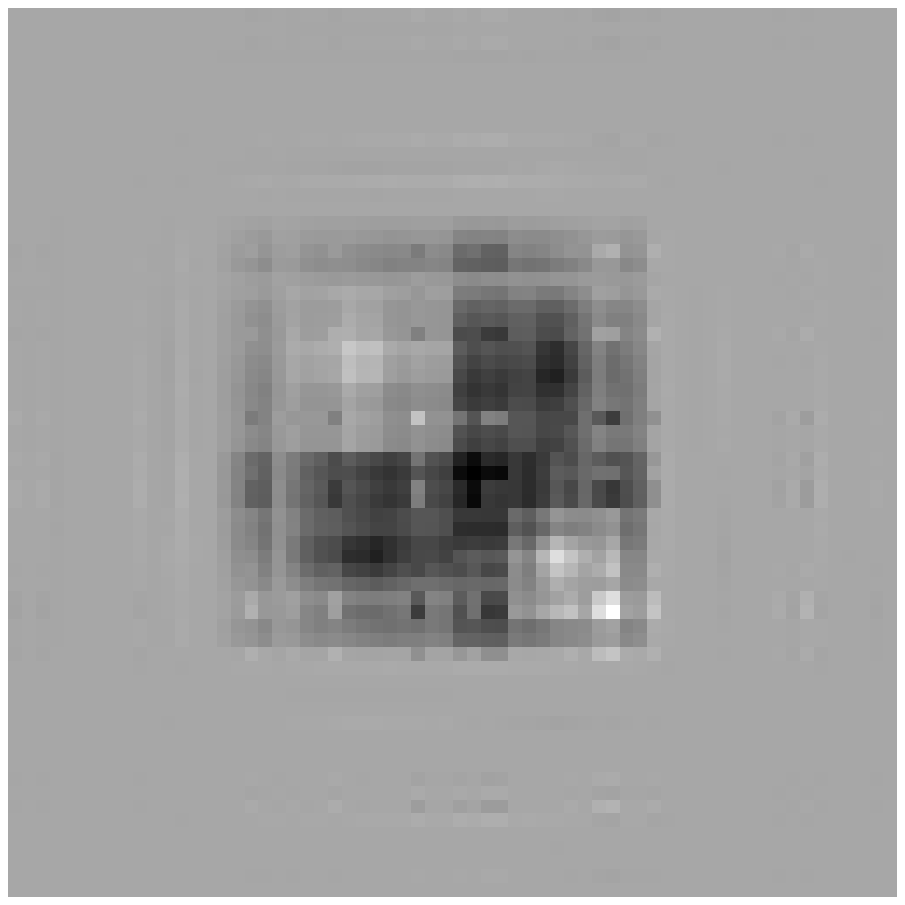} \\  \hline

\includegraphics[scale=0.11]{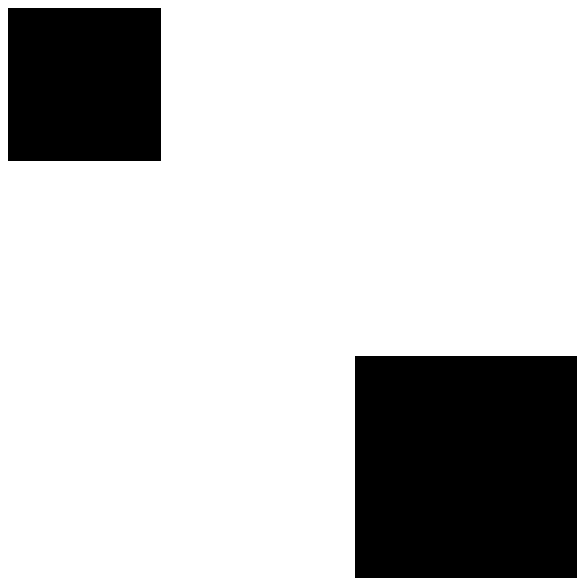} & \includegraphics[scale=0.11]{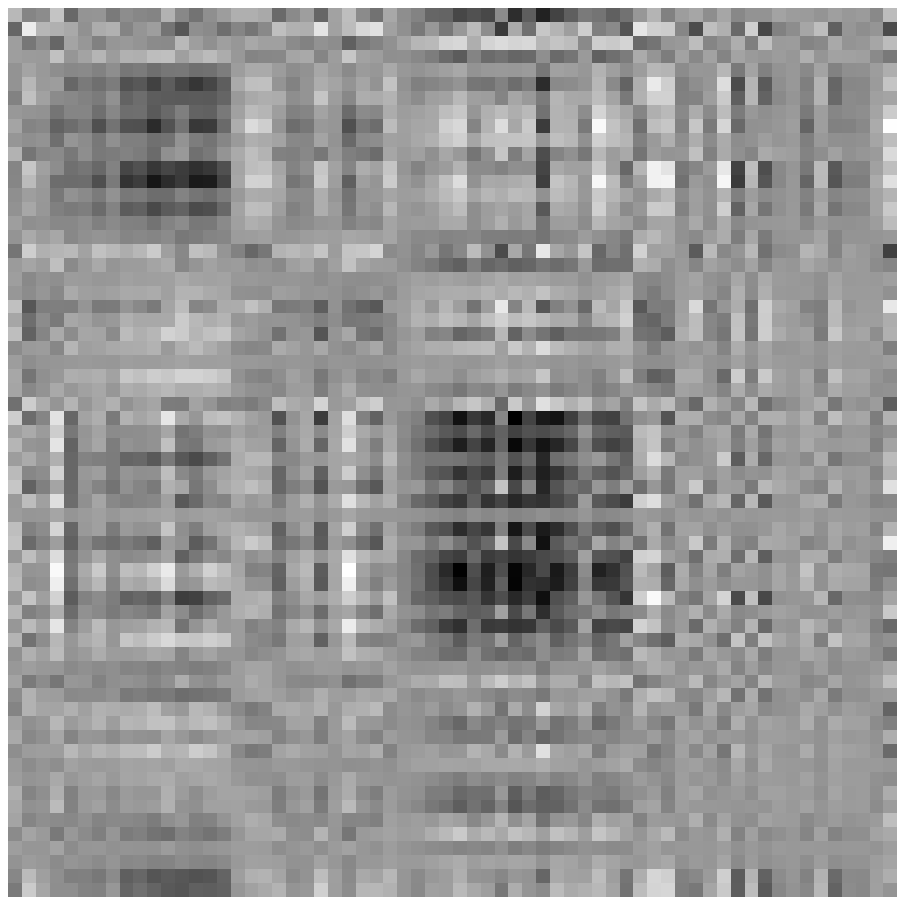} & \includegraphics[scale=0.11]{./fig-twobox_500_TR1.eps}  & \includegraphics[scale=0.11]{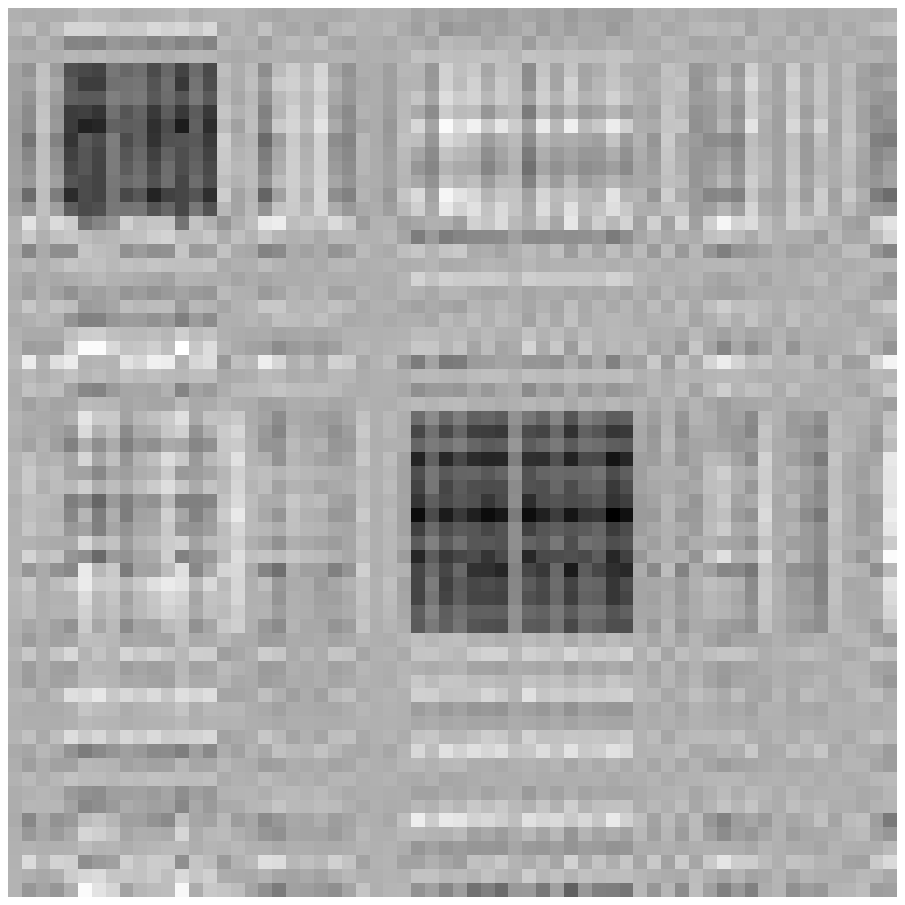} & \includegraphics[scale=0.11]{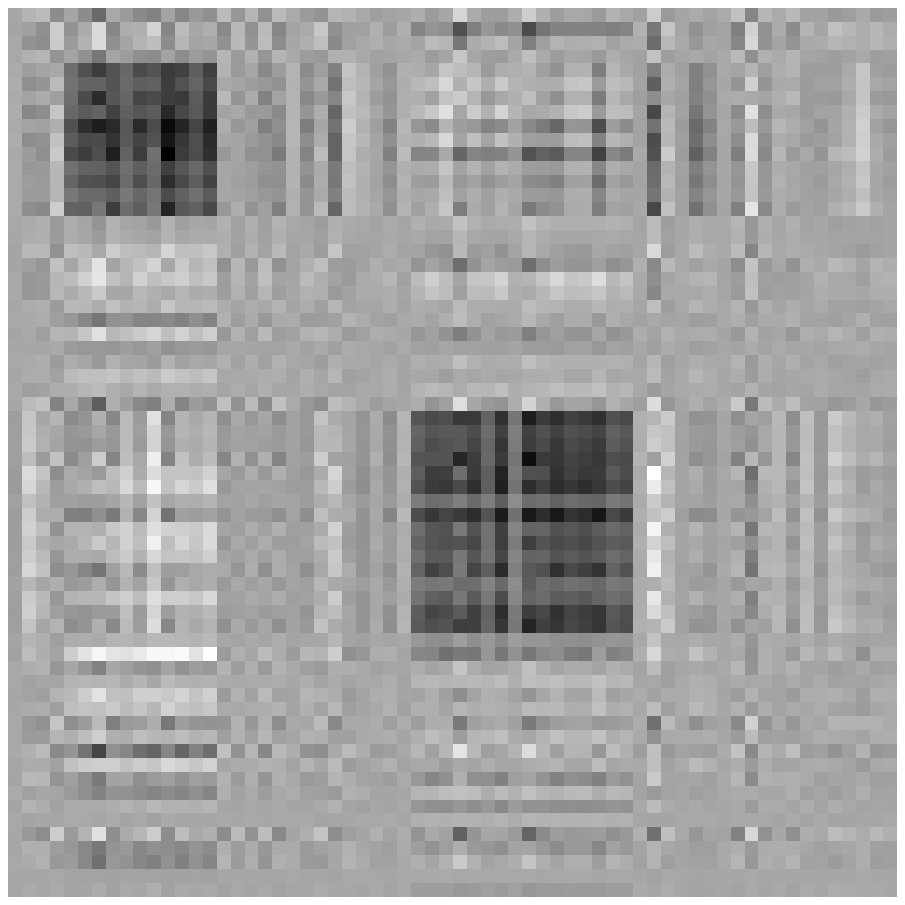} \\

&  \includegraphics[scale=0.11]{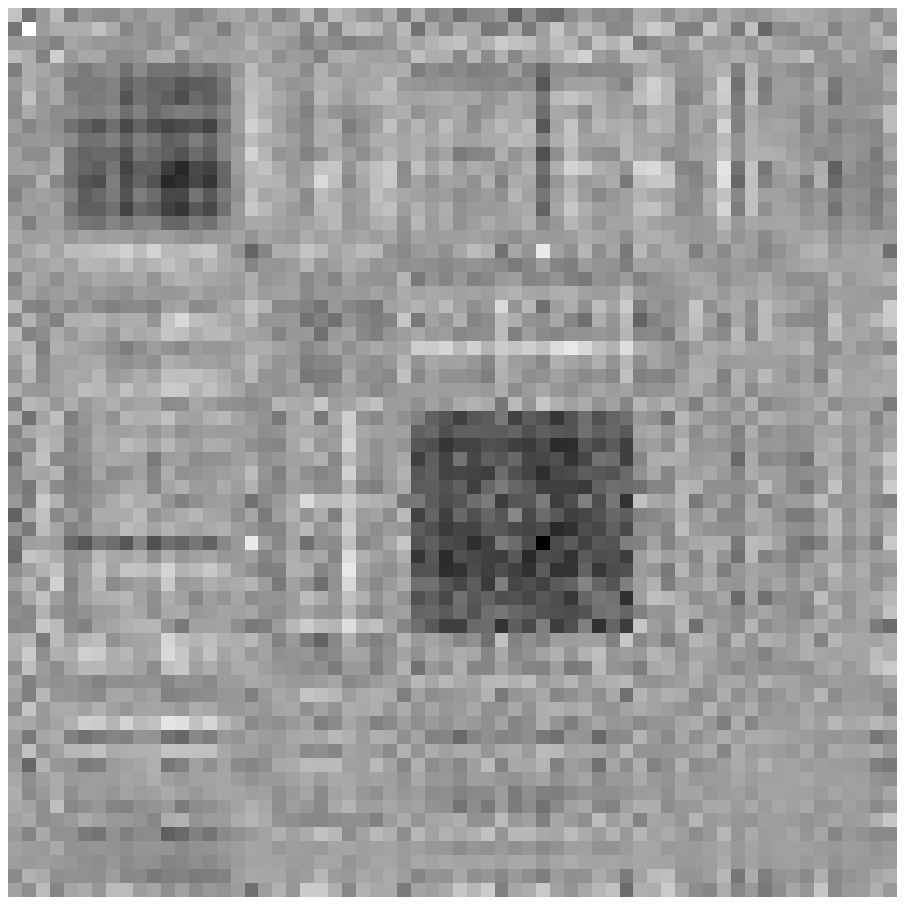} & \includegraphics[scale=0.11]{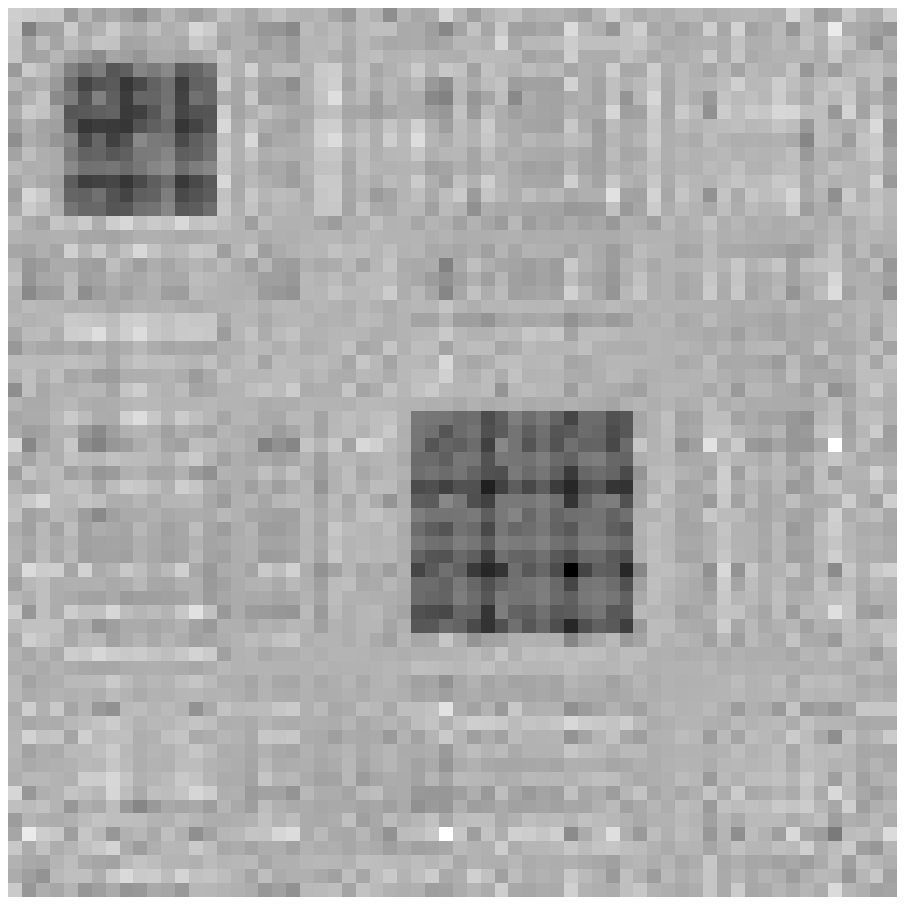}  & \includegraphics[scale=0.11]{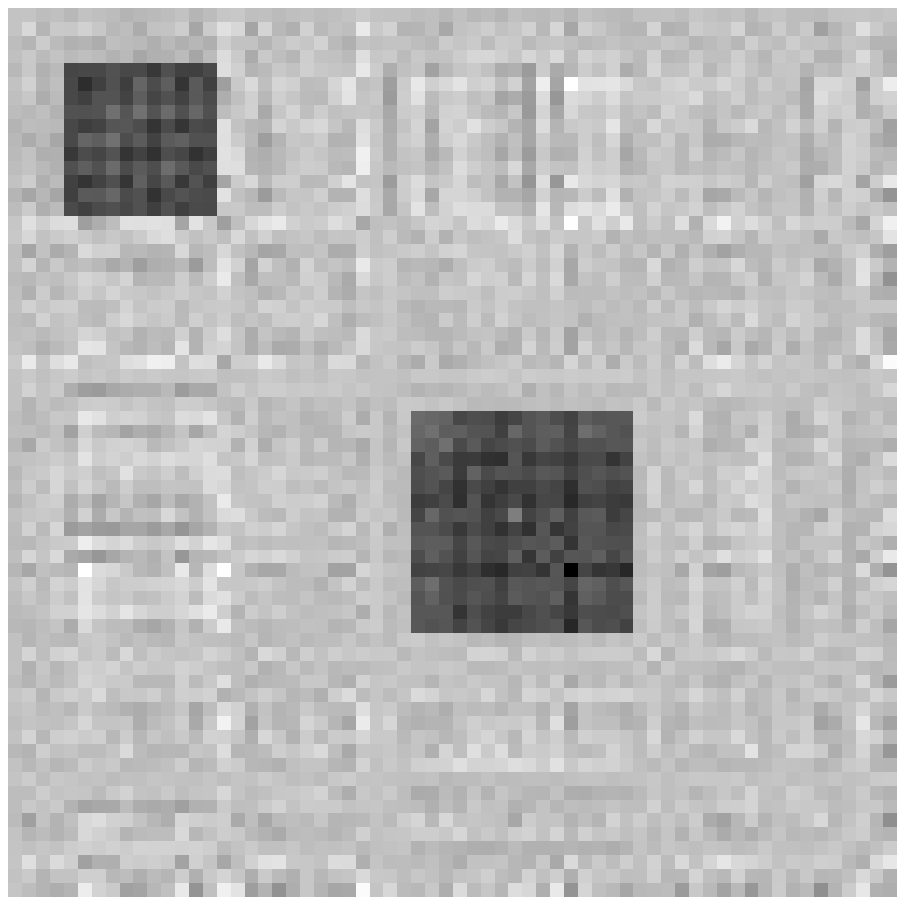} & \includegraphics[scale=0.11]{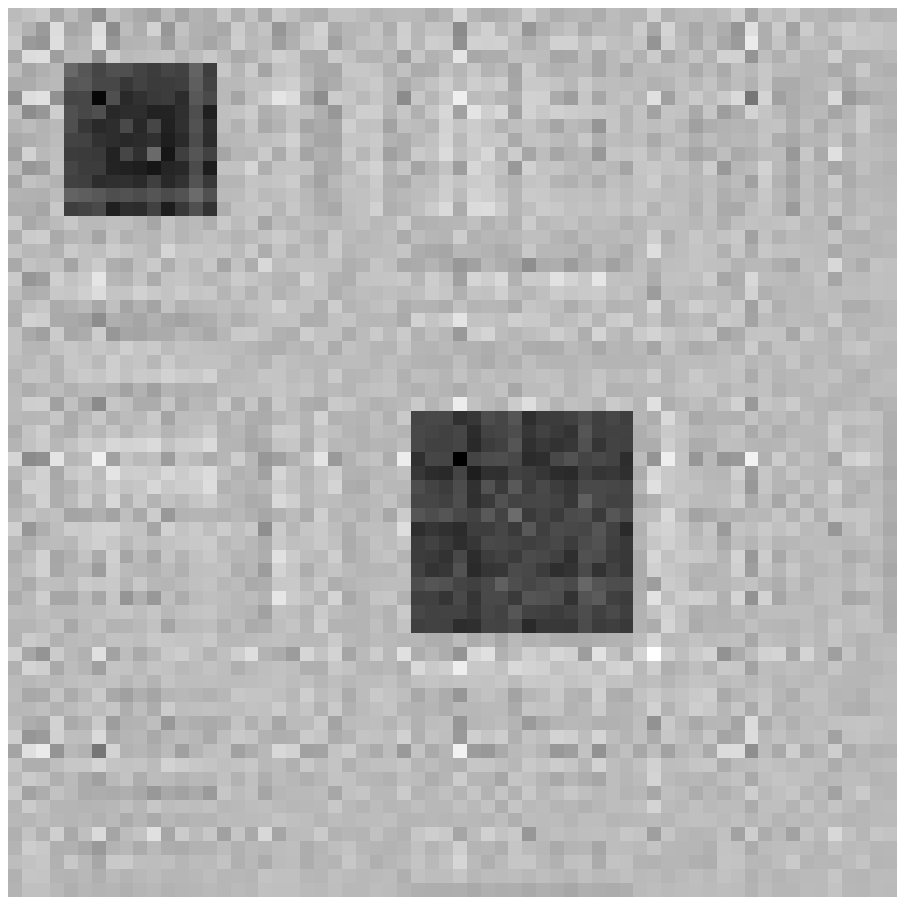} \\

&  \includegraphics[scale=0.11]{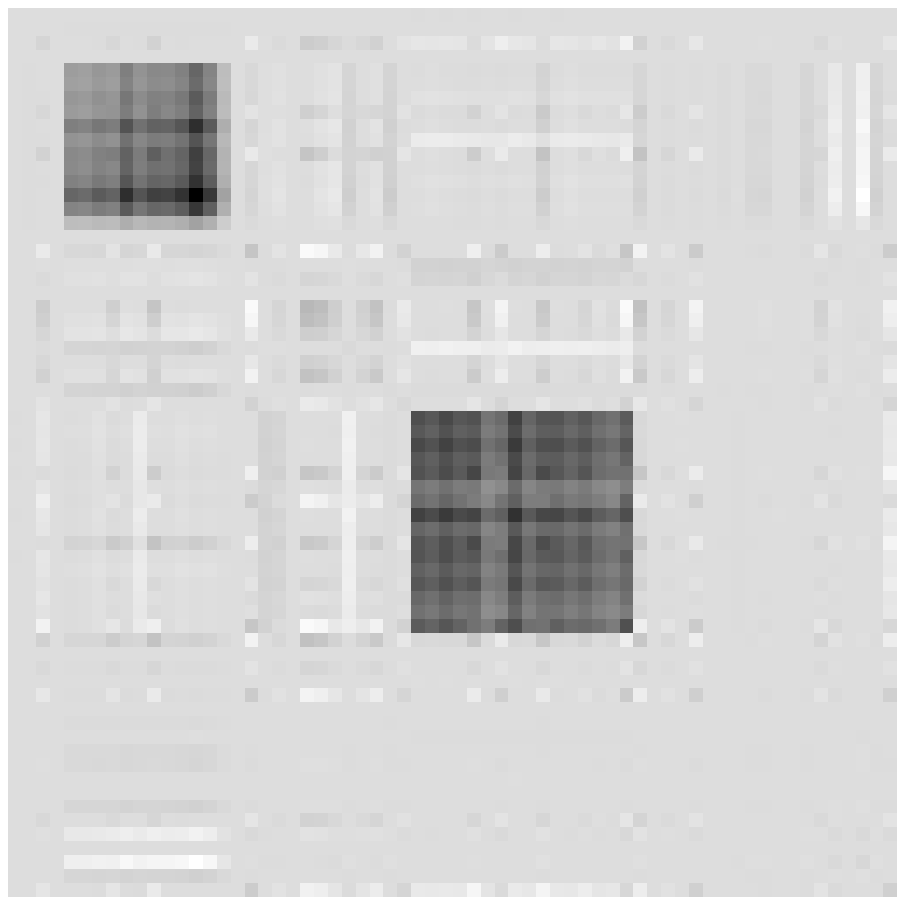} & \includegraphics[scale=0.11]{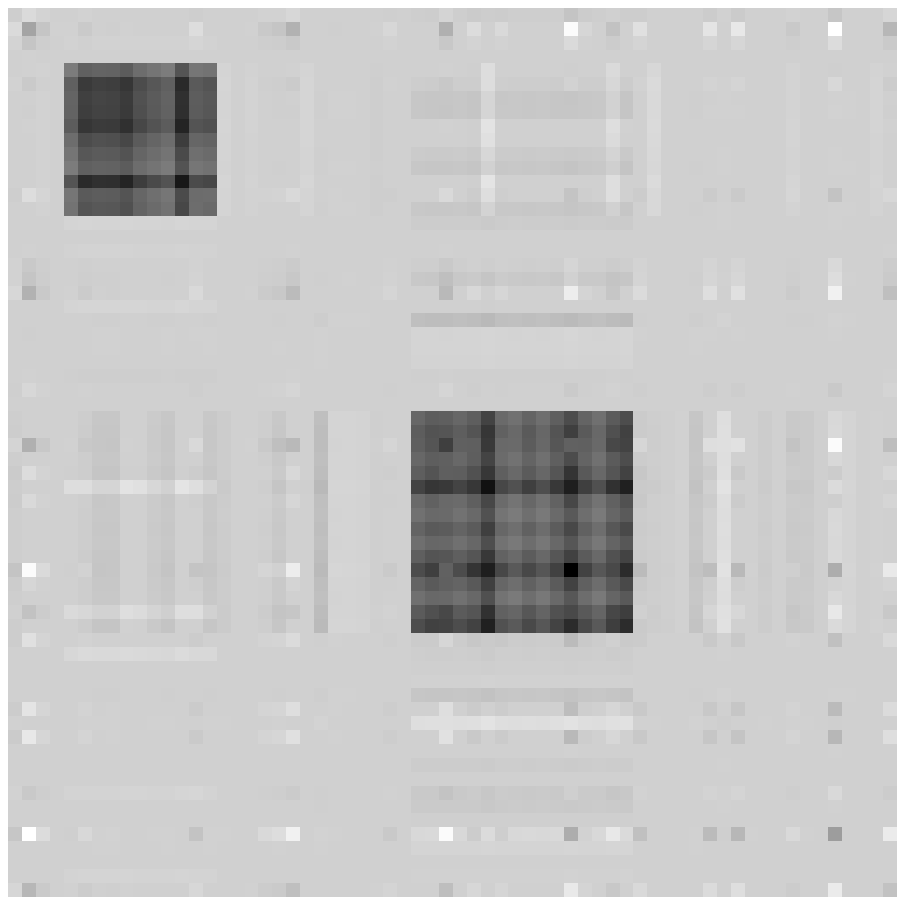}  & \includegraphics[scale=0.11]{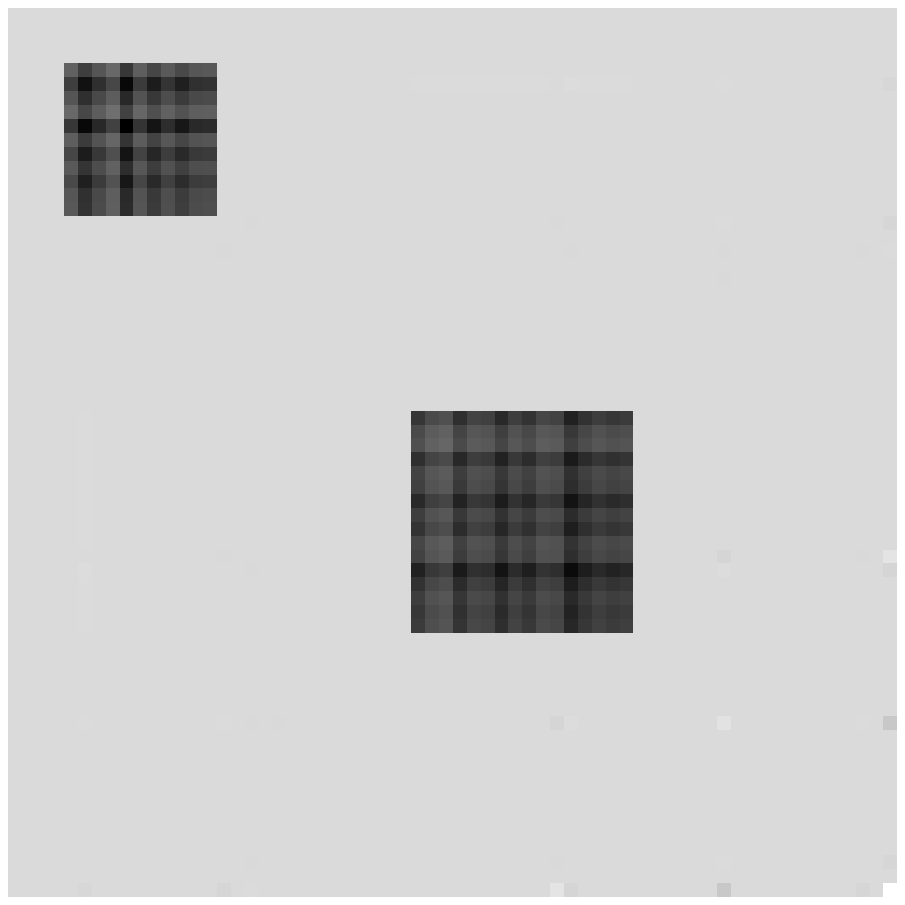} & \includegraphics[scale=0.11]{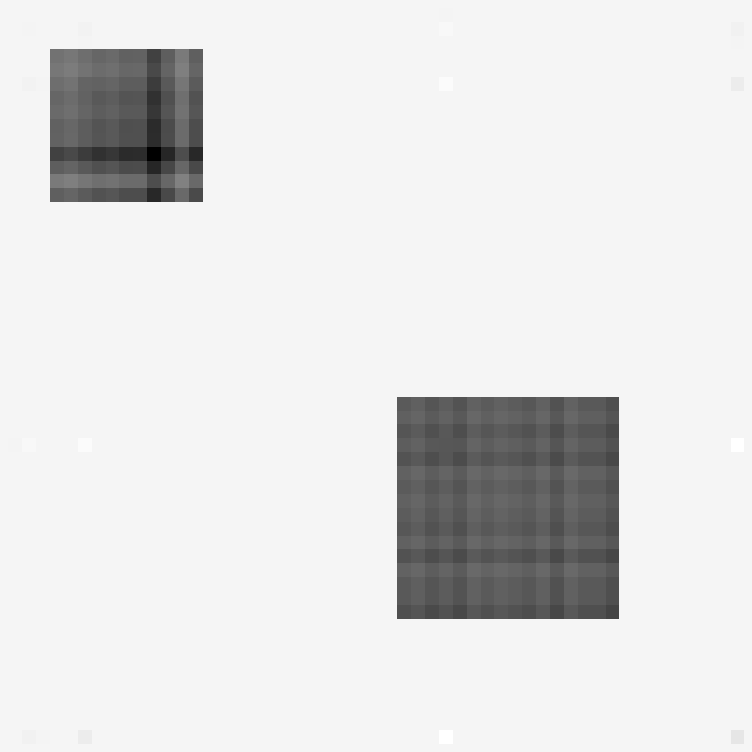} \\ \hline

\includegraphics[scale=0.11]{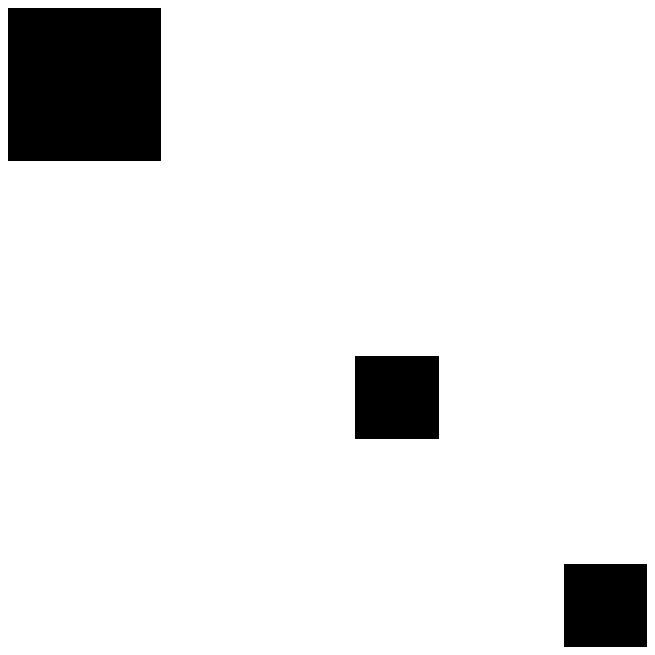} & \includegraphics[scale=0.11]{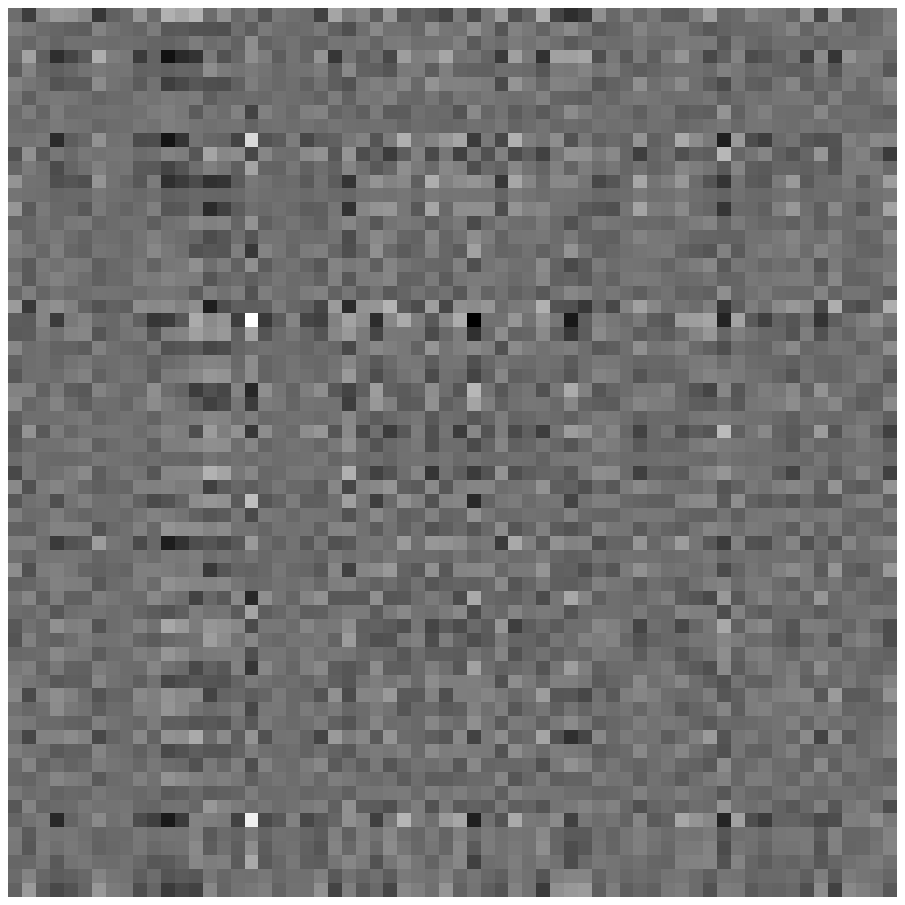} & \includegraphics[scale=0.11]{./fig-threebox_500_TR1.eps}  & \includegraphics[scale=0.11]{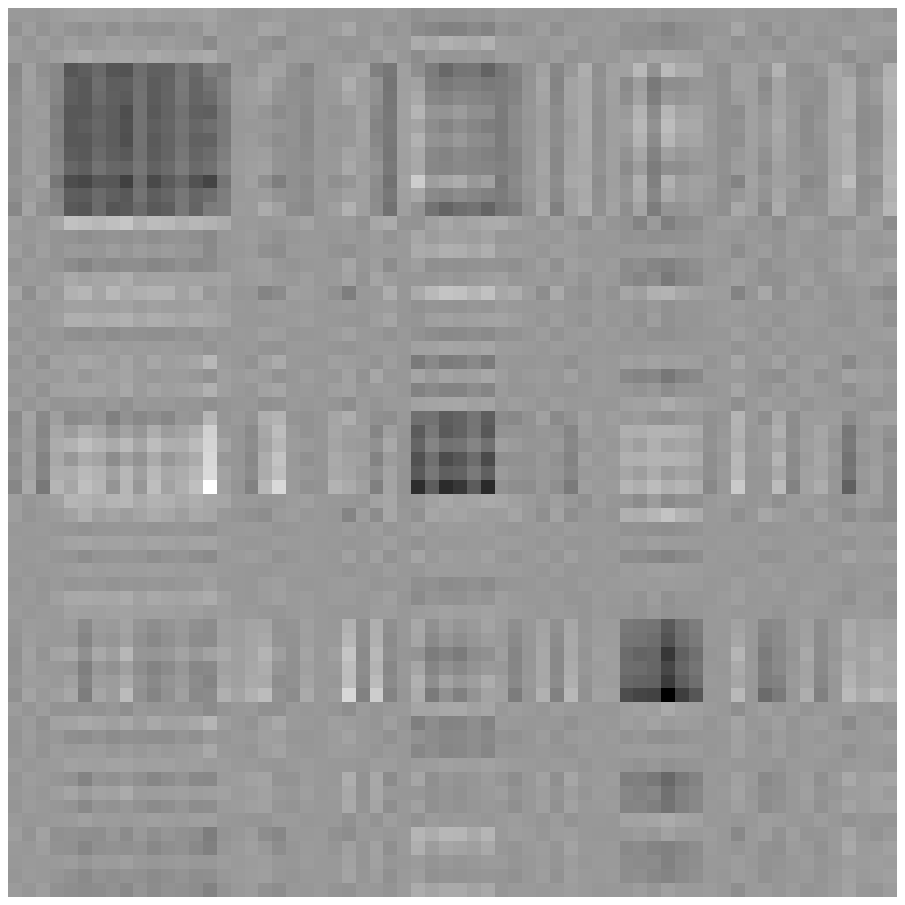} & \includegraphics[scale=0.11]{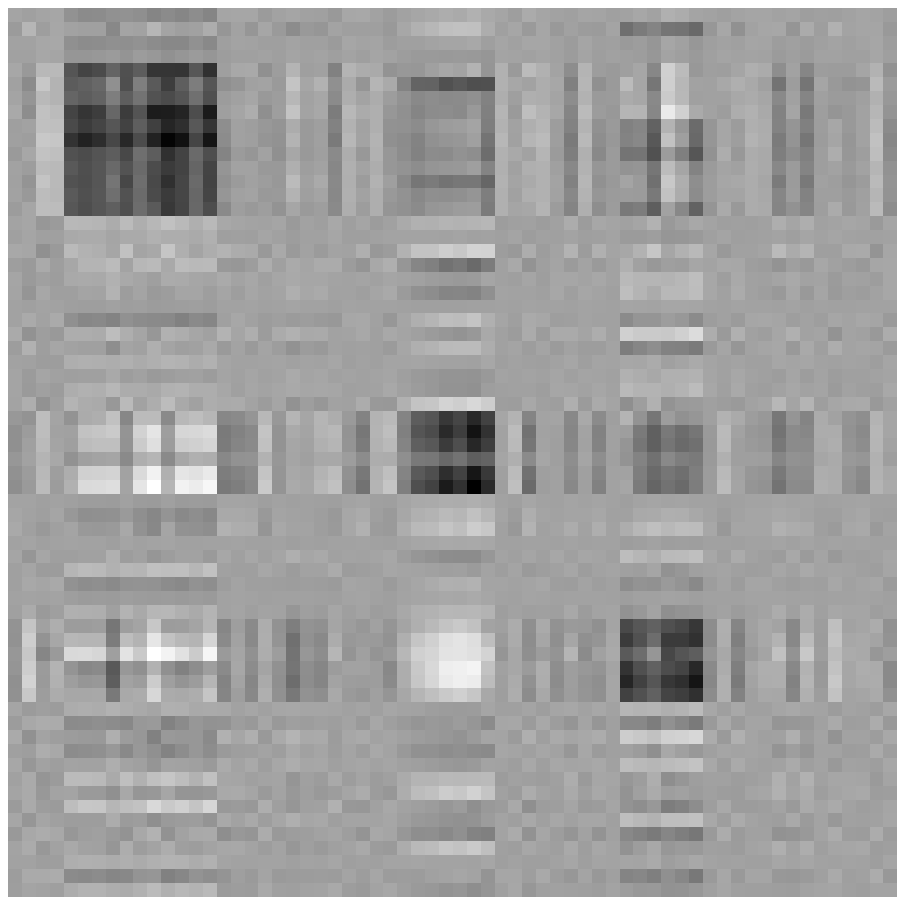} \\

&  \includegraphics[scale=0.11]{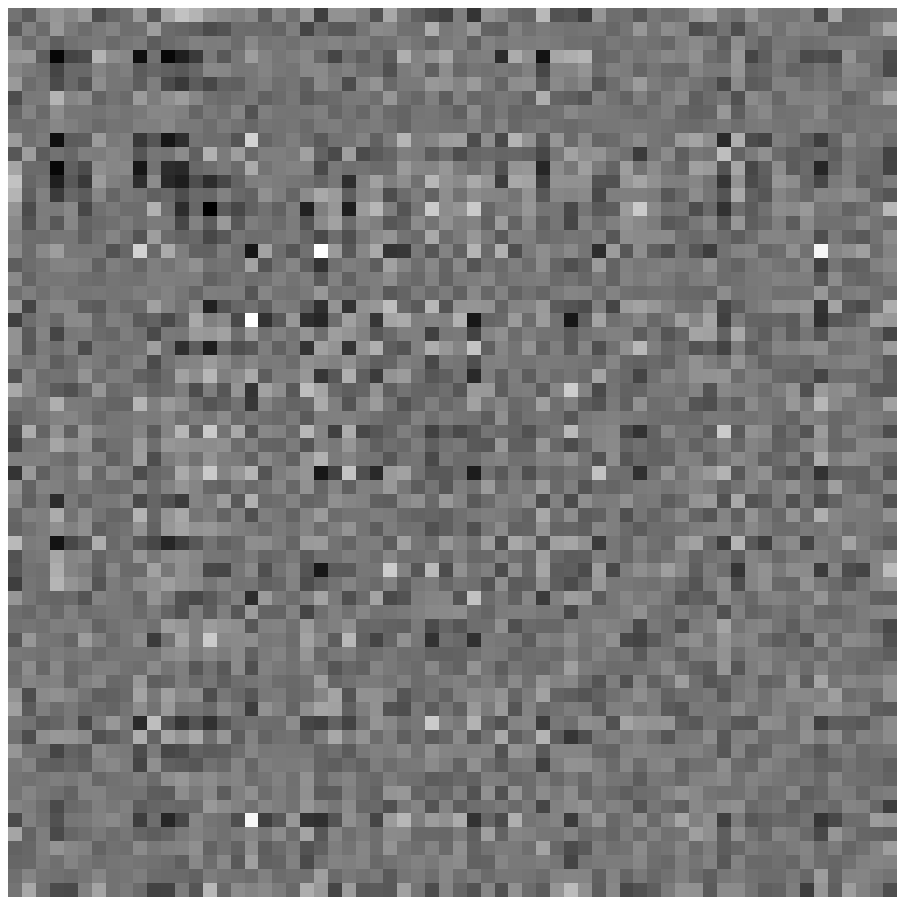} & \includegraphics[scale=0.11]{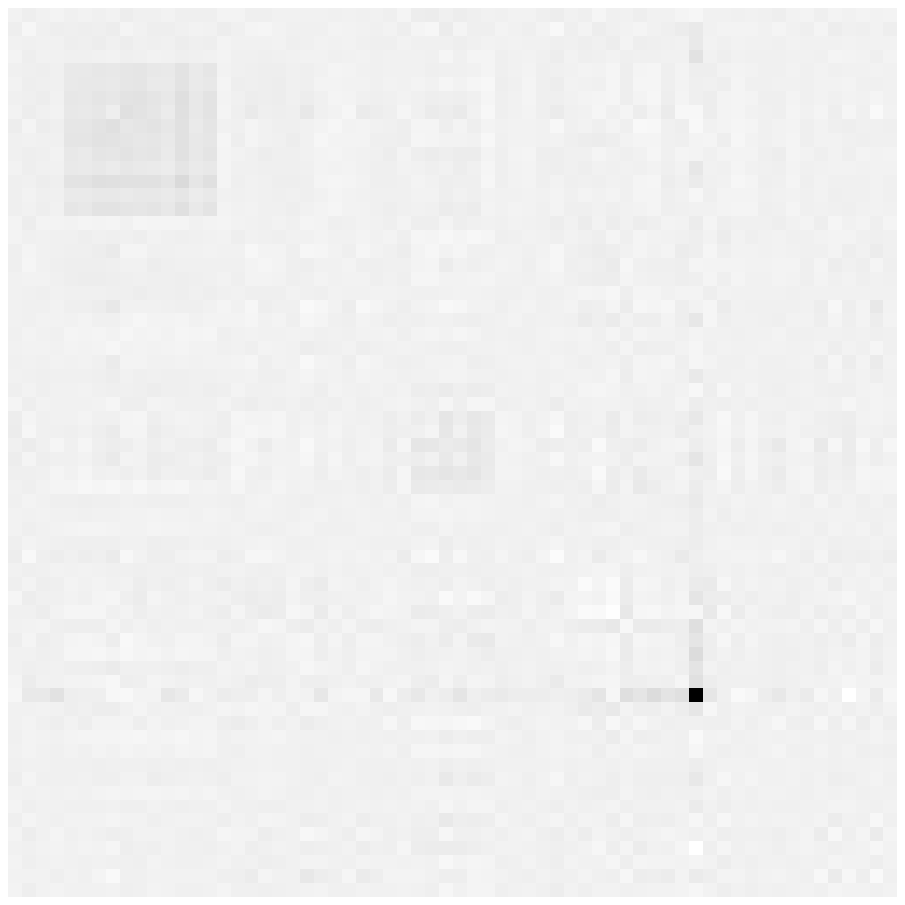}  & \includegraphics[scale=0.11]{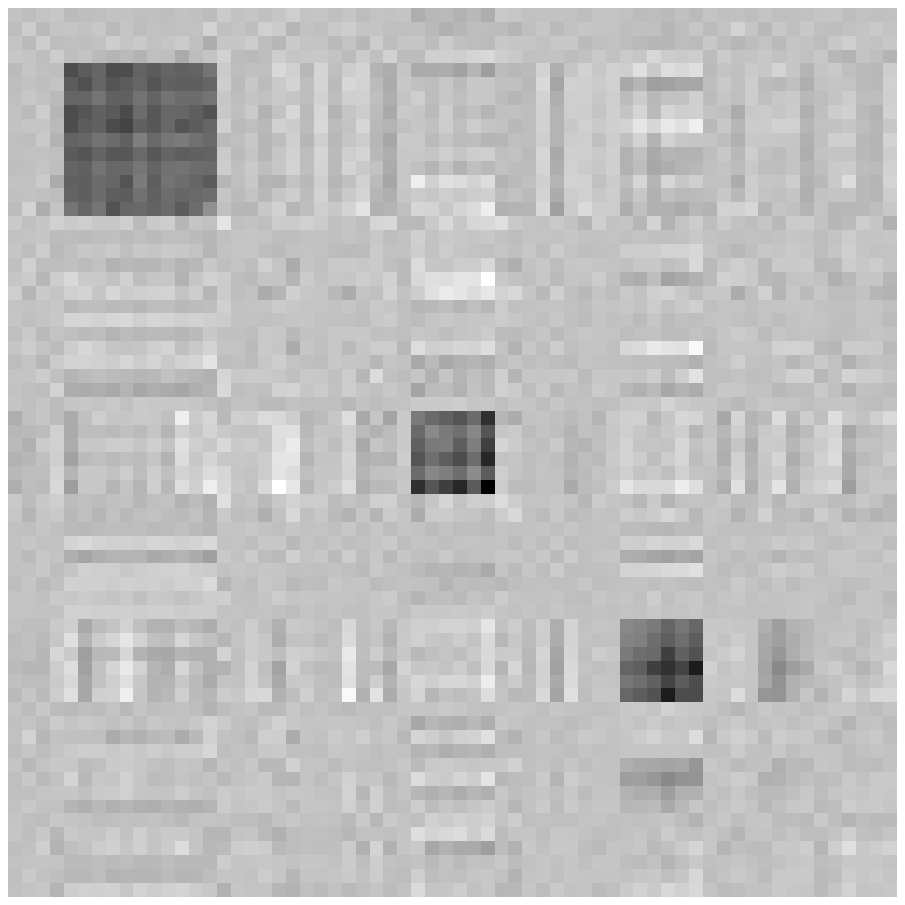} & \includegraphics[scale=0.11]{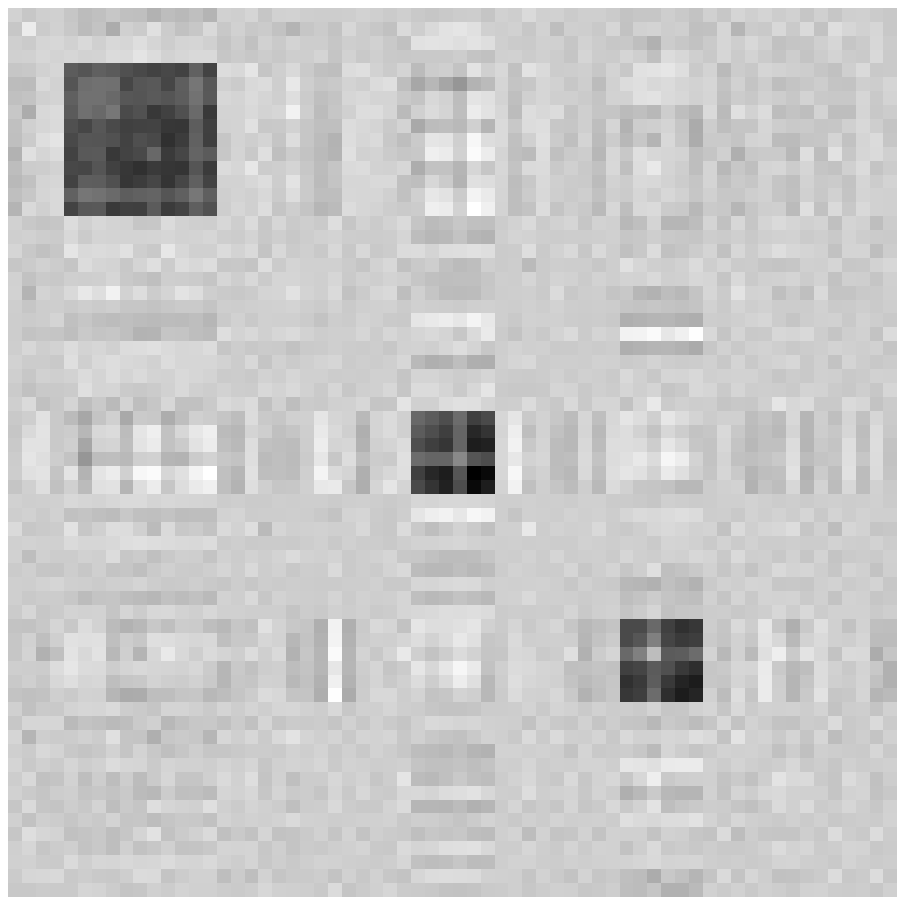} \\

&  \includegraphics[scale=0.11]{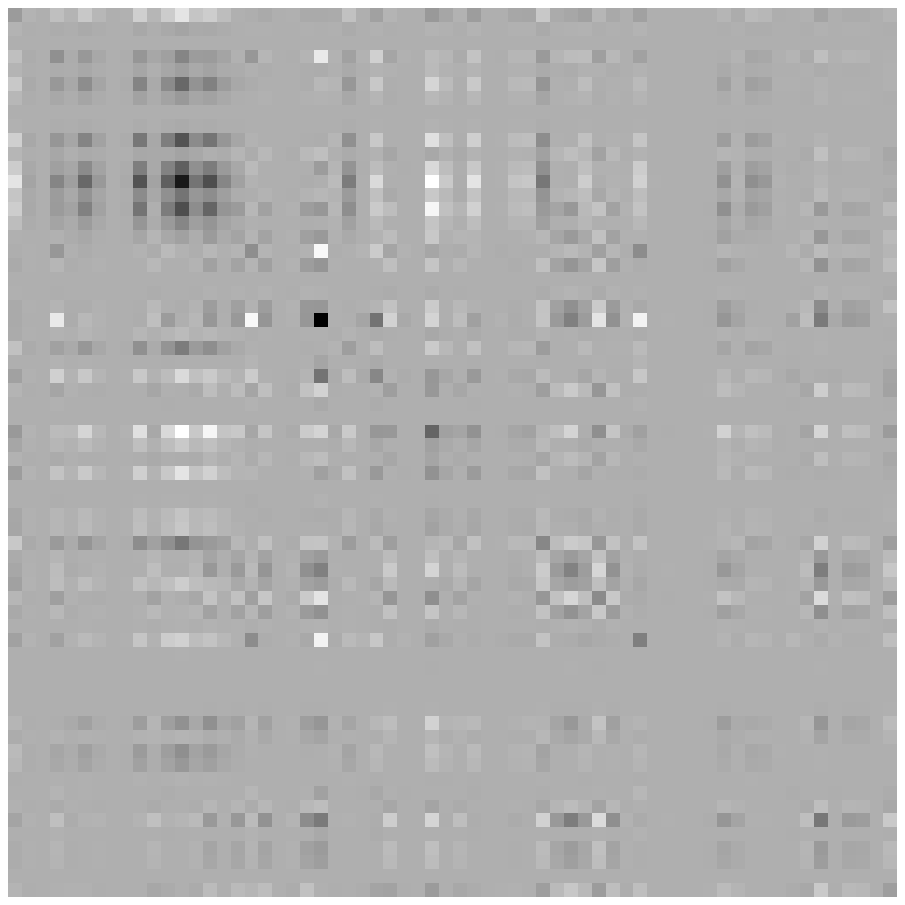} & \includegraphics[scale=0.11]{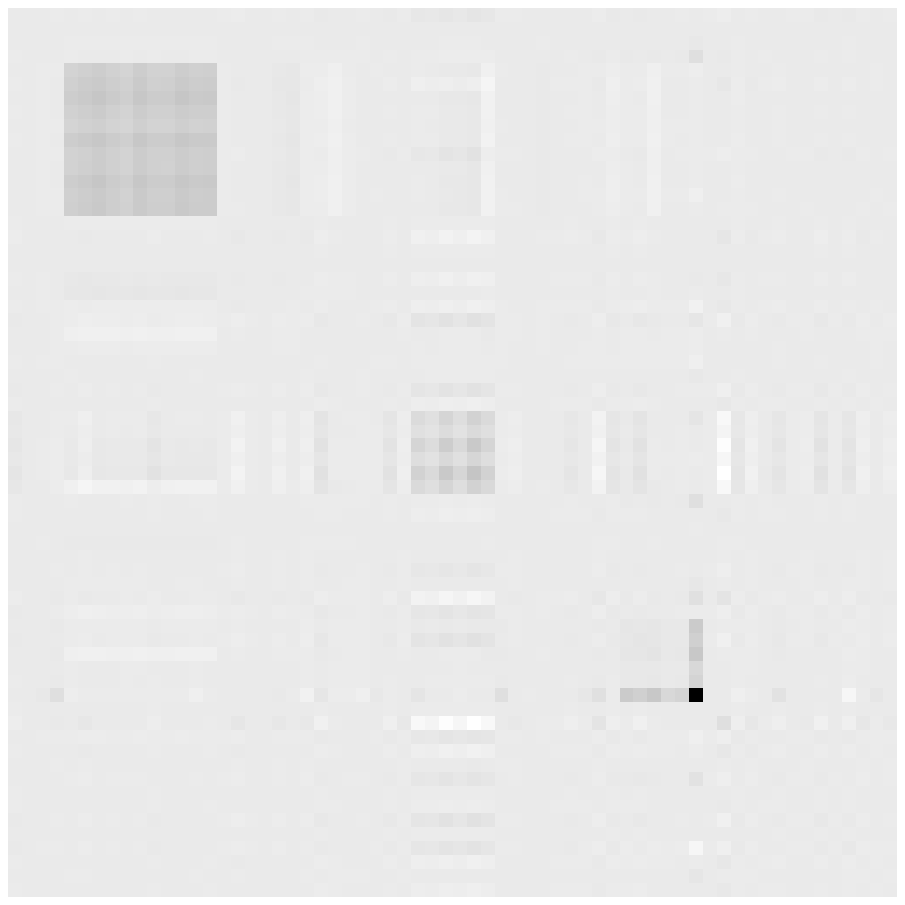}  & \includegraphics[scale=0.11]{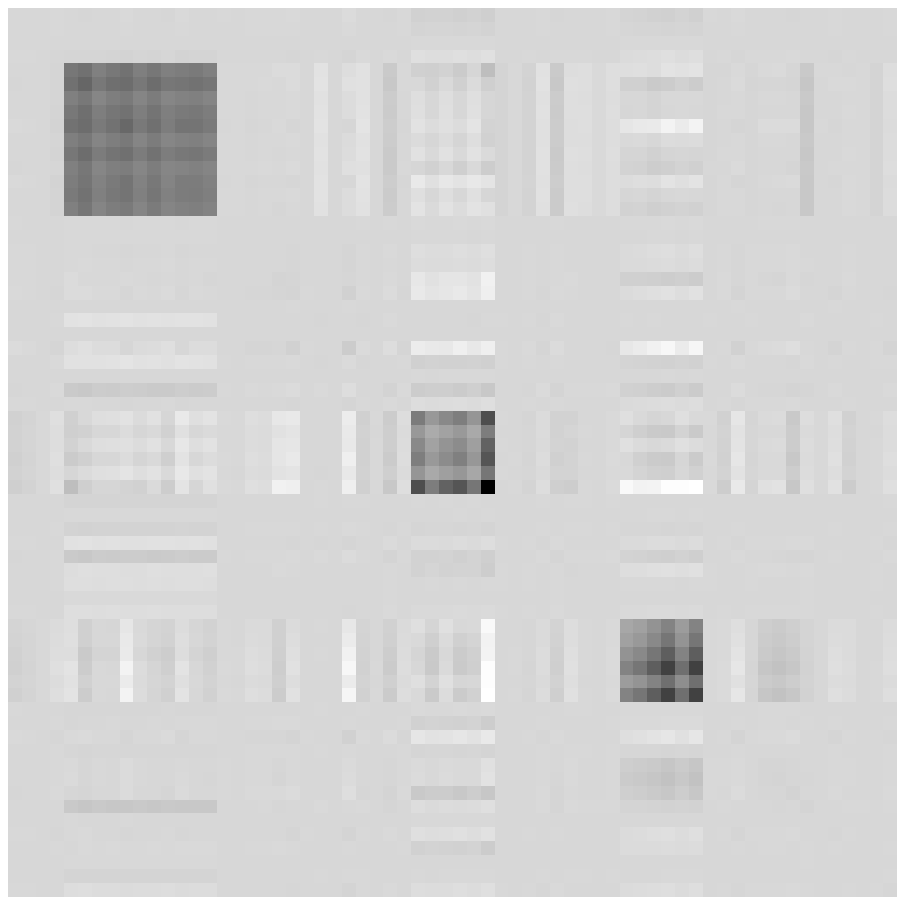} & \includegraphics[scale=0.11]{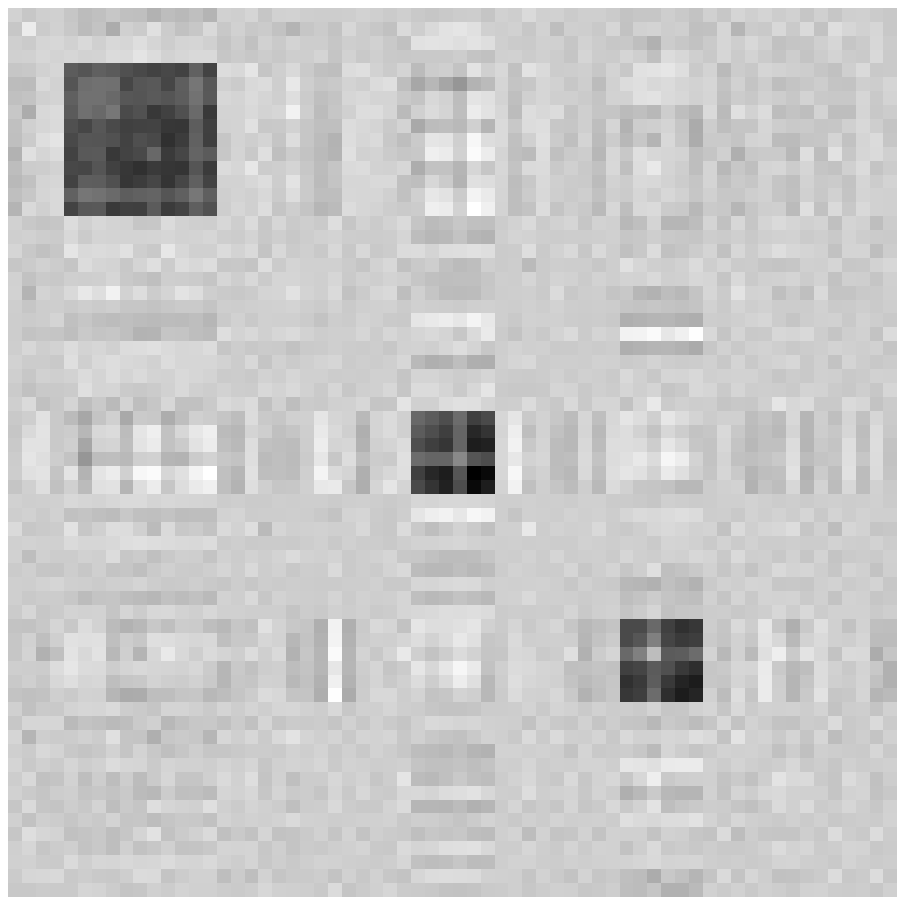} \\ \hline
\end{tabular}
\caption{A snapshot of the outcome for our model, CP and symmetric CP regression based on one random data generation. For each $\Bcal_0$, the top row gives the outcome of the standard CP tensor regression method of \cite{ZhouLiZhu2013}, the second row gives the symmetric CP regression, the third row gives the proposed symmetry tensor regression outcome. }
\label{fig:sim-shape}
\end{center}
\end{figure}

We conduct simulation studies to examine the performance of the proposed symmetric tensor regression, comparing with the standard CP tensor regression and symmetric CP regression, under a variety of signals and sample sizes. We choose these two baselines among others because their solutions also have low-rank structures. Also, the superiority of tensor regression compared with other baselines methods have been shown in \cite{zhou2013tensor,li2018tucker}. We also show that the additional symmetry constraint in our tensor regression model improves the estimation and prediction accuracy when using the symmetric covariate matrix as input. 
Toward that end, we generate the response $y_i$ according to a normal distribution with the mean value given by:
\begin{eqnarray*}
\mu_i = \gammabf\trans \zbf_i + \langle \Bcal, \xbf_i \rangle, \quad i=1, \ldots, n, 
\end{eqnarray*}
with standard Gaussian noise. Here, $\zbf_i \in \real{5}$ is the vector of scalar predictors also generated from the standard normal distribution. The corresponding coefficient $\gammabf \in \real{5}$ has all elements equal to one, and $\xbf_i \in \real{64 \times 64}$ is the symmetric correlation matrix predictor generated by the \emph{gallary} function in Matlab\footnote{The \emph{gallary} function is designed to generate random test matrices: https://www.mathworks.com/help/matlab/ref/gallery.html}. We also use $\Bcal$ to denote the symmetric signal matrix. In this simulation, we choose $\Bcal$s from both the low-rank and high-rank structures. Also, we experiment on $\Bcal$ that have single and multiple connected components. For instance, the \texttt{butterfly} shape has one connected component and high-rank structure, the \texttt{two box} and \texttt{three box} shapes have several connected components and low-rank structure (Figure \ref{fig:sim-shape}).

We experiment with the sample size n = $\{500, 700, 900, 1100\}$, and fix the rank at $R=3$ for our model, the symmetric and standard CP tensor regressions. We comment that the true signal tensor rarely possesses a low-rank structure in the real world. However, given the limited sample size in the relevant imaging and functional connectivity studies, a low-rank estimation, such as with rank=3, often provides a reasonable estimation even when the true signal has a higher rank \cite{ZhouLiZhu2013,li2018tucker}. In practice, the rank can always be treated as tuning parameter and selected using the cross-validation. Here, we also use cross-validation to select the tuning parameter $\rho$. The initial values for our model are constructed according to the method in Section \ref{sec:initial}.

To evaluate the accuracy of the estimations, we report a snapshot of the estimated $\Bcal$ based on one single data generation, which we show in Figure~\ref{fig:sim-shape}. The \emph{average mean squared error} of the estimation of parameters $\Bcal$, and the prediction of $Y$ for the 100 data replications are reported in Table~\ref{tab:sim-shape1} and Table~\ref{tab:sim-shape2}. Finally, we choose $\Bcal$ from the shapes of \texttt{circle}, \texttt{cross}, \texttt{butterfly}, \texttt{two boxes} and \texttt{three boxes}.

\begin{table}[hbt]
\caption{Average MSE for $\hat{\Bcal}$ obtained from the standard CP regression, symmetric CP regression and the proposed symmetric tensor regression, under n=500, 700, 900, 1100, based on 100 replications. In the parenthesis are the standard deviation. We use bold-font letters to highlight the best outcome.}
\label{tab:sim-shape1}
\begin{center}
\begin{tabular}{|c|c|ccc|}
\hline
$\Bcal$ & Sample Size & CP Reg. & Symmetric CP Reg. & Symmetric Tensor Reg. \\ \hline
\multirow{ 4}{*}{\texttt{Circle}} & n=500 & 0.433(0.290) & 0.106(0.018) & \textbf{0.019}(0.005) \\
& n=700 & 0.156(0.023) & 0.045(0.005) & \textbf{0.015}(0.003) \\
&n=900 & 0.018(0.009) & 0.032(0.003) & \textbf{0.012}(0.002) \\
& n=1100 & 0.096(0.015) & 0.023(0.002) & \textbf{0.010}(0.001) \\ \hline

\multirow{ 4}{*}{\texttt{Cross}} & n=500 & 0.433(0.290) & 0.106(0.018) &  \textbf{0.019}(0.005) \\
& n=700 & 0.186(0.110) & 0.048(0.004) & \textbf{0.011}(0.002) \\
&n=900 & 0.080(0.053) & 0.023(0.015) & \textbf{0.007}(0.005) \\
&n=1100 &0.069(0.020) &0.023(0.001) & \textbf{0.007}(0.001) \\ \hline

\multirow{ 4}{*}{\texttt{Butterfly}} &n=500 &0.457(0.082) &0.182(0.026) &\textbf{0.050}(0.011) \\
&n=700 &0.239(0.040) &0.071(0.008) &\textbf{0.030}(0.004) \\
&n=900 &0.179(0.027) &0.043(0.004) &\textbf{0.024}(0.002) \\
&n=1100 &0.148(0.025) &0.030(0.003) &\textbf{0.021}(0.002) \\ \hline

\multirow{ 4}{*}{\texttt{Two Box}} &n=500 &0.217(0.037) &0.083(0.015) &\textbf{0.007}(0.003) \\
&n=700 &0.119(0.028) &0.034(0.004) &\textbf{0.004}(0.002) \\
&n=900 &0.082(0.014) &0.022(0.004) &\textbf{0.004}(0.001) \\
&n=1100 &0.062(0.014) &0.015(0.003) &\textbf{0.003}(0.001) \\ \hline

\multirow{ 4}{*}{\texttt{Three Box}} &n=500 &0.368(0.091) &0.175(0.033) &\textbf{0.035}(0.012) \\
&n=700 &0.144(0.063) &0.058(0.027) &\textbf{0.012}(0.007) \\
&n=900 &0.062(0.031) &0.026(0.014) &\textbf{0.005}(0.004) \\
&n=1100 &0.052(0.025) &0.015(0.009) &\textbf{0.004}(0.003) \\ \hline
\end{tabular}
\end{center}
\end{table}

\begin{table}[hbt]
\caption{Average MSE for $\hat{Y}$ with the standard CP regression, symmetric CP regression and the proposed symmetric tensor regression model, under n=500, 700, 900, 1100, based on 100 replications.}
\label{tab:sim-shape2}
\begin{center}
\begin{tabular}{|c|c|ccc|}
\hline
$\Bcal$ & Sample Size & CP Reg. & Symmetric CP Reg. & Symmetric Tensor Reg. \\ \hline
\multirow{ 4}{*}{\texttt{Circle}} & n=500 & 5.194(0.784) & 5.194(0.784) & \textbf{1.485}(0.204) \\
& n=700 & 3.045(0.244) & 3.045(0.244) & \textbf{1.628}(0.156) \\
&n=900 & 2.367(0.181) & 2.367(0.181) & \textbf{1.518}(0.096) \\
&n=1100 & 1.981(0.118) & 1.981(0.118) & \textbf{1.436}(0.072) \\ \hline

\multirow{ 4}{*}{\texttt{Cross}} &n=500 &  5.194(0.784) & 5.194(0.784) & \textbf{1.485}(0.204) \\
&n=700 & 2.876(0.262) & 2.876(0.262) & \textbf{1.339}(0.081) \\
&n=900 & 1.596(1.031) & 1.596(1.031) & \textbf{0.912}(0.589) \\
&n=1100 &1.323(0.882) &1.323(0.882) &\textbf{0.884}(0.570) \\ \hline

\multirow{ 4}{*}{\texttt{Butterfly}} &n=500 &8.485(1.151) &8.485(1.151) &\textbf{2.980}(0.484) \\
&n=700 &3.996(0.410) &3.996(0.410) &\textbf{2.309}(0.199) \\
&n=900 &2.829(0.220) &2.829(0.220) &\textbf{2.056}(0.151) \\
&n=1100 &2.319(0.172) &2.319(0.172) &\textbf{1.945}(0.118) \\ \hline

\multirow{ 4}{*}{\texttt{Two Box}} &n=500 &4.397(0.652) &4.397(0.652) &\textbf{1.235}(0.129) \\
&n=700 &2.468(0.176) &2.468(0.176) &\textbf{1.224}(0.074) \\
&n=900 &1.938(0.132) &1.938(0.132) &\textbf{1.197}(0.062) \\
&n=1100 &1.699(0.098) &1.699(0.098) &\textbf{1.194}(0.050) \\ \hline

\multirow{ 4}{*}{\texttt{Three Box}} &n=500 &8.159(1.368) &8.159(1.368) &\textbf{2.242}(0.492) \\
&n=700 &3.088(0.539) &3.088(0.539) &\textbf{1.478}(0.229) \\
&n=900 &2.053(0.312) &2.053(0.312) &\textbf{1.324}(0.158) \\
&n=1100 &1.660(0.134) &1.660(0.134) &\textbf{1.222}(0.071) \\ \hline
\end{tabular}
\end{center}
\end{table}

We first observe from Figure \ref{fig:sim-shape} that the symmetric CP regression and the proposed symmetric tensor regression are able to correctly rover the shape of the signals, while the standard CP regression sometimes fails for such as the \texttt{butterfly} and \texttt{cross} settings. Moreover, we see that under the same sample size, our method recovers the signal more accurately than the baselines, which is also suggested by the numerical results in Table \ref{tab:sim-shape1}. The improvements partly owing to the fact that symmetric tensor regression employs much fewer parameters than the other two models, so under small sample sizes, we are able to achieve more efficient estimation.

Table~\ref{tab:sim-shape1} and Table~\ref{tab:sim-shape2} show that for all the sample sizes and signals that we experiment with, the symmetric tensor regression outperforms the other two methods, both in recovering the true signal and predicting the response. It is also worth mentioning that our approach has the smallest variances across all the settings, which again demonstrates the stability of the proposed symmetric tensor regression.

\subsection{Berkeley Aging Cohort Study data analysis}
\label{sec:realdata}
%\subsection{Exploratory data analysis}
%\label{sec:EDA}
We analyze the BAC data reviewed in Section~\ref{sec:introduction} that motivates our work. The dataset consists of 111 controls and 29 cases. Each subject has a PIB measure (that corresponds to the A$\beta$ deposition), age and gender. For our purpose, we use the PIB measure as a continuous response, together with the age and gender as the demographical covariates. The input tensor covariates are of dimension $80\times 80$ for each subject, which is the connectivity matrix for the 80 regions of interest (ROI) computed from different measures. The distribution of PIB measure is highly skewed, so by the standard practice for dealing with the skewed response, we use the logarithm of PIB measure (log(PIB)) as the response.

The $80 \times 80$ connectivity matrix for each individual is computed from their resting-state functional magnetic resonance imaging data. As we have mentioned in Section~\ref{sec:introduction}, the widely-used measures for connectivity matrix include the \emph{Pearson correlation} (\textbf{Corr}), \emph{mutual information} (\textbf{Minfo}), \emph{partial correlation} (\textbf{Pcorr}) and \emph{partial mutual information} (\textbf{Pminfo}). Here we compute all four of them for each individual. 
We apply the proposed symmetric tensor regression to detect ROI functional connectivities that show a strong association with the response. Due to the small sample size, we use 3-fold cross-validation to select tuning parameter $\rho$. We fix the rank $R=2$, because $R=2$ will result in approximately 150 free parameters, which is comparable to the sample size. Due to the imbalance of our dataset (111 controls and 29 cases), the samples in each fold in cross-validation are drawn separately from cases and controls and then combined together. 

\begin{figure}[hbt]
\caption{Quantile-quantile plots of Y and $\hat{Y}$ for all four types of connectivity measures. In the first row, the PIB measure is used as the response, and in the second row, the log(PIB) is used as the response.}
\label{fig:qqplot}
\begin{center}
\begin{tabular}{m{.22\textwidth}m{.22\textwidth}m{.22\textwidth}m{.22\textwidth}}
Corr &Minfo &Pcorrr &Pminfo \\
\includegraphics[scale=0.119]{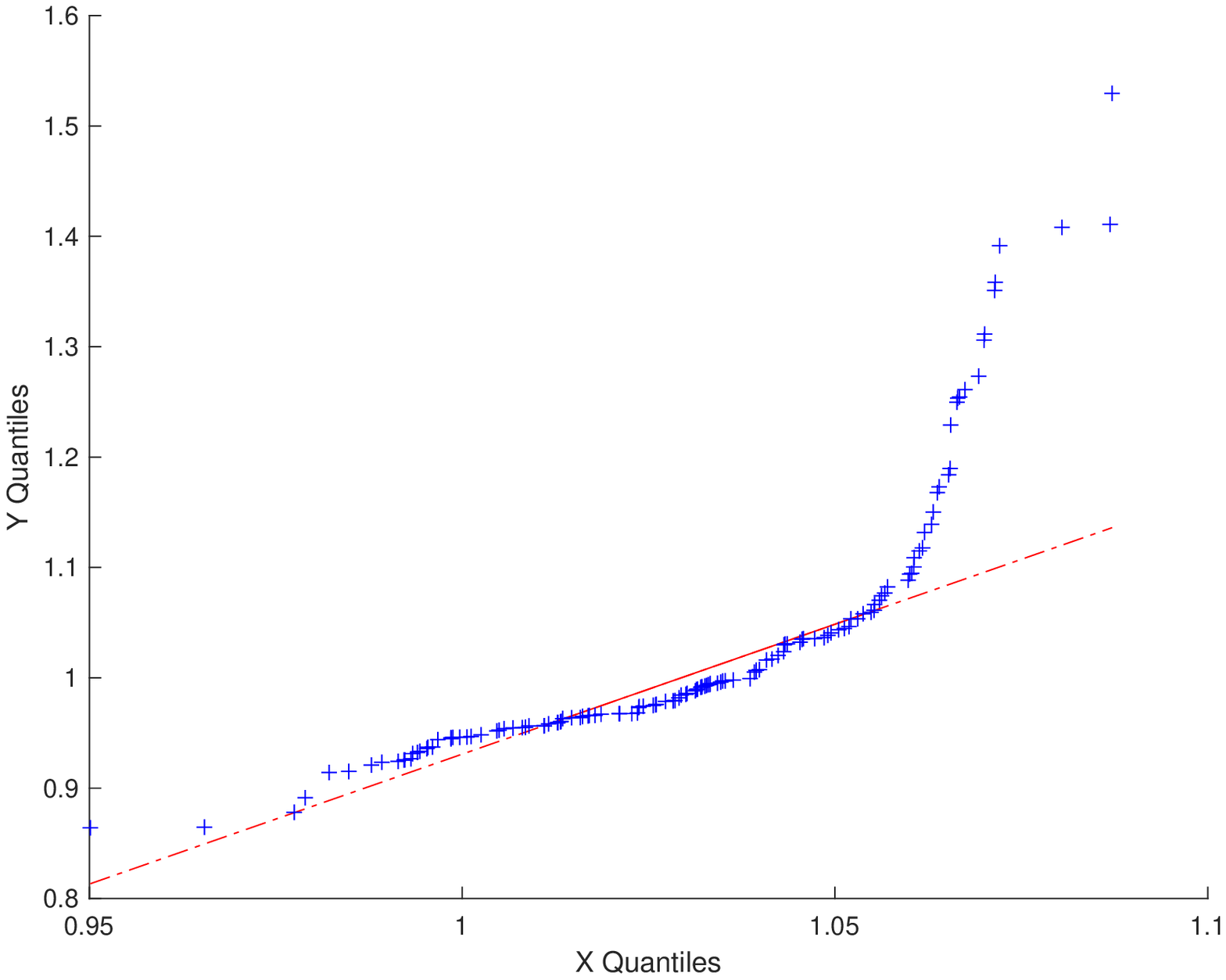} & \includegraphics[scale=0.119]{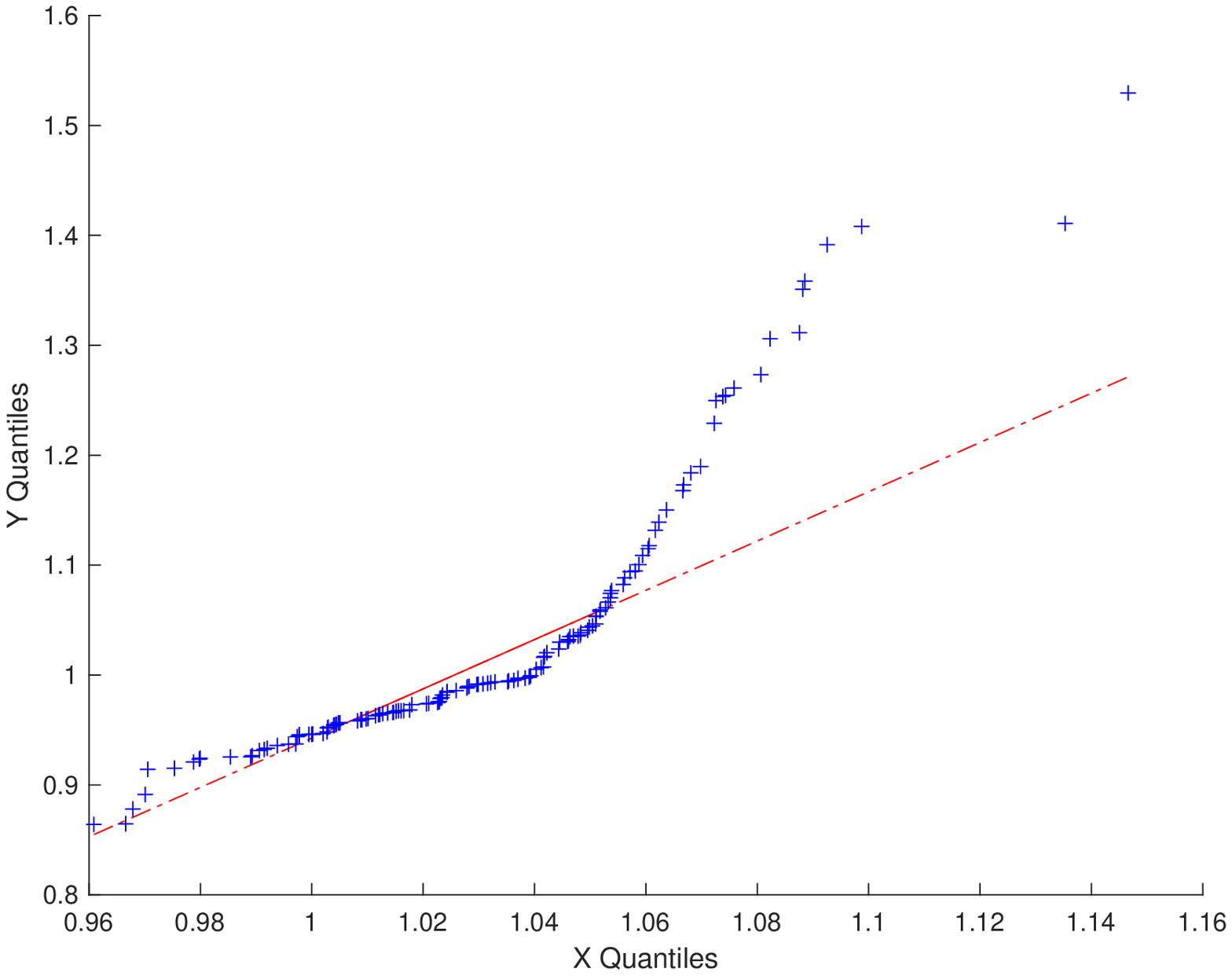} & \includegraphics[scale=0.119]{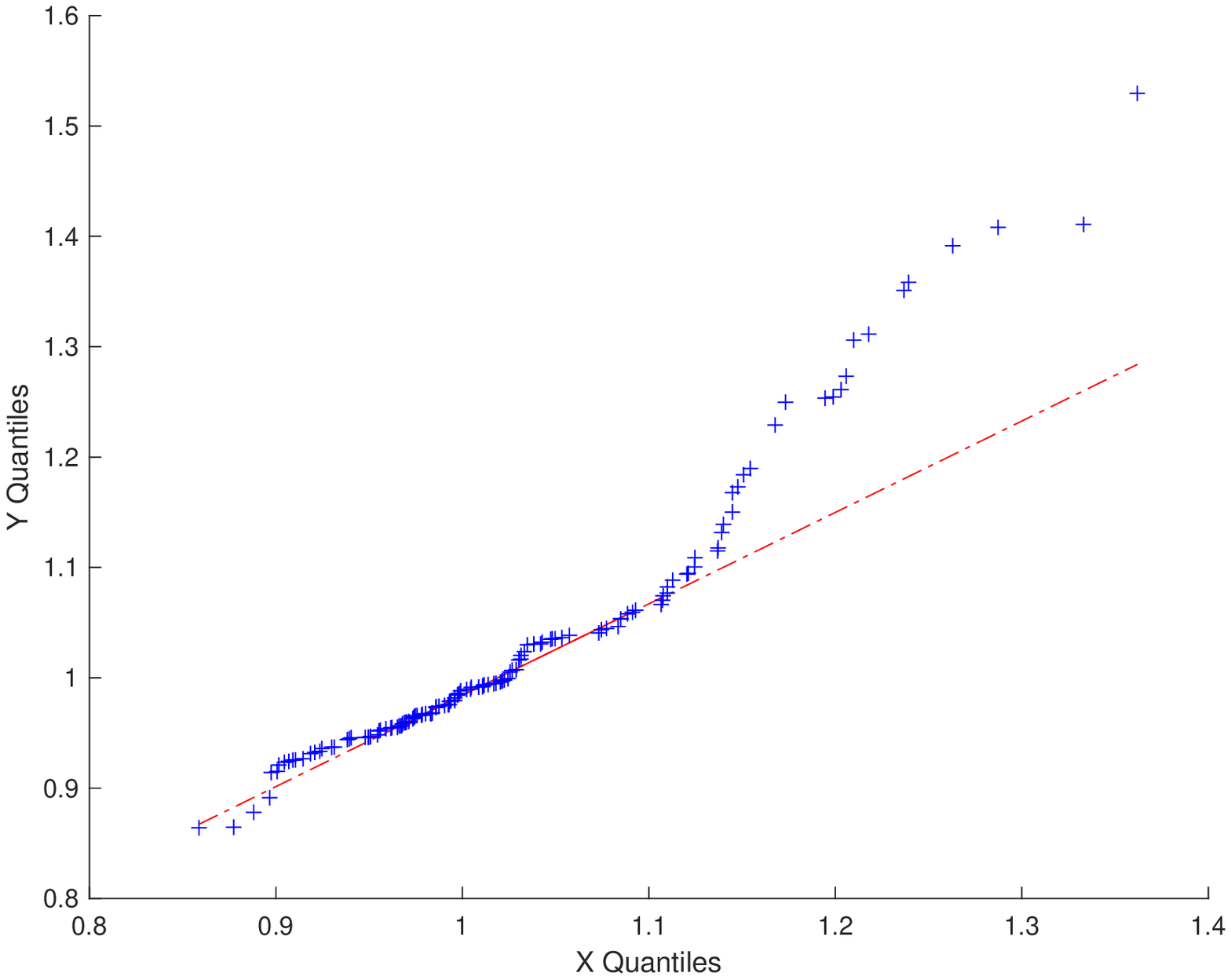} & \includegraphics[scale=0.119]{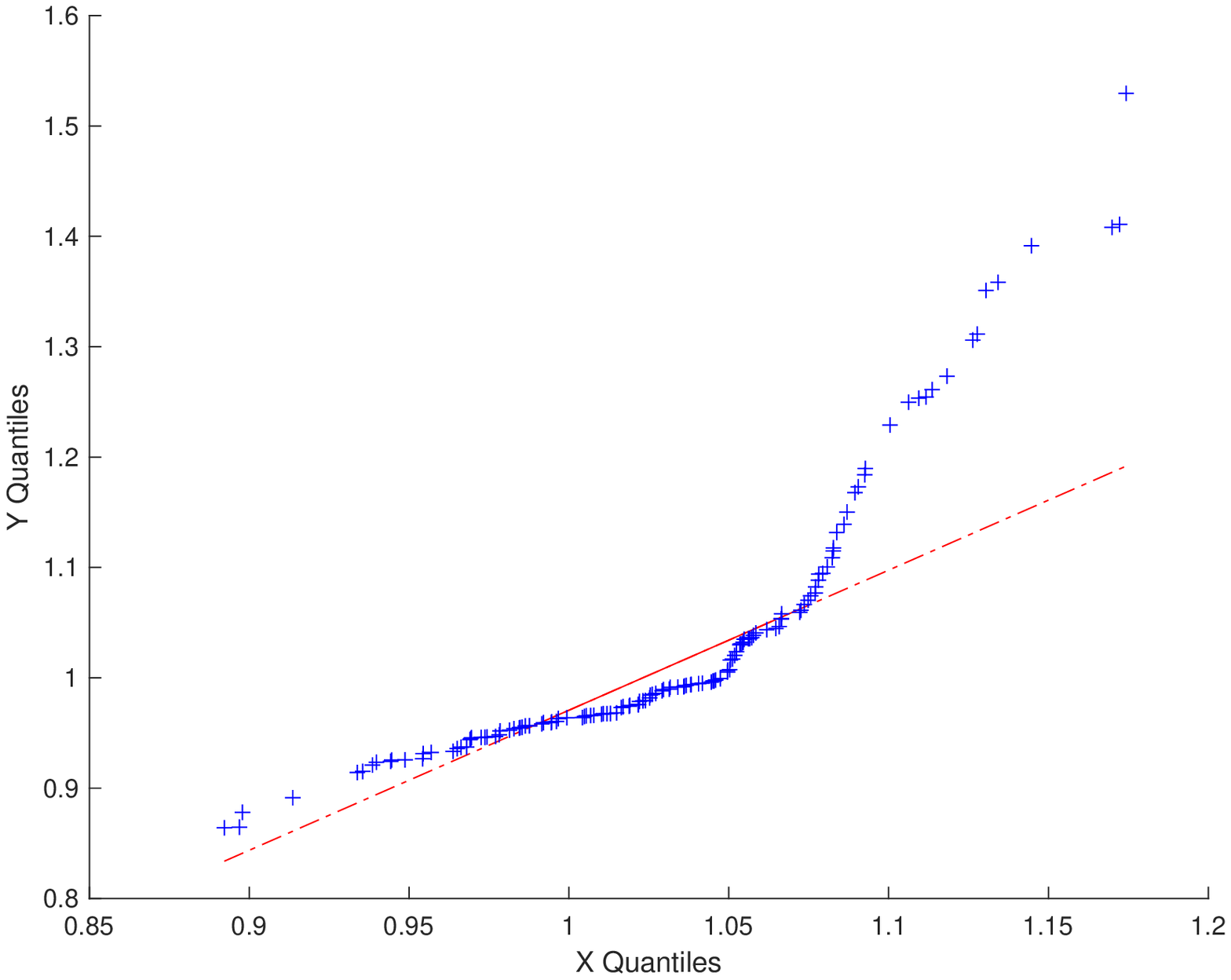} \\
\includegraphics[scale=0.119]{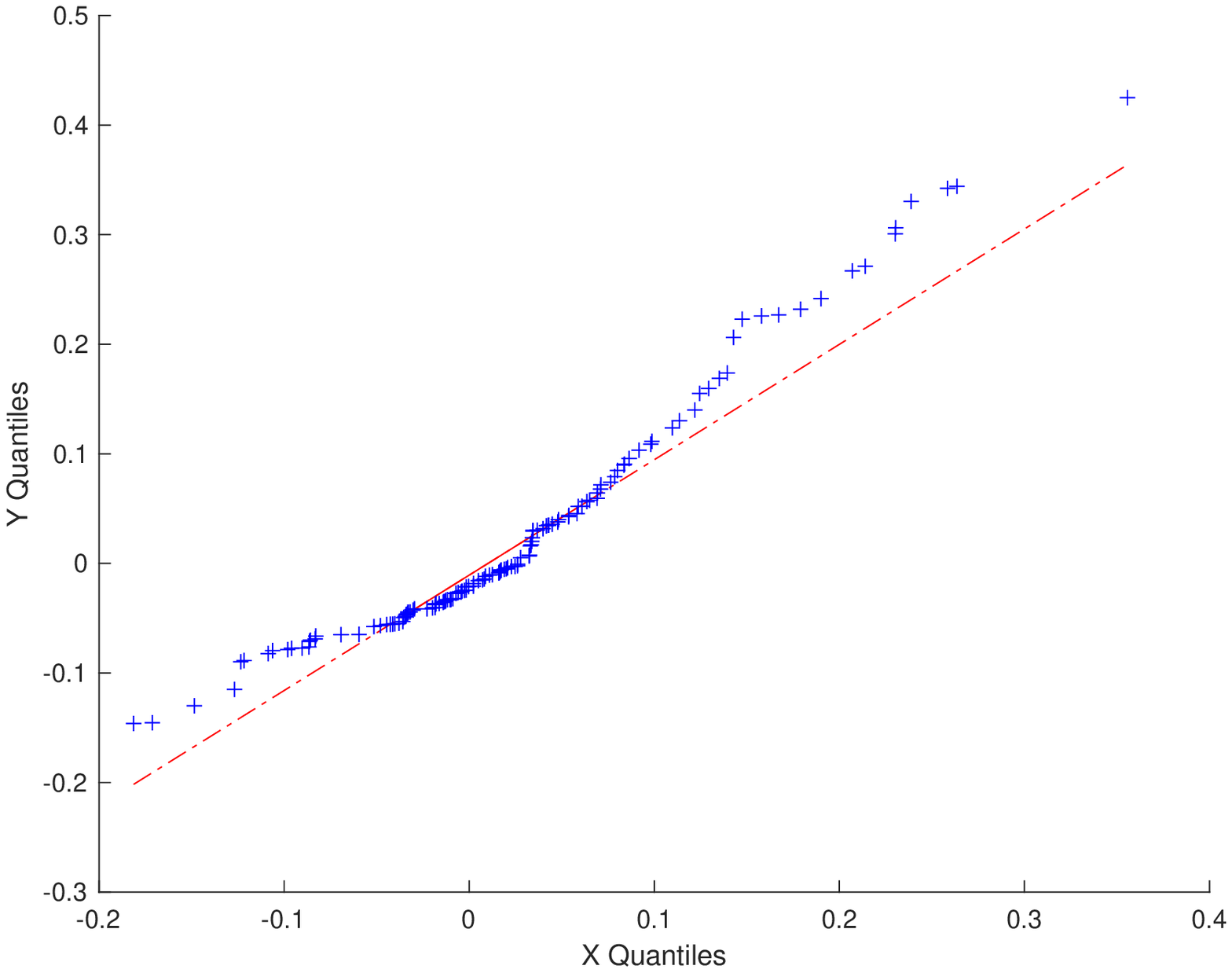} & \includegraphics[scale=0.119]{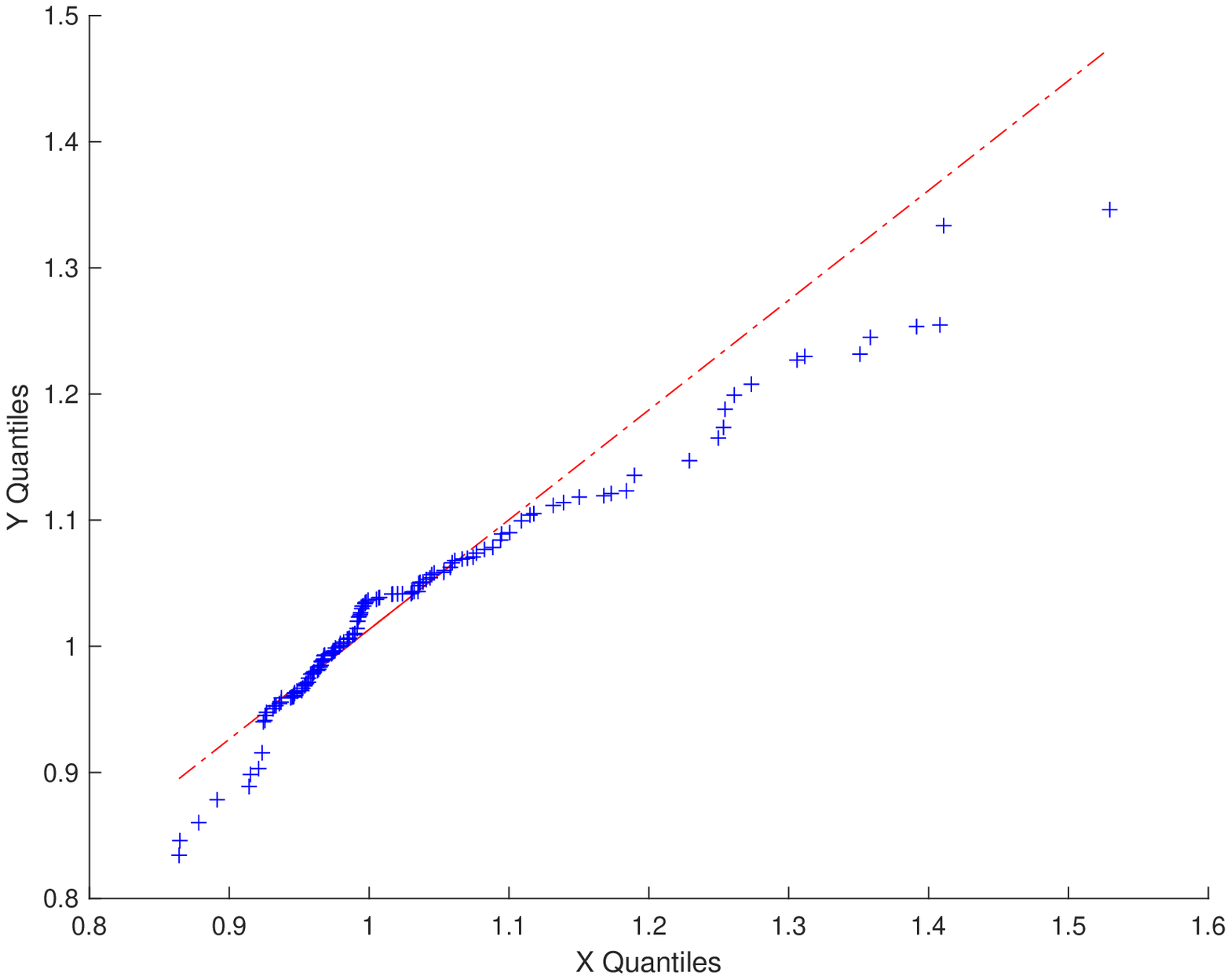} & \includegraphics[scale=0.119]{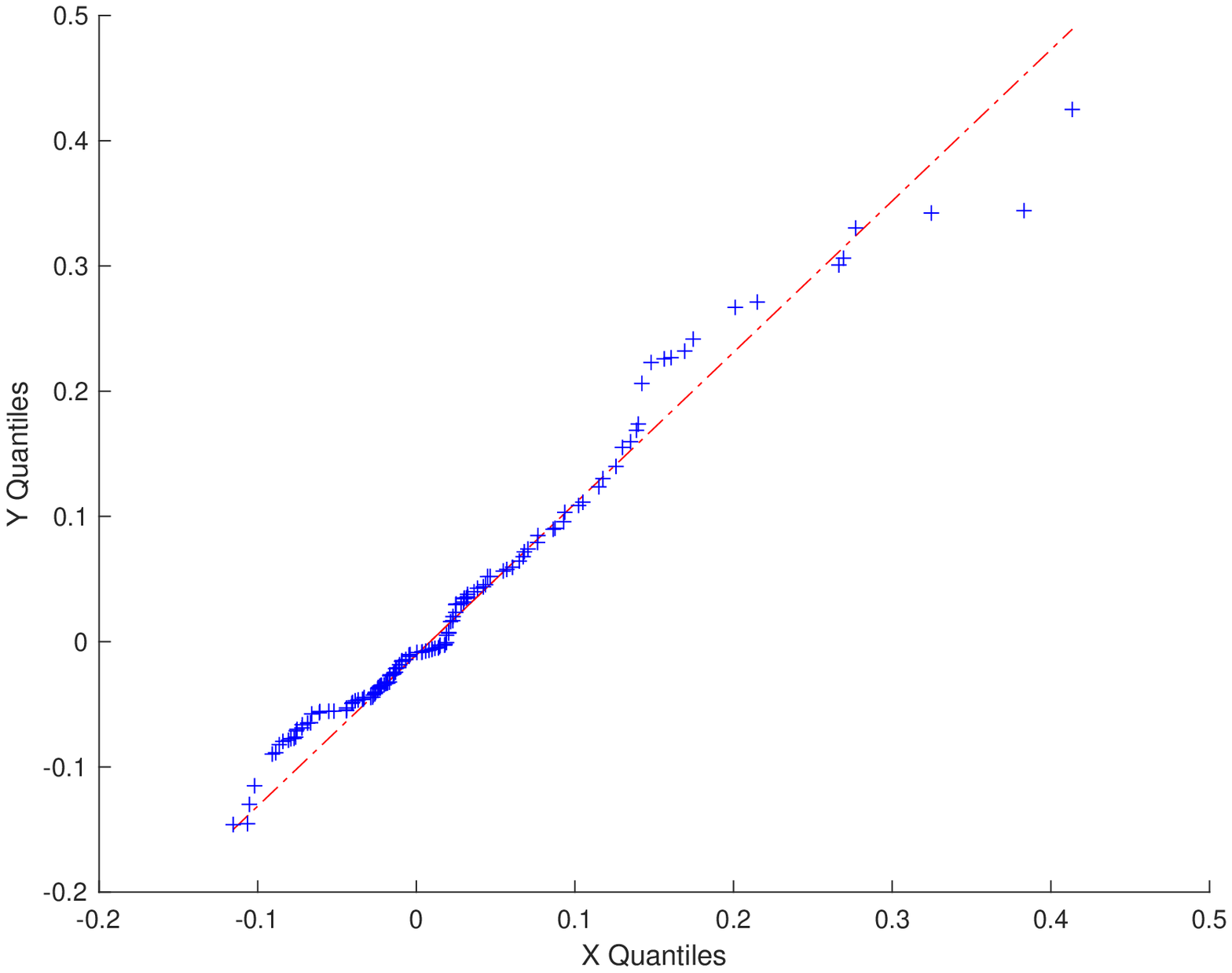} & \includegraphics[scale=0.119]{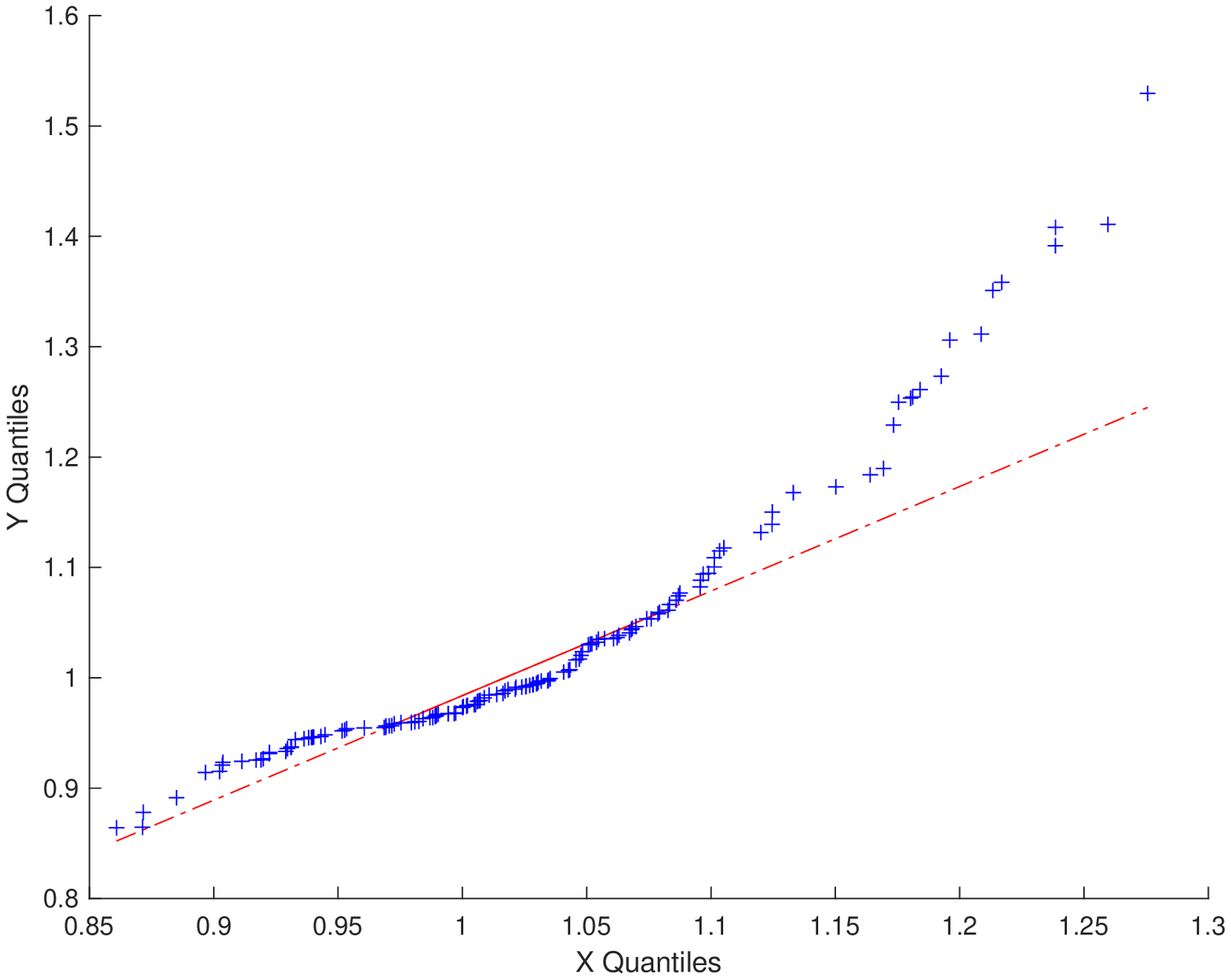} \\
\end{tabular}
\end{center}
\end{figure}

We report the MSE for the fold-wise prediction in cross-validation under the selected tuning parameter, as well as the MSE for final prediction on the whole dataset (averaged over ten runs). Table~\ref{tab:cv-pib} shows the result when using PIB measure as the response, and Table~\ref{tab:cv-logpib} shows that of using log(PIB) as the response. To further evaluate the performance under different types of connectivity measures with PIB and log(PIB) as the response, we provide the quantile-quantile plot for $\hat{Y}$ and $Y$ under all scenarios, in Figure~\ref{fig:qqplot}.

\begin{table}[htb]
\caption{Cross-validation MSE in each fold (MSE$(\hat{Y}_k, Y_k)$ where the subscript $k$ indicates the $k^{th}$ fold in the cross-validation), with PIB as response (averaged over 10 runs). \textbf{Corr} denotes connectivity matrix computed via the Pearson correlation, \textbf{Minfo} denotes mutual information, \textbf{Pcorr} denotes or partial correlation and \textbf{Pminfo} denotes partial mutual information.}
\label{tab:cv-pib}
\begin{center}
\begin{tabularx}{\textwidth}{|X|X|X|X|X|}
\hline
MSE& \textbf{Corr} & \textbf{Minfo} & \textbf{Pcorr} & \textbf{Pminfo} \\ \hline
Fold 1 & 0.011 & 0.010 & 0.014 & 0.008 \\ \hline
Fold 2 & 0.016 & 0.014 & 0.027  & 0.013\\ \hline
Fold 3 & 0.015 & 0.017 & 0.022 & 0.015\\ \hline
MSE$(\hat{Y}, Y)$ & 0.014 & 0.012 & 0.018 & 0.010 \\ \hline
\end{tabularx}
\end{center}
\end{table}

\begin{table}[hbt]
\caption{Cross-validation MSE in each fold with log(BIP) as the response (averaged over 10 runs).}
\label{tab:cv-logpib}
\begin{center}
\begin{tabularx}{\textwidth}{|X|X|X|X|X|}
\hline
MSE& \textbf{Corr} & \textbf{Minfo} & \textbf{Pcorr} & \textbf{Pminfo} \\ \hline
Fold 1 & 0.023 & 0.025 & 0.008 & 0.037 \\ \hline
Fold 2 & 0.026 & 0.012 & 0.014  & 0.028 \\ \hline
Fold 3 & 0.026 & 0.029 & 0.011 & 0.029 \\ \hline
MSE$(\hat{Y}, Y)$ & 0.004 & 0.004 & \textbf{0.0027} & 0.004 \\ \hline
\end{tabularx}
\end{center}
\end{table}

Table~\ref{tab:cv-pib} and Table~\ref{tab:cv-logpib} show that when using log(PIB) as the response, symmetric tensor regression leads to better and more stable predictions, partly because the log transformation alleviates the skewness in the data. The quantile-quantile plots also justify this in Figure~\ref{fig:qqplot}, where we notice that the symmetric tensor regression model does not the distribution tails well when using PIB measure as response. So we move on with the results where log(PIB) measure is used as the response. We also notice that that when using the partial correlation as connectivity measures, we obtain the best predictions in terms of MSE. As a matter of fact, partial correlation reflects the marginal correlation among ROIs and has been successfully applied to many neuroimaging analysis. So we proceed with the $\hat{\Bcal}$ where the connectivity matrix is computed from partial correlation and log(PIB) is used as the response.

\begin{figure}[hbt]
\caption{Top 20 connectivities with largest coefficients in $\hat{\mathcal{B}}$, with log(PIB) as the response and the connectivity matrix is computed via the partial correlation. The \textbf{red} edges indicate connectivities with positive coefficients, the \textbf{blue} edges indicate connectivities with negative coefficients.}
\label{fig:connec-plot}
\begin{center}
\includegraphics[scale=0.6]{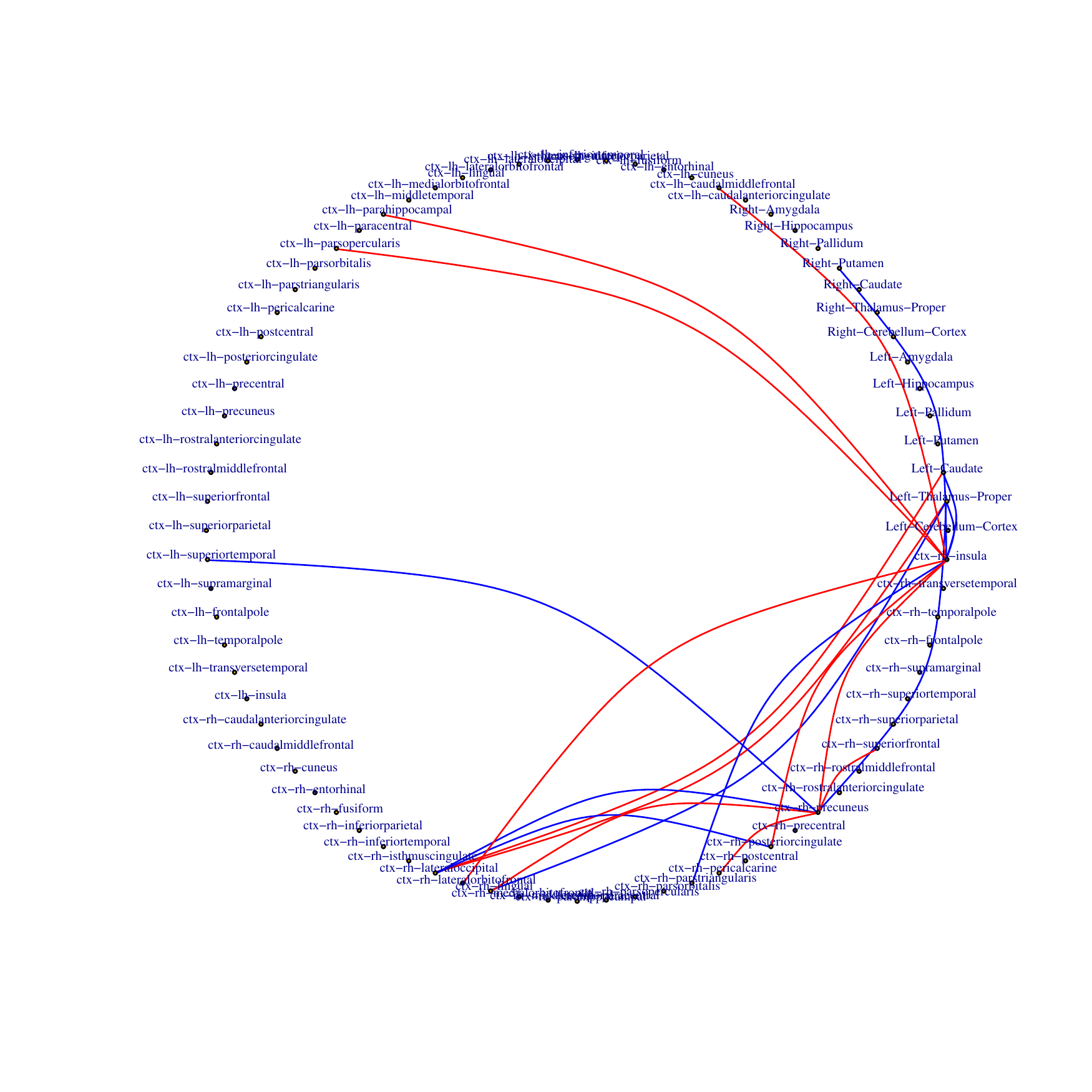}
\end{center}
\end{figure}

\begin{figure}
\caption{The heatmap of coefficients for: (a). all pairs of regions (left); (b). pairs of top 10 regions with largest overall absolute coefficients in $\hat{\mathcal{B}}$ (right).}
\label{fig:connec-heatmap}
\begin{center}
\begin{tabular}{m{.5\textwidth}m{.5\textwidth}}
(a) All regions & (b) Top 10 regions \\
\includegraphics[scale=0.35]{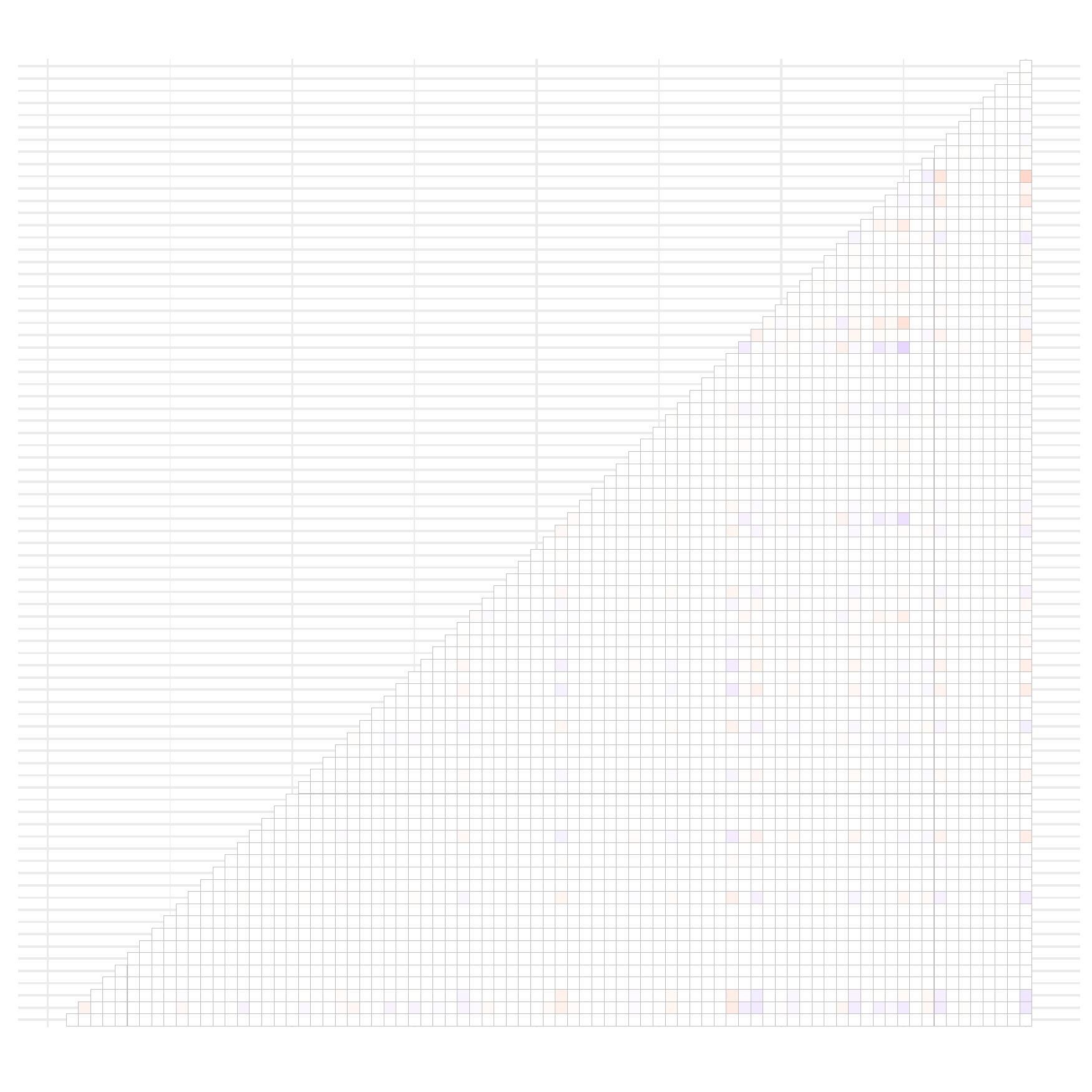} & \includegraphics[scale=0.45]{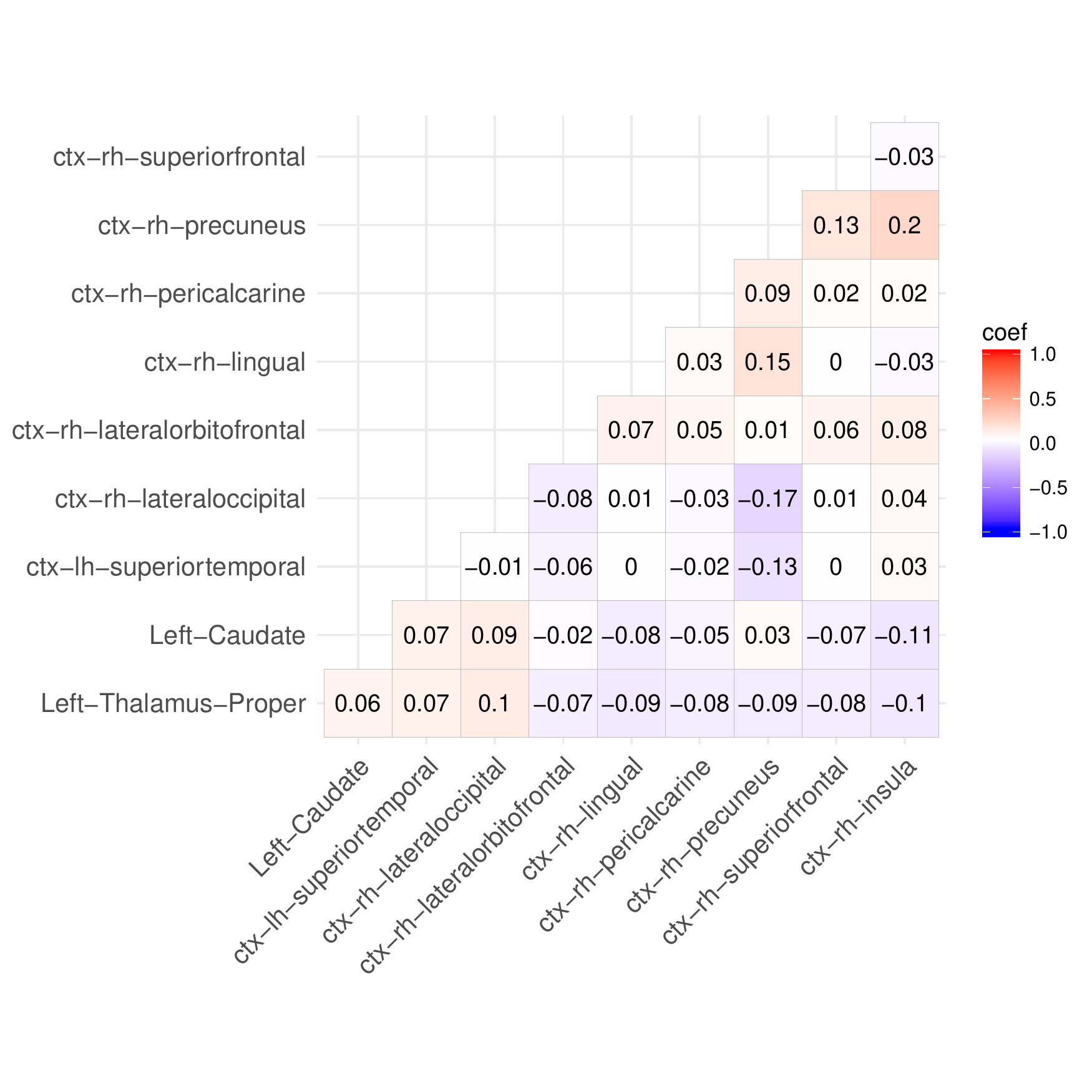} 
\end{tabular}
\end{center}
\end{figure}

We first visualize the estimated coefficient matrix $\hat{\mathcal{B}}$ in Figure~\ref{fig:connec-heatmap}. The high degree of sparsity of the outcome is partly due to the variable-selection ability of our method. We also visualize the most significant ROIs according to the estimated coefficient matrix in Figure~\ref{fig:connec-heatmap}. Although the estimated $\hat{\mathcal{B}}$ is sparse, for the sake of space, we also visualize the top 20 connected pairs that have the strongest association, i.e. largest coefficients (in magnitude), in Figure~\ref{fig:connec-plot} and list them in Table~\ref{tab:top-connec}. From Table~\ref{tab:top-connec}, we observe that the top six coefficients in magnitude correspond to connectivities that involve the \texttt{Precuneus} region. The \texttt{Insula} region is also crucial for the functional connectivity network in association with the response, since 10 of the top 20 associations correspond to connectivities that involve the \texttt{Insula} region. The functional connectivity with the strongest association is shown between the \texttt{Insula-Precuneus} pair, whose coefficient is significantly larger than the others. 

\begin{table}[hbt]
\caption{Top 20 connectivities with largest coefficients (in magnitude) in $\hat{\mathcal{B}}$, with log(PIB) as the response and the connectivity matrix is computed via the partial correlation. The table is arranged in a decreasing order according to the magnitude in coefficients.}
\label{tab:top-connec}
\begin{center}
\begin{tabularx}{\textwidth}{|c|X|X|c|}
\hline
Rank & ROI & ROI & $\hat{\mathcal{B}_{ij}}$ \\ \hline
1& ctx-rh-precuneus& ctx-rh-insula& 0.202 \\ \hline
2& ctx-rh-precuneus& ctx-rh-lateraloccipital& -0.172 \\ \hline
3&  ctx-rh-precuneus& ctx-rh-lingual& 0.148 \\ \hline
4& ctx-rh-precuneus& ctx-rh-superiorfrontal& 0.128 \\ \hline
5& ctx-rh-precuneus & ctx-lh-superiortemporal& -0.126 \\ \hline
6& Left-Caudate& ctx-rh-insula& -0.106 \\ \hline
7& ctx-rh-posteriorcingulate& ctx-rh-insula& 0.104 \\ \hline
8& Left-Thalamus-Proper& ctx-rh-lateraloccipital& 0.102 \\ \hline
9& Left-Thalamus-Proper& ctx-rh-insula& -0.096 \\ \hline
10& Left-Caudate& ctx-rh-lateraloccipital &0.093\\ \hline
11& ctx-rh-lateraloccipital &ctx-rh-posteriorcingulate& -0.092 \\ \hline
12& Left-Thalamus-Proper& ctx-rh-lingual& -0.088 \\ \hline
13& ctx-rh-precuneus & ctx-rh-pericalcarine &0.088 \\ \hline
14& ctx-lh-caudalmiddlefrontal& ctx-rh-insula& 0.088 \\ \hline
15& ctx-lh-parahippocampal& ctx-rh-insula& 0.087 \\ \hline
16& ctx-lh-parsopercularis& ctx-rh-insula& 0.087 \\ \hline
17& ctx-rh-precuneus & Left-Thalamus-Proper& -0.086 \\ \hline
18& ctx-rh-parstriangularis& ctx-rh-insula& -0.086 \\ \hline
19& Right-Putamen& ctx-rh-insula& -0.084 \\ \hline
20& ctx-rh-lateralorbitofrontal& ctx-rh-insula& 0.081 \\ \hline

\end{tabularx}
\end{center}
\end{table}

The literature also supports our findings in the neuroscience domain. In one paper, the authors have thoroughly studied various aspects of the functional connectivity mapping of the human \texttt{Precuneus} by resting-state fMRI \cite{zhang2012functional}. In their article, they review a number of studies which have described the altered pattern of functional connectivity of the \texttt{Precuneus} as a pathognomonic marker of the early Alzheimer’s disease, and further elucidate the functions of the \texttt{Precuneus}. In another study, the authors discover that the connectivity maps of commonly atrophied regions-of-interest support the \texttt{Precuneus} network involvement across AD variants \cite{lehmann2013intrinsic}. More recent work reveals that aging has a major effect on functional brain interactions throughout the entire brain, whereas AD is distinguished by additional diminished posterior \texttt{Precuneus} connectivity with the default mode network \cite{klaassens2017diminished}. Also, alternation in the \texttt{Insula} connectivity has been consistently implicated in diseases, including autism, frontotemporal dementia, and schizophrenia \cite{chand2017racial}. Several MRI studies have identified the \texttt{insular} grey matter loss, abnormal \texttt{insular} activity and disrupted \texttt{insular} network in the early stage of AD \cite{liu2018altered}. Another study finds altered functional connectivity of the \texttt{insular} subregions in Alzheimer’s disease, and concluded that functional connectivities in those subregions are affected in the AD patients \cite{liu2018altered}.

\section{Discussion}
\label{sec:discussion}
We propose a symmetric tensor regression model based on the symmetric tensor decomposition. The model is motivated by real-world applications where the input has super-symmetric structures such as the functional connectivity matrix in rs-fMRI studies. We provide detailed comparisons, both analytically and numerically, between the symmetric tensor regression, standard CP regression and symmetrized CP regression. We show that the symmetric tensor regression model better leverages the problem structure and leads to improved performances in a variety of settings. For the optimization part, we develop an efficient proximal gradient algorithm, and leverage the more flexible symmetrized CP regression to construct the initial values that are shown to be effective empirically. Our approach helps identify several regions of interests for Alzheimer’s disease in the Berkeley Aging Cohort Study.

% \bibliographystyle{unsrt}  

\clearpage
\bibliographystyle{abbrv}
\bibliography{references}

\end{document}